%% file: main.tex
\pdfoutput=1

\documentclass[11pt]{article}

\usepackage[final]{acl}
\usepackage{times}
\usepackage{latexsym}
\usepackage{placeins}
\usepackage[T1]{fontenc}
\usepackage[utf8]{inputenc}
\usepackage{microtype}
\usepackage{inconsolata}
\usepackage{graphicx}
\usepackage{microtype}
\usepackage{graphicx}
\usepackage{booktabs} 
\usepackage{tcolorbox}
\usepackage{caption}
\usepackage{subcaption}
\DeclareCaptionFormat{custom}
{%
    \textbf{#1#2}{#3}
}
\usepackage{float} 
\usepackage{wrapfig}
\usepackage{tabularx}
\newcolumntype{Y}{>{\centering\arraybackslash}X}
\usepackage{svg}
\usepackage{hyperref}

\usepackage{amsmath,amsfonts,bm}
\usepackage{amssymb}
\usepackage{amsthm}
\usepackage{bbm}
\usepackage{mathtools}      
\usepackage{xfrac}          
\usepackage[capitalize,noabbrev]{cleveref}
\usepackage[group-separator={,},group-minimum-digits={3}]{siunitx}
\usepackage{enumitem}

\theoremstyle{plain}

\theoremstyle{definition}

\theoremstyle{remark}

\usepackage{listings}
\usepackage{upquote}  
\usepackage{xcolor}
\definecolor{codegreen}{rgb}{0,0.6,0}
\definecolor{codegray}{rgb}{0.5,0.5,0.5}
\definecolor{codepurple}{rgb}{0.58,0,0.82}
\definecolor{backcolour}{rgb}{0.95,0.95,0.92}
\lstdefinestyle{mystyle}{
    backgroundcolor=\color{backcolour},   
    commentstyle=\color{codegreen},
    keywordstyle=\color{magenta},
    numberstyle=\tiny\color{codegray},
    stringstyle=\color{codepurple},
    basicstyle=\ttfamily\footnotesize,
    breakatwhitespace=false,         
    breaklines=true,                 
    captionpos=b,                    
    keepspaces=true,                 
    numbers=left,                    
    numbersep=5pt,                  
    showspaces=false,                
    showstringspaces=false,
    showtabs=false,                  
    tabsize=2,
    xleftmargin=\parindent,
}
\input{tex/math_commands.tex}

\input{tex/util_commands.tex}

\title{
Training Dynamics Underlying Language Model Scaling Laws:\\ Loss Deceleration and Zero-Sum Learning
}

\author{
 \textbf{Andrei Mircea \textsuperscript{1,2,3}}, 
 \textbf{Supriyo Chakraborty \textsuperscript{3}}, 
 \textbf{Nima Chitsazan\textsuperscript{3}}, 
 \\
  \textbf{Milind Naphade\textsuperscript{3}}, 
  \textbf{Sambit Sahu\textsuperscript{3}}, 
 \textbf{Irina Rish\textsuperscript{1,2}}, 
 \textbf{Ekaterina Lobacheva\textsuperscript{1,2}}
\\
 \textsuperscript{1}Mila -- Quebec AI Institute, 
 \textsuperscript{2}Université de Montréal, 
 \textsuperscript{3}Capital One 
\\
 \small{
   \textbf{Correspondence:} \href{mailto:email@domain}{mirceara@mila.quebec}
 }
}

\begin{document}
\maketitle
\begin{abstract}
This work aims to understand how scaling improves language models, specifically in terms of training dynamics. 
We find that language models undergo \textit{loss deceleration} early in training—an abrupt slowdown in the rate of loss improvement, resulting in piecewise linear behaviour of the loss curve in log-log space.
Scaling up the model mitigates this transition by (1) decreasing the loss at which deceleration occurs, and (2) improving the log-log rate of loss improvement after deceleration. 
We attribute loss deceleration to a type of degenerate training dynamics we term \textit{zero-sum learning} (ZSL).
In ZSL, per-example gradients become systematically opposed, leading to destructive interference in per-example changes in loss. As a result, improving loss on one subset of examples degrades it on another, bottlenecking overall progress. 
Loss deceleration and ZSL provide new insights into the training dynamics underlying language model scaling laws, and could potentially be targeted directly to improve language models independent of scale. 
We make our code and artefacts available at: \\
\url{https://github.com/mirandrom/zsl} \\
\end{abstract}

\section{Introduction}
\label{sec.intro}
\parnl{What mechanisms underlie scaling laws?}
Increasing language model (LM) size empirically improves cross-entropy loss with power-law behaviour, which can be accurately described with scaling laws \cite{kaplan_scaling_2020}. Despite their predictive capabilities, scaling laws offer limited insight into the underlying mechanism \cite{stumpf_critical_2012}; i.e. they do not explain \textit{how} scaling improves loss. 
This question is particularly interesting because, by identifying and understanding such a mechanism \cite{glennan_routledge_2017}, we may become able to target it directly and improve models independent of scale. 

While several recent works have sought to explain scaling laws based on notions of e.g. intrinsic model capacity \cite{sharma_scaling_2022}, data distribution properties \cite{michaud_quantization_2023}, or asymptotic behaviour \cite{bahri_explaining_2024}, mechanistic explanations that can inform new approaches and drive principled progress  (beyond resource-intensive scaling) remain under-explored. In particular, little is known about the changes in training dynamics that underlie scaling improvements, which our work addresses.

\begin{figure}[t]
     \centering 
    \includegraphics[width=\linewidth]{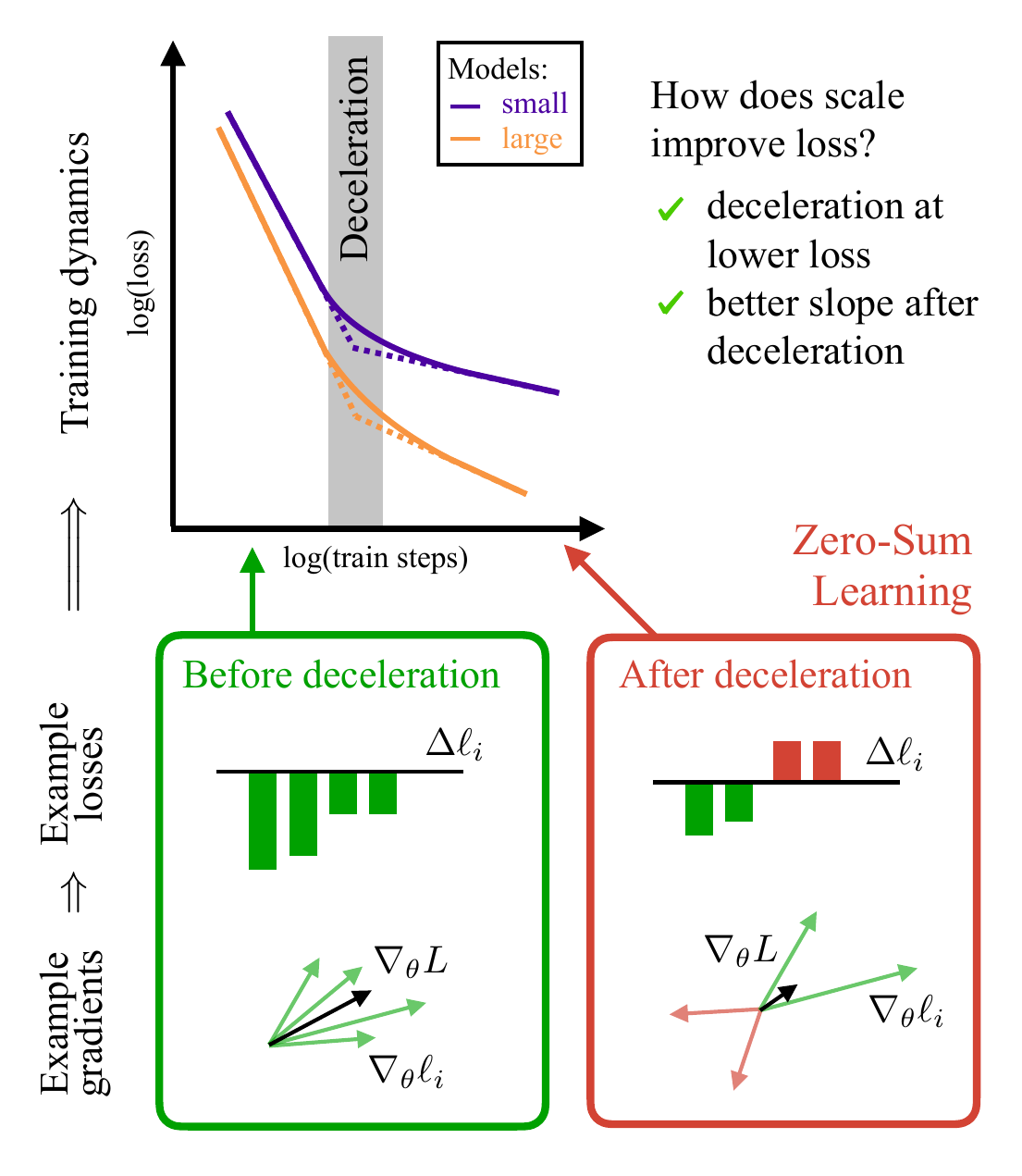}
    \caption{Paper overview. Top: Language model loss curves exhibit deceleration early in training. Scaling model size affects at which loss level this transition happens and how severe it is. Bottom: Loss deceleration can be attributed to \textit{zero-sum learning} dynamics: per-example gradients $\dldwx{i}$ become systematically opposed, leading to competing changes in per-example losses $\Dlx{i}$ and an overall slowdown in model loss improvement.
    }
    \label{fig.overview}
\end{figure}

\parnl{Loss deceleration underlies scaling laws.}
We find that scaling improvements can be explained in terms of training dynamics via \textit{loss deceleration}, a phenomenon where rates of loss improvement slow down abruptly, resulting in loss curves that are piecewise-linear in log-log space (see~\reffig{fig.overview}, top). Importantly, scaling improvements can be decomposed in terms of deceleration measurements.
Specifically, we show that scaling up the model size \textbf{(1)} decreases the loss at which deceleration occurs, and \textbf{(2)} increases the log-log rate of loss improvement after deceleration. 
This connection suggests that the mechanism behind deceleration (and the mitigating effects of scale) can help understand how scaling improves final model loss.

The piecewise-linear nature of deceleration suggests a qualitative transition in training dynamics. 
To the best of our knowledge, deceleration and the underlying transition in training dynamics has not been addressed in relevant prior works on e.g. loss plateaus \cite{yoshida_data-dependence_2020}, learning curve shapes \cite{hutter_learning_2021,viering_shape_2022}, or LM saturation \cite{godey_why_2024,mircea_gradient_2024}. 
Our work fills this gap by proposing a mechanistic explanation for loss deceleration and showing how it underlies scaling laws. 

\parnl{A mechanistic explanation of deceleration.}
We hypothesize that deceleration occurs as a result of degenerate zero-sum learning dynamics (see ~\reffig{fig.overview}, bottom). In ZSL, per-example gradients become systematically opposed, leading to destructive interference in loss improvements. In other words, loss can not be improved on one set of examples without degrading on another, thus bottlenecking the rate at which overall loss can improve. 
We verify this hypothesis against alternative explanations, characterizing and validating the proposed mechanism with a complementary empirical and theoretical results. 

As a mechanistic explanation \cite{kaplan_explanatory_2011}, zero-sum learning describes how the training dynamics of individual examples (i.e. their loss and gradients) behave and interact with one another to produce loss deceleration. This approach of understanding learning dynamics from the perspective of per-example gradient alignment and opposition is similar to \citet{mircea_gradient_2024}, but otherwise under-explored outside of tangential areas of research on e.g. improving multi-task learning \cite{liu_conflict-averse_2021}, or characterizing outliers in SGD \cite{rosenfeld_outliers_2023}. 

Importantly, we believe zero-sum learning 
can potentially be mitigated directly to improve loss independent of scale. Our findings offer new insights into how scaling improves loss by mitigating deceleration, and can provide a foundation for future work in this direction. 

\paragraph{Summary of findings and contributions} 
In \refsec{sec.decel} we identify loss deceleration as a novel qualitative transition in LM training dynamics. In particular, we show how scaling improvements can be explained in terms of mitigating deceleration. In \refsec{sec.expl}, we propose and validate a mechanistic explanation of deceleration based on destructive interference in per-example gradients and loss improvements. Lastly, in \refsec{sec.impr}, we connect these mechanisms to scaling improvements, showing how they are mitigated in ways that could be targeted directly and independent of scale. 

\paragraph{Methodology} We adapt the training setup of \citet{groeneveld_olmo_2024} and scaling experiments of \citet{kaplan_scaling_2020}, training and analyzing models between 14M and 472M parameters. Details on training and model analyses are in \refapp{app.meth}. 
We also provide ablation experiments with different model architectures, datasets, optimizers and training hyperparameters in \refapp{app.ablation}. 

\section{Loss deceleration in language models}
\label{sec.decel}
\begin{figure}[t]
     \centering 
    \includegraphics[width=0.8\linewidth]{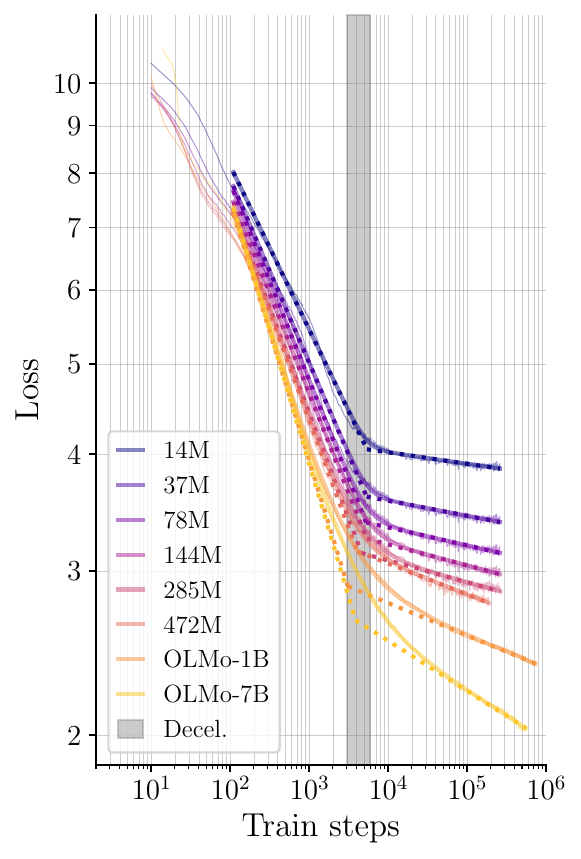}
    \vspace{-1em}
    \caption{
    Loss curves exhibit deceleration early in training (grey fill), and can be parametrically described with a one-break BNSL (\refeqn{eqn.bnsl}). The resulting BNSL fits are shown in bold, with the underlying piecewise-linear components shown as dashed lines.   
    We also include the OLMo 1B and 7B models \cite{groeneveld_olmo_2024}, showcasing similar behaviour at larger scales.
    }
    \label{fig.bnsl_loss_fit}
\end{figure}

\parnl{Characterizing loss deceleration.}
%
We find that LM loss curves typically exhibit an abrupt slow down in the rate of loss improvements early during training, in a transition we refer to as \textit{loss deceleration}. Notably, we see in \reffig{fig.bnsl_loss_fit} that loss deceleration is characterized by piecewise-linear behaviour in log-log space, consistent across different model scales and training setups, suggesting a qualitative transition in training dynamics.

An important observation from \reffig{fig.bnsl_loss_fit} is that loss improvements from scaling can be framed in terms of mitigating this transition, i.e. by improving: 
\setlist[enumerate,1]{leftmargin=3em}
\begin{enumerate}[
    align=parleft, topsep=0ex,itemsep=0ex,partopsep=1ex,parsep=0.5ex, labelwidth=1em
]
    \item[$\bm{(1)}$] the loss at which deceleration occurs; and 
    \item[$\bm{(2)}$] the log-log loss slope after deceleration. 
\end{enumerate}
This suggests that, by understanding the mechanism underlying loss deceleration (and the mitigating effects of scale), we can shed light on \textit{how} scaling improves loss in terms of training dynamics. Such an understanding could in turn inform methods that directly target and mitigate deceleration independent of scale. However, to study how scale mitigates deceleration, we must first measure it. 

\parnl{Measuring loss deceleration with BNSL.}
%
In measuring loss deceleration, we want to capture the log-log piecewise-linear behaviour observed in \reffig{fig.bnsl_loss_fit} and quantify how it changes with scaling. Luckily, this type of function can be parametrically described and fit with smoothly broken power laws such as BNSL \cite{caballero_broken_2023}, particularly in the simplified one-break form:
\begin{equation}\label{eqn.bnsl}
    L(t) - a = \pp{bt^{-c_0}}\pp{1 + \pp{t/d_1}^{1/f_1}}^{-c_1f_1},
\end{equation}
where $L(t)$ is the loss at step $t$, and the remaining variables are the parameters being fit: $a$ represents the irreducible loss; $b$ the y-axis intercept $L(0)$; $c_0$ the log-log slope of the first linear segment; $c_1$ the difference between the slope of the second segment and the first; $d_1$ the step at which the break occurs; and $f_1$ the smoothness of the transition between segments. However, these parameters provide limited insight into how deceleration relates to loss.

\parnl{Quantifying the effect of deceleration on loss.}
%
For a more interpretable but nevertheless quantifiable connection between deceleration and loss, we can tease out the linear segments underlying \refeqn{eqn.bnsl}. 
Concretely, an estimate $\hat{L}_T$ of the loss $L_T$ at step $T>d_1$ can be expressed in terms of three measurements grounded in BNSL parameters\footnotemark: 
\footnotetext{See Appendix A.2 in \cite{caballero_broken_2023}.}
\begin{flalign}
   &\bm{\log(\hat{L}_T)} = \log\pp{L_d} - r_d\log\pp{T/t_d}&&
   \label{eqn.loss_estimate_decel}
   \\
   &\bm{\hat{L}_T} = L_d \pp{t_d / T}^{r_d}&&
   \notag
\end{flalign}
\setlist[enumerate,1]{leftmargin=2.5em}
\begin{enumerate}[align=parleft, topsep=0pt,itemsep=-1ex,partopsep=1ex,parsep=1ex, labelwidth=1em]
    \item[$\bm{t_d}$] $= d_1$, the step where deceleration occurs, or where the two segments intersect; 
    \item[$\bm{L_d}$] $= b {d_1}^{-c_0}$, the loss where deceleration occurs, or where the two segments intersect;
    \item[{$\bm{r_d}$}] $= c_0 + c_1$, the log-log rate of loss improvement after deceleration, or the negative log-log slope of the second segment.
\end{enumerate}
\vspace{1em}
Intuitively, we see that final loss is fundamentally a function of $L_d$, $r_d$, $t_d$; such that scaling improvements can be explained solely in terms of increased $r_d$ and decreased $L_d$, $t_d$. 
These measurements are reported in \reftab{tab.decel_measurements}, where we indeed observe monotonic improvements in $L_d$, $r_d$ and $t_d$ with increased model size\footnotemark. 
We also confirm that $\hat{L}_T$ is a valid approximation, typically within $1\%$ of $L_T$.
\footnotetext{
One notable outlier is $t_d$ in OLMo-$7$B, likely attributable to OLMo-$7$B using a warmup of \num{5000} rather than \num{2000} steps.
}

\begin{table}[!h]
  \caption{Loss deceleration measurements from \refeqn{eqn.loss_estimate_decel}: larger models have lower $L_d$, $t_d$ and higher $r_d$.
  }
  \label{tab.decel_measurements}
  \centering
  \resizebox{\columnwidth}{!}{
    \begin{tabular}{lccccccccc}
    \toprule
    Model && $\downarrow L_d$ & $\downarrow t_d$ & $\uparrow r_d$ && $\hat{L}_T$ & $L_T$ \\ 
    \midrule
    14M && $4.05$ & $5900$ & $0.013$ & & $3.86$ & $3.88$ \\ 
    37M && $3.60$ & $5900$ & $0.016$ & & $3.39$ & $3.40$ \\ 
    78M && $3.38$ & $5900$ & $0.020$ & & $3.14$ & $3.15$ \\ 
    144M && $3.25$ & $6000$ & $0.023$ & & $2.98$ & $2.99$ \\ 
    285M && $3.14$ & $5300$ & $0.025$ & & $2.85$ & $2.87$ \\ 
    472M && $3.16$ & $4600$ & $0.035$ & & $2.77$ & $2.80$ \\ 
    \midrule
    OLMo-1B && $2.86$ & $3700$ & $0.034$ & & $2.39$ & $2.40$ \\ 
    OLMo-7B && $2.64$ & $4600$ & $0.053$ & & $2.04$ & $2.03$ \\ 
    \bottomrule
  \end{tabular}
  }
\end{table}

Crucially, these are interpretable measurements of loss deceleration, allowing us to naturally describe and reason about scaling improvements in terms of training dynamics. For example, \refeqn{eqn.loss_estimate_decel} forms the basis of a novel scaling law functional form, with improved explanatory power as a result of being grounded in these interpretable quantities. 
While beyond the scope of this paper, we include preliminary results in \refapp{app.decel_scaling_laws}. 

More generally, this means we can shift our goal from understanding how scaling improves loss to understanding the mechanism underlying deceleration and how scaling improves $L_d$, $r_d$ and $t_d$. In the next section, we focus on the first question of understanding deceleration. 

\clearpage
\section{Explaining Loss Deceleration}
\label{sec.expl}
The log-log piecewise-linear behaviour of loss deceleration suggests that a qualitative change in training dynamics underlies the abrupt slowdown in loss improvements. 
Our goal in this section is to characterize this transition in training dynamics and provide a mechanistic explanation for loss deceleration. By ``mechanistic explanation'', we mean identifying and formalizing an underlying mechanism as defined in \citet{glennan_routledge_2017}.
To this end, we propose and verify the hypothesis that loss deceleration is a transition in training dynamics characterized by \textit{zero-sum learning}. 
\paragraph{Zero-sum learning (ZSL)} 
describes degenerate training dynamics where per-example gradients become systematically opposed, leading to significant destructive interference between per-example changes in loss. 
Put differently, ZSL corresponds to regions in parameter space where gradient descent cannot improve loss on one set of examples without commensurate loss degradation on another, effectively bottlenecking the overall rate of loss improvement. 
ZSL could therefore be a mechanistic explanation of how per-example gradients and loss changes interact to produce the abrupt slowdown in overall loss improvement seen in deceleration.

\paragraph{Verifying the ZSL hypothesis}
In the following sections, we break down the hypothesis that ZSL explains loss deceleration into atomic claims that we validate with empirical and theoretical results. 

\setlist[enumerate,1]{leftmargin=2em}
\begin{enumerate}[
    align=parleft, topsep=0ex,itemsep=0ex,partopsep=1ex,parsep=0.5ex
]
    \item[\textbf{\ref{sec.expl.measure}}] Introduces and defines measures of destructive interference used throughout our analysis.
    \item[\textbf{\ref{sec.expl.cooccur}}]  Confirms deceleration \textit{co-occurs} with increased destructive interference in per-example loss improvements and gradients.
    \item[\textbf{\ref{sec.expl.attrib}}] Demonstrates deceleration is \textit{primarily attributable} to destructive interference in per-example loss improvements.
    \item[\textbf{\ref{sec.expl.grads}}] Demonstrates destructive interference in loss improvements is \textit{primarily attributable} to destructive interference in gradients.
\end{enumerate}

\paragraph{Notation} 
    Let $\lossx{i}$ denote loss for example (i.e. token) $i$ in dataset $\dataset$, with overall loss $\Loss = \tsum_i \lossx{i} / \abstt{\dataset}$. Conversely, change in loss between steps $t_1,t_2$ is denoted as $\Delta_{t_1}^{t_2} \Loss = \tsum_{i}\Delta_{t_1}^{t_2} \loss_i / \abstt{\dataset}$. To reduce notation clutter, $t_1,t_2$ are sometimes omitted when evident from context or not relevant. Similarly, the overall gradient is denoted $\dLdw = \tsum_{i}\dldwx{i} / \abstt{\dataset}$ where $\dldwx{i}$ is the gradient for token $i$.

\subsection{Measuring Destructive Interference}\label{sec.expl.measure}
To measure zero-sum learning, we define destructive interference in per-token loss improvements $\Delta \lossx{i}$ as the proportion with which they cancel out in overall loss improvements $\Delta \Loss = \tsum_i \Delta \lossx{i} / \abstt{\dataset}$, with respect to an ideal scenario where there is no interference $\Delta \Loss^* = \tsum_i \abstt{\Delta \lossx{i}} / \abstt{\dataset}$: 
\begin{equation}\label{eqn.di}
    \DI{\Dlx{}} = 
        \frac
            {\Delta \Loss^* - \abs{\Delta \Loss}}
            {\Delta \Loss^*} 
        = 1 - \frac
            {\abs{\sum_i \Delta \lossx{i}}}
            {\sum_i \abs{\Delta \lossx{i}}}
\end{equation}

Similarly, we use coordinate-level destructive interference to measure gradient opposition, typically reporting $\DI{\dldwx{}}$ as the average over coordinates: 
\begin{equation}\label{eqn.di_grad}
    \vDI\pp{\dldwx{}}
        = 1 - \frac
            {\abs{\sum_i \dldwx{i}}}
            {\sum_i \abs{\dldwx{i}}}
\end{equation}

\noindent
Intuitively, $\DI{\Dlx{}}$ and $\DI{\dldwx{}}$ increase and approach $1$ with ZSL. Conversely, $\DI{\Dlx{}}$ and $\DI{\dldwx{}}$ decrease and approach $0$ with no ZSL. 

\subsection{Confirming Deceleration Occurs with ZSL}\label{sec.expl.cooccur}
\begin{figure}[t]{
     \centering
     \includegraphics[height=0.9\linewidth]{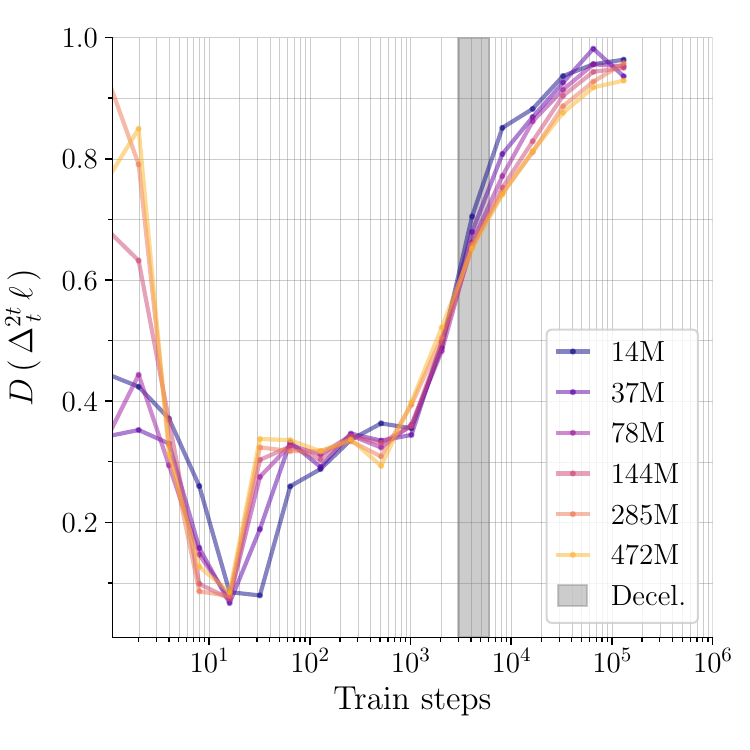}
    \caption{
        ZSL throughout training, as measured by destructive interference in per-token loss improvements. Deceleration co-occurs with a sharp increase in ZSL.
    }\label{fig.zsl.zsl_multistep_log2_DI}
    }
\end{figure}

To show that loss deceleration co-occurs with ZSL, we analyse the behavior of destructive interference in both loss improvements and gradients during training. 
In \reffig{fig.zsl.zsl_multistep_log2_DI}, we measure 
destructive interference in loss improvements
$\DI{\Dltt}$.   We observe that it exhibits a sharp increase, beginning just before deceleration, then converging towards its maximum. 
One important consideration is that these measurements are based on $\Dltt$ to smooth out noise from loss oscillations on too-small timescales. 
In practice, we find that $\DI{\Delta_{t_1}^{t_2} \ell}$ is mitigated as the number of steps $t_2 - t_1$ increases, such that \reffig{fig.zsl.zsl_multistep_log2_DI} is effectively under-reporting ZSL (\refapp{app.zsl_incr_steps}).

\begin{figure}[t]{
     \centering
     \includegraphics[height=0.9\linewidth]{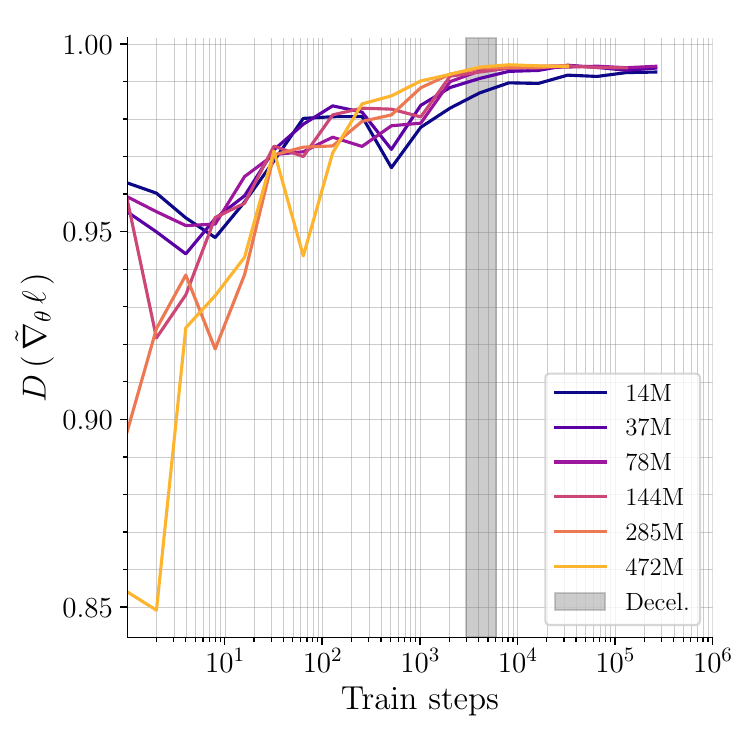}
    \caption{
        Gradient destructive interference (averaged over parameters) converges to a maximum with deceleration.
    }\label{fig.sgo.di_grads_avg}
    }
\end{figure}

In \reffig{fig.sgo.di_grads_avg}, we measure gradient destructive interference\footnotemark $\DI{\ptg{}{}}$ and find that it also converges to a maximum at the same time as deceleration. 
Surprisingly, gradient opposition turns out to be quite high even at the start of training.
Despite this high starting point, the increase in destructive interference leading up to deceleration still has a significant effect. In fact, increasing destructive interference in a sum beyond $0.9$ rapidly decreases the magnitude of that sum by several orders of magnitude, as can be seen in \reffig{fig.zsl.multistep.d_vs_m} and \refeqn{eqn.di_vs_ma} in the next section for $\DI{\Dlx{}}$ and $\DL$.
\footnotetext{A tractable proxy for per-example gradients, described and empirically validated in \refapp{app.measure_ptg}.} 

These results confirm that ZSL indeed co-occurs with loss deceleration, but are not sufficient evidence that ZSL is the underlying mechanism. 
The following sections will demonstrate how, in terms of per-token loss behaviour, deceleration is driven by ZSL rather than the alternative explanation.

\subsection{The Role of ZSL in Deceleration}\label{sec.expl.attrib}
Our framing of ZSL hypothesizes that deceleration is a result of destructive interference in loss improvements. In this section, we quantify the contribution of this destructive interference to overall loss improvements and show that it is indeed the main contributor to deceleration. 

\parnl{Quantifying the role of ZSL in deceleration.}
In terms of per-token loss improvements $\Delta \lossx{i}$, loss deceleration can occur for two (non mutually exclusive) reasons: 
(1) $\Dlx{i}$ increasingly cancel one another out due to ZSL; or 
(2) $\Dlx{i}$ increasingly shrink in magnitude across tokens. 
Destructive interference $\DI{\Dlx{}}$ in \refeqn{eqn.di} captures (1); while average magnitude $\MA{\Dlx{}}$ in \refeqn{eqn.ma} captures (2):
\begin{equation}\label{eqn.ma}
    \MA{\Dlx{}} = 
        \frac
            {\sum_i  \abs{\Delta \lossx{i}}}
            {\abstt{\dataset}}
\end{equation}
Importantly, we can express the absolute change in loss $\abstt{\DL}$ in terms of these two quantities:
\begin{equation}\label{eqn.di_vs_ma}
    \abs{\DL} 
        = \frac
            {\abs{\sum_i \Delta \lossx{i}}}
            {\abs{\dataset}}
        = \MA{\Dlx{}}(1 - \DI{\Dlx{}}) 
\end{equation}
If loss is monotonically decreasing, $\abstt{\DL}$ corresponds to overall loss improvements, such that we can effectively quantify and disentangle the relative contributions of increasing $\DI{\Dlx{}}$ from decreasing $\MA{\Dlx{}}$ in loss deceleration.

\begin{figure}[t!]{
     \centering
     \includegraphics[width=\linewidth]{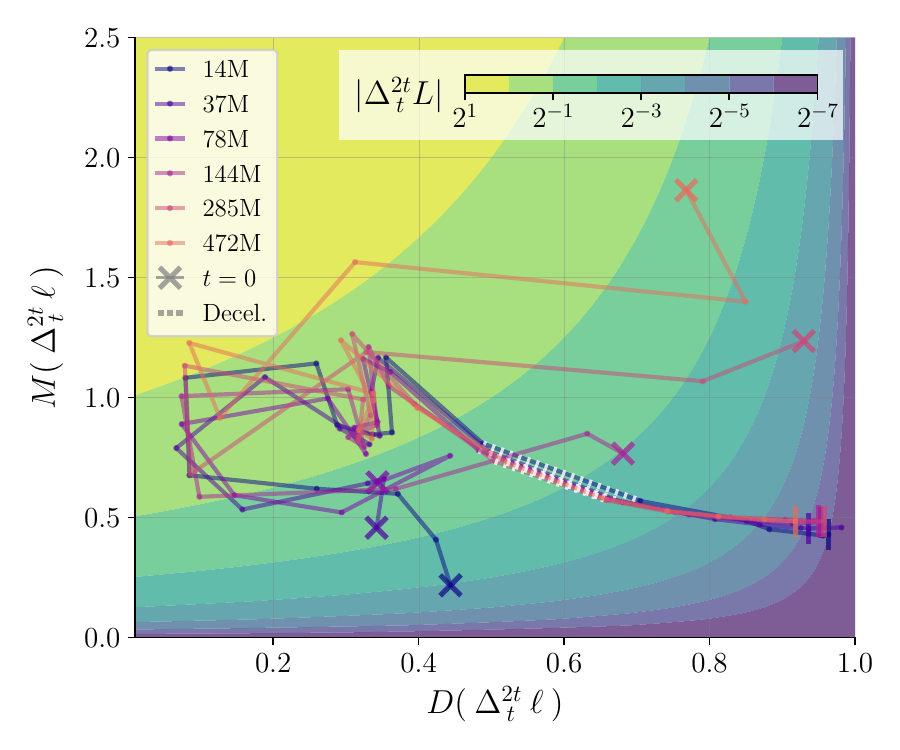}
    \caption{
        Disentangling the relative contributions of increased ZSL ($\DI{\Dlx{}}$) and decreased token-level loss improvements ($\MA{\Dlx{}}$) towards decreased overall loss improvements ($\abstt{\DL}$). 
        Model training trajectories, plotted with respect to these values, show that ZSL dominates decreases in $\abstt{\DL}$ after deceleration.
    }\label{fig.zsl.multistep.d_vs_m}
    }
\end{figure}

\paragraph{Showing ZSL is responsible for deceleration.} ~\\
In \reffig{fig.zsl.multistep.d_vs_m}, we plot model training trajectories with respect to the terms in \refeqn{eqn.di_vs_ma}.
This allows us to visually determine and quantify how increases in $\DI{\Dlx{}}$ map to decreases in $\abstt{\DL}$; i.e. the contribution of ZSL to loss deceleration. 
Notably, we see that during and after deceleration, reductions in $\abstt{\DL}$ are largely attributable to changes in $D$ rather than $M$. Concretely, we know from \refeqn{eqn.di_vs_ma} that the observed reduction in $M$ during deceleration, from $0.75$ to $0.5$, corresponds to a $1.5$x reduction in $\abstt{\DL}$. In contrast, the increase in $D$ observed in that same period, from $0.5$ to $0.95$, corresponds to a $10$x reduction in loss improvements. 

More generally, we see that as $D$ increases and approaches $1.0$, the required increase in $M$ to maintain $\abstt{\DL}$ explodes such that ZSL effectively bottlenecks loss improvements and leads to deceleration. 
These results corroborate that, while decreasing magnitude across per-token loss improvements plays a role in deceleration, the effect of ZSL is almost an order of magnitude greater and effectively bottlenecks loss improvements.

\subsection{The Role of Gradient Opposition in ZSL}\label{sec.expl.grads}
Implicit to our framing of ZSL is the idea that destructive interference in loss improvements is a result of destructive interference in gradients. In this section, we make this assumption explicit and show how systematic gradient opposition, where $\vDI\pp{\dldwx{}} \to \mathbf{1}$, fundamentally leads to ZSL under first-order training dynamics. We will then verify the validity of the first-order training dynamics assumption empirically. Lastly, we rule out an alternate explanation based on progressive sharpening.

\parnl{Under first-order training dynamics}
If weight updates $\Dw$ are sufficiently small, first-order training dynamics apply where changes in overall or per-token losses are approximable via first-order Taylor expansion: 
\begin{align}\label{eqn.c2.fote}
    \DLfote 
        &= \dotp{\Dw}{\dLdw} 
        = \tsum_i {\Dlxfote{i}} \: / \: {\abstt{\dataset}}
    \\
    \dLdw 
        &= \tsum_i {\dldwx{i}} \: / \: {\abstt{\dataset}}
    \notag 
    \\
    \Dlxfote{i} 
        &= \dotp{\Dw}{\dldwx{i}}
    \notag
\end{align}

\noindent
In such cases, ZSL is intrinsically a result of destructive interference in $\Dlxfote{}$ across tokens $i$:

\begin{align}\label{eqn.c2.fote_ldi}
    \DI{\Dlxfote{}} 
        &= 1 - 
        \frac
            {\abs{\tsum_i \dotp{\Dw}{\dldwx{i}}}}
            {\tsum_i \abs{\dotp{\Dw}{\dldwx{i}}}}
        \\[2ex]
        &= 1 - 
        \frac
            {\abs{\dotp{\Dw}{\dLdw}}}
            {\frac{1}{\abs{\dataset}} \tsum_i \abs{\dotp{\Dw}{\dldwx{i}}}}
        \notag
\end{align}

\begin{figure}[t]{
     \centering
     \includegraphics[width=\linewidth]{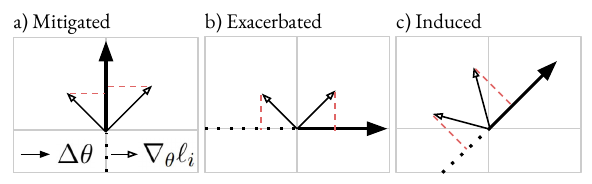}
    \caption{
    $\Dlxfote{i}$ is a projection of $\dldwx{i}$ onto $\Dw$, such that the effect of gradient destructive interference on ZSL can be \textbf{a) mitigated} or \textbf{b) exacerbated}  depending on the alignment between $\vDI\pp{\dldwx{}}$ and $\Dw$. 
    Moreover, destructive interference in loss improvements $\DI{\Dlxfote{}}$ can be \textbf{c) induced} even when gradients are not opposed, but their difference is aligned with $\Dw$.
    }\label{fig.update_effect_on_sgo}
    }
\end{figure}

Notably, $\DI{\Dlxfote{}}$ is a function of $\Dw$ as well as $\dldwx{i}$ and is not necessarily proportional to gradient destructive interference. 
In particular, we see in \reffig{fig.update_effect_on_sgo} that the effect of gradient destructive interference can be mitigated or exacerbated depending on its alignment with $\Dw$. 
In some cases, $\DI{\Dlxfote{}}$  can be "induced" without destructive interference in $\dldwx{}$ if $\Dw$ is aligned with differences in $\dldwx{}$.

\begin{figure}[t]{
     \centering
     \includegraphics[width=\linewidth]{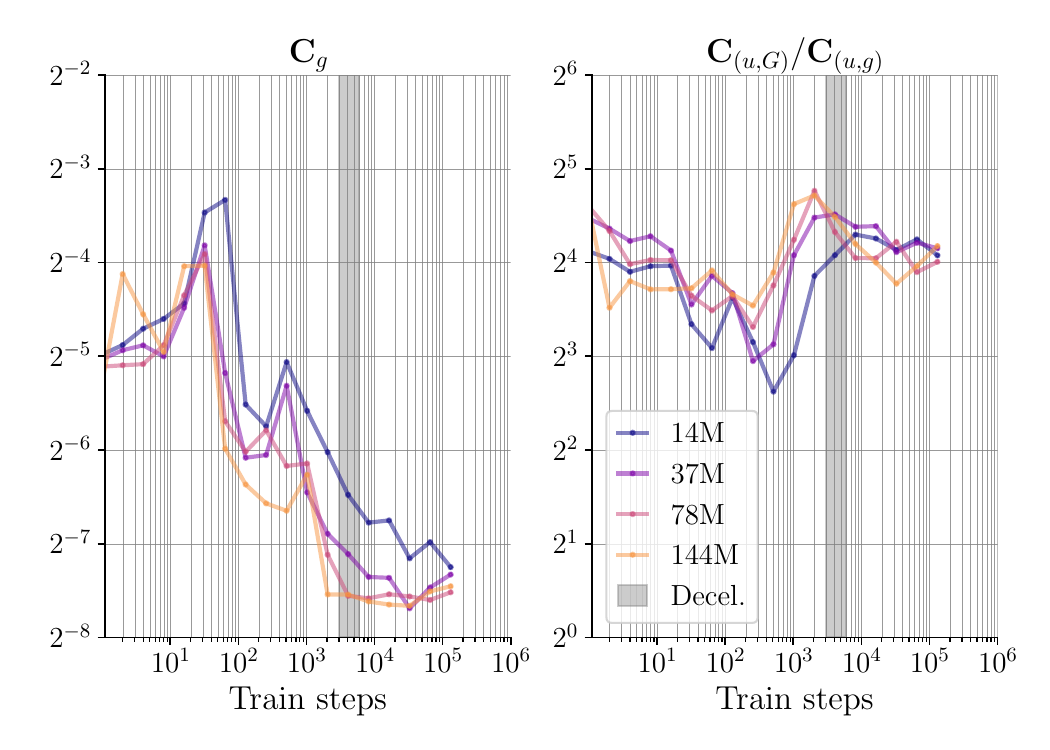}
    \caption{
        $\cg$ approaches $0$ leading up to and during deceleration, while $\cug / \cu$ increases and remains relatively constant. As a result, we know from \refeqn{eqn.cucg} that $\cg$ is the main contributor to $\DI{\Dlxfote{}}$.
    }\label{fig.cucg}
    }
\end{figure}

In light of this, we attempt to disentangle the contribution of gradient opposition to $\DI{\Dlxfote{}}$ in \refapp{app.cucg}, decomposing \refeqn{eqn.c2.fote_ldi} into interpretable components:

\begin{equation}\label{eqn.cucg}
    \DI{\Dlxfote{}} 
        = 1 - \cg \cdot \frac{\cug}{\cu}
\end{equation}

\noindent
Each component captures "constructive" interference attributable to weight updates, per-example gradients and overall gradients (respectively denoted here as $u$, $g$ and $G$ for compactness). 
Constructive interference is simply one minus destructive interference. 
$\cg \in [0,1]$ measures constructive interference in per-example gradients, taking into account the exacerbating or mitigating effects of $\Dw$. 
In contrast, $\cu,\cug \in [0,1]$ capture constructive interference that is induced by projecting $\dldwx{i}$ and $\dLdw$ onto $\Dw$ independent of gradient opposition.

\clearpage
More specifically, $\cg$ measures a weighted average of constructive interference in per-example gradients, across coordinates $j$: 
\begin{align}
\cg &= \tsum_j W_j \cdot (1 - \vDI\pp{\dldwx{}}_{[j]})
\\ \notag
W_j &\propto \tsum_i \abstt{\Dw_{[j]} \cdot \dldwx{i,[j]}}
\end{align}
where $\tsum_j W_j = 1$ and $W_j$ captures the mitigating or exacerbating effect of $\Dw$ on coordinate $j$. 
Notably, $\cg \leq \max(\mathbf{1} - \vDI\pp{\dldwx{}})$ by convexity, such that systematic gradient opposition where $\vDI\pp{\dldwx{}} \to \mathbf{1}$ implies $\cg \to 0$. We validate this empirically on a subset of models in \reffig{fig.cucg} and confirm that $\DI{\Dlxfote{}} \to 1$ is indeed primarily attributable to $\cg \to 0$; i.e. that destructive interference in loss improvements under first-order training dynamics is primarily attributable to destructive interference in gradients.

\parnl{Testing the first-order dynamics assumption.}
Our hypothesis relies on the assumption that $\Dlxfote{}$ is a valid approximation of $\Dlx{}$ such that destructive interference in $\Dlxfote{}$ is reflective of destructive interference in $\Dlx{}$. To validate our hypothesis, we must therefore validate this assumption.
However, computing $\dldwx{i}$ and the corresponding $\Dlxfote{i} = \dotp{\Dw}{\dldwx{i}}$ for each token is intractable. 

\begin{figure}[t]
    \centering    
    \includegraphics[width=0.9\linewidth]{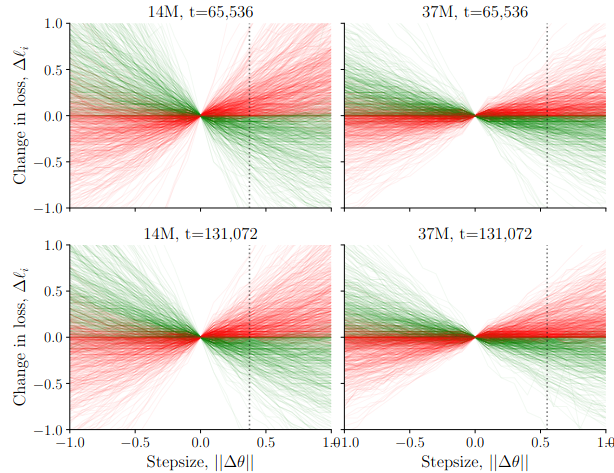}
    \caption{Per-token loss landscapes at step $t$ along $\Delta_t \theta$.
    Dashed vertical lines indicate $\theta_{t+1} = \theta_t + \Delta_t \theta$. 
    Tokens which improve in loss after the update are indicated in green, and tokens which degrade are indicated in red
    }
    \label{fig.landscape_xsection.sample}
\end{figure}

Instead, to empirically measure  $\Dlxfote{}$, we compute $1$D cross-sections of per-token loss landscapes by evaluating model checkpoints along increments of their next weight update $\Dw$, with $\theta(\alpha) = \theta + \alpha \Dw / \norm{\Dw} , \: \alpha \in [-10, 10]$. This allows us to tractably measure $\Dlxfote{}$ as a linearization around $\alpha=0$ where $\Dlxfote{}(\alpha) = \alpha \pp{\frac{\lossx{\theta + \epsilon} - \lossx{\theta}}{\norm{\epsilon}}}$. 
A sample of \num{1000} such per-token loss landscapes is shown in \reffig{fig.landscape_xsection.sample}, with the complete set in \refapp{app.res.xsections}. 
Generally, these appear linear in the vicinity of weight updates, suggesting that actual changes in per-token losses $\Dlx{}$ are well captured by their first-order approximation $\Dlxfote{}$. 

\begin{figure}[t]
    \centering    
    \includegraphics[width=\linewidth]{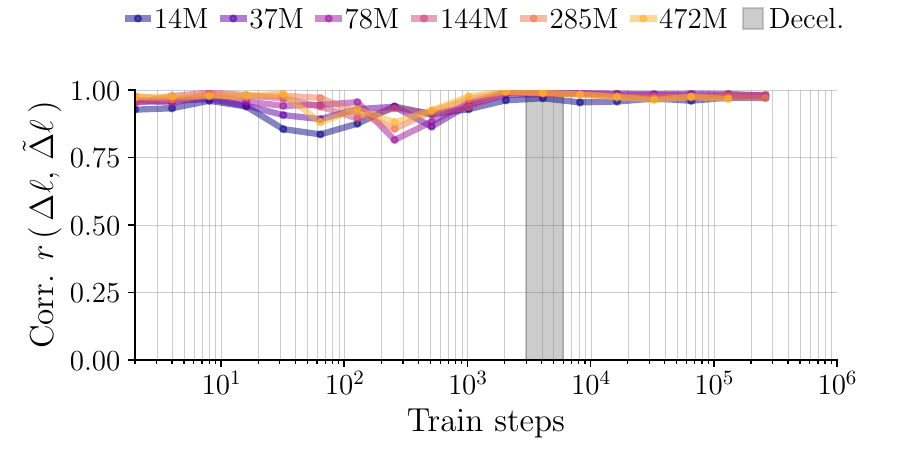}
    \caption{Correlation between $\Dlx{}$ and it's first-order approximation $\Dlxfote{}$ is close to $1.0$ after deceleration.}
    \label{fig.c2.fote_dl_corr}
\end{figure}

However, to more quantifiably verify that this indeed is the case, we measure and plot the Pearson correlation coefficient between  $\Dlx{}$ and $\Dlxfote{}$ throughout training in \reffig{fig.c2.fote_dl_corr}. 
We find strong correlation after deceleration where we observe destructive interference in loss improvements and gradients, validating our hypothesis by validating the underlying assumption of first-order dynamics on which our reasoning depends.

\begin{figure}[t]
    \centering    
    \includegraphics[width=0.9\linewidth]{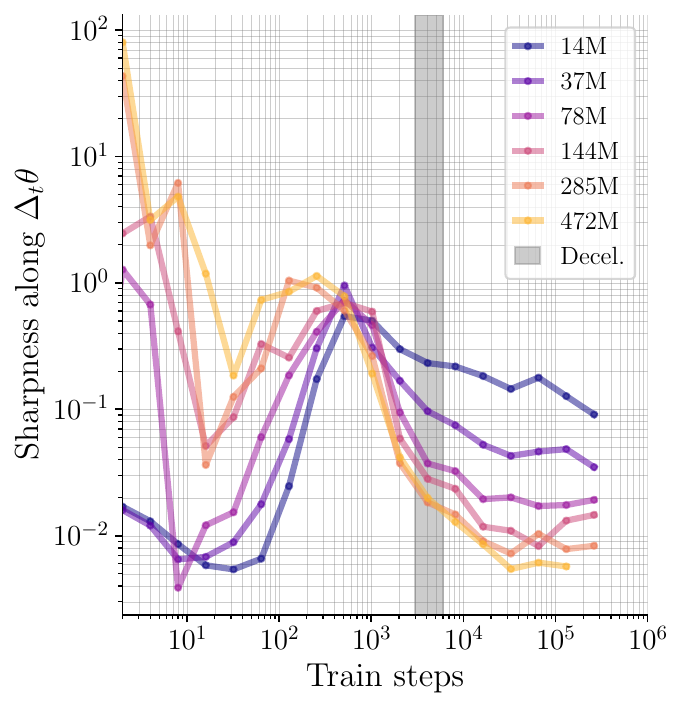}
    \caption{Sharpness decreases with loss deceleration.}
    \label{fig.c2.sharpness}
\end{figure}

\clearpage
\paragraph{Ruling out the role of progressive sharpening.}~\\
As an alternative explanation for ZSL, one might consider progressive sharpening \cite{cohen_gradient_2022, rosenfeld_outliers_2023} where $\Dw$ might overshoot local minima for some examples but not others. 
Surprisingly, and perhaps counter to conventional wisdom, we observe in \refapp{app.res.landscapes} that loss landscapes instead become significantly flatter with deceleration; following an initial phase of high sharpness before deceleration. 

To quantify this observation, we measure the sharpness of loss landscapes along update directions. Specifically, we fit a quadratic to the loss landscape cross-section, using the second order term as a measure of sharpness. In \reffig{fig.c2.sharpness}, we see the same trend, with sharpness peaking and immediately begin decreasing before deceleration. 
While the relationship between loss sharpness and zero-sum learning is outside the scope of this work, it appears there might be an interesting connection.

\section{Explaining Scaling Improvements}
\label{sec.impr}
In \refsec{sec.decel} we showed how scaling improves loss by mitigating loss deceleration, specifically by decreasing the loss $L_d$ and step $t_d$ at which it occurs, and increasing the subsequent log-log rate of loss improvement $r_d$ (\reftab{tab.decel_measurements}). 
Conversely, in \refsec{sec.expl} we proposed a mechanistic explanation of loss deceleration based on interactions at the levels of per-example loss improvements, and of per-example gradients. 
Specifically, we showed that loss deceleration is a transition in training dynamics characterized by the emergence of near-complete destructive interference in per-example gradients and loss improvements; i.e. ZSL. 

In this section, we will attempt to connect these findings, and shed light on \textit{how} scaling mitigates deceleration based on the underlying mechanisms we identified. Specifically, we will focus on understanding how scale improves $L_d$ and $r_d$. While \reftab{tab.decel_measurements} also suggests scale decreases $t_d$, this effect is not as consistent, and is negligible relative to the total number of training steps.

\paragraph{Decomposing loss improvements.}~\\
Similar to \refsec{sec.expl.grads}, we decompose the first-order Taylor expansion for changes in loss from \refeqn{eqn.c2.fote} into interpretable components that enable a finer-grained analysis of training dynamics, specifically the cosine similarity and $L^2$ norms of weight updates $\Dw$ and gradients $\dLdw$:
\begin{align}
    \DLfote 
        &= \sqnorm{\Dw} \sqnorm{\dLdw} \: \cossim{\Dw}{\dLdw}  
    \label{eqn.dl_decomposition}
\end{align}
We show these values across training steps and model scales in \reffig{fig.dl_decomposition}, and will discuss their interpretation in the following sections.

\subsection{Improving Loss Before Deceleration ($L_d$)}
\label{sec.impr.ld}
Surprisingly, we find in \reftab{tab.early_impr} that most of the scaling improvements in loss at deceleration $L_d$ are already established by step $t=32$. 
\begin{figure}[t]
     \centering
     \captionsetup{width=\linewidth}
     \includegraphics[width=\linewidth]{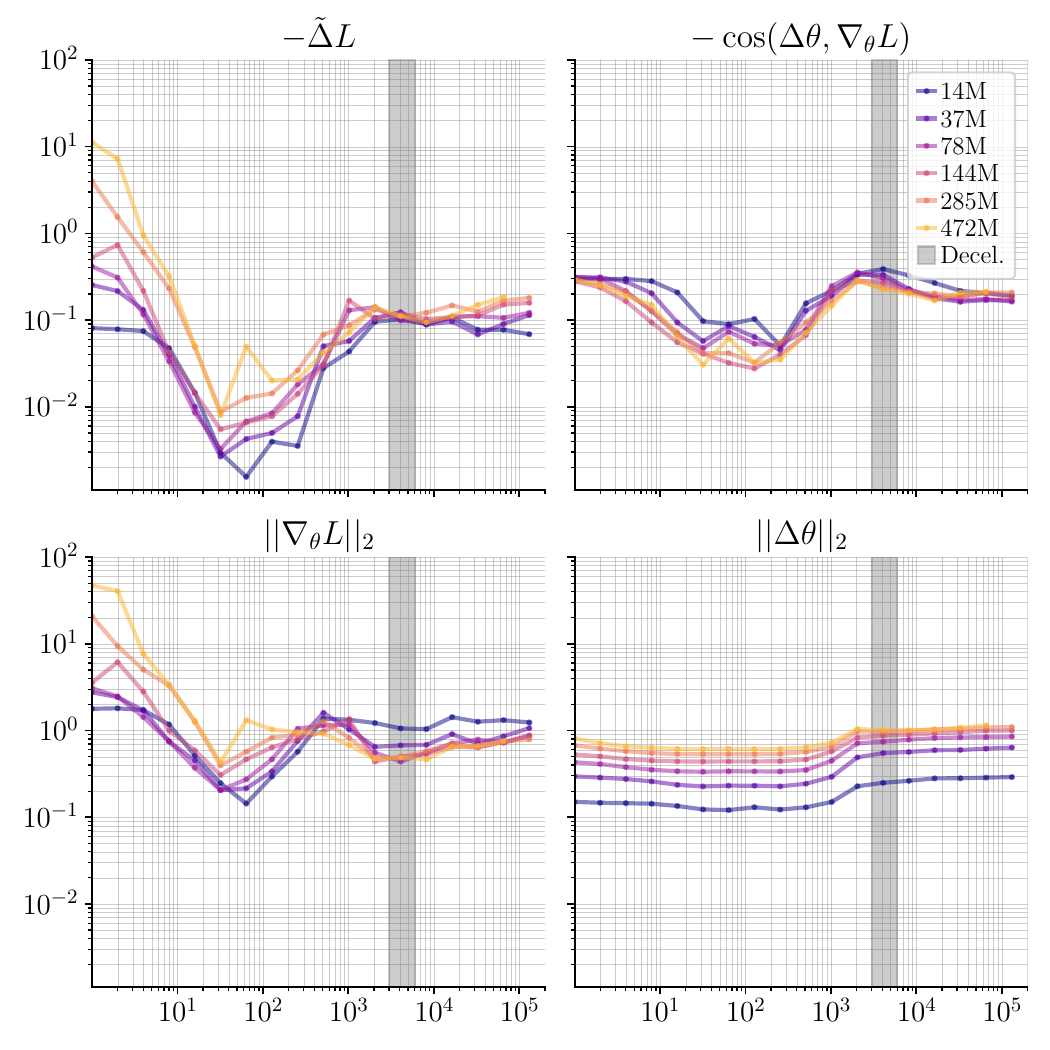}
    \caption{
        First-order approximation of loss improvements with terms from \refeqn{eqn.dl_decomposition} plotted throughout training. Note that because
        $\log(\Tilde\Delta L)$ is a sum of the log of its terms, the shared log-scale allows us to easily gauge how different terms contribute to changes in $\Tilde\Delta L$.
    }
    \label{fig.dl_decomposition}
\end{figure}

\begin{table}[t]
  \caption{Scaling improvements in loss at deceleration $L_d$ are established early during training.}
  \label{tab.early_impr}
  \centering
  \resizebox{\columnwidth}{!}{
    \begin{tabular}{lccccc}
    \toprule
    Loss Improvement && $t=32$ && $t=4096$ & $t=8192$ \\
    \midrule
    $14\text{M} \to 37\text{M}$   && $0.76$  && $0.43$ & $0.45$ \\
    $37\text{M} \to 78\text{M}$   && $0.29$  && $0.20$ & $0.21$ \\
    $78\text{M} \to 144\text{M}$  && $0.15$  && $0.12$ & $0.12$ \\
    $144\text{M} \to 285\text{M}$ && $0.15$  && $0.11$ & $0.11$ \\
    $285\text{M} \to 472\text{M}$ && $0.05$  && $0.06$ & $0.07$ \\
    \bottomrule
  \end{tabular}
  }
\end{table}
From \refeqn{eqn.dl_decomposition} and \reffig{fig.dl_decomposition}, the underlying reason becomes apparent. Scaling models improves $\DLfote$ primarily by improving gradient norms $\sqnorm{\dLdw}$ in the beginning of training. Beyond $t=32$, the effects of scaling become less significant, with improvements in $\DLfote$ and $\sqnorm{\dLdw}$ orders of magnitude smaller and eventually reversed leading up to deceleration. In contrast, scaling degrades gradient-update alignment $-\cossim{\Dw}{\dLdw}$, and results in consistent but relatively insignificant improvements in $\sqnorm{\Dw}$. These effects are trivially explained by an increased number of parameters and appear unrelated to deceleration, however it remains an open question how similar effects can be achieved independent of scale.

\subsection{Improving Loss After Deceleration ($r_d$)}
\label{sec.impr.rd}
We see in \reffig{fig.zsl.zsl_multistep_log2_DI} that post-deceleration ZSL is mitigated by scaling model size, which we know results in greater loss improvements from \refeqn{eqn.di_vs_ma} that can explain how scaling improves $r_d$. Unfortunately, the way in which scaling reduces ZSL after deceleration is not as immediately obvious. 

\begin{figure}[t]
     \centering 
    \includegraphics[width=\linewidth]{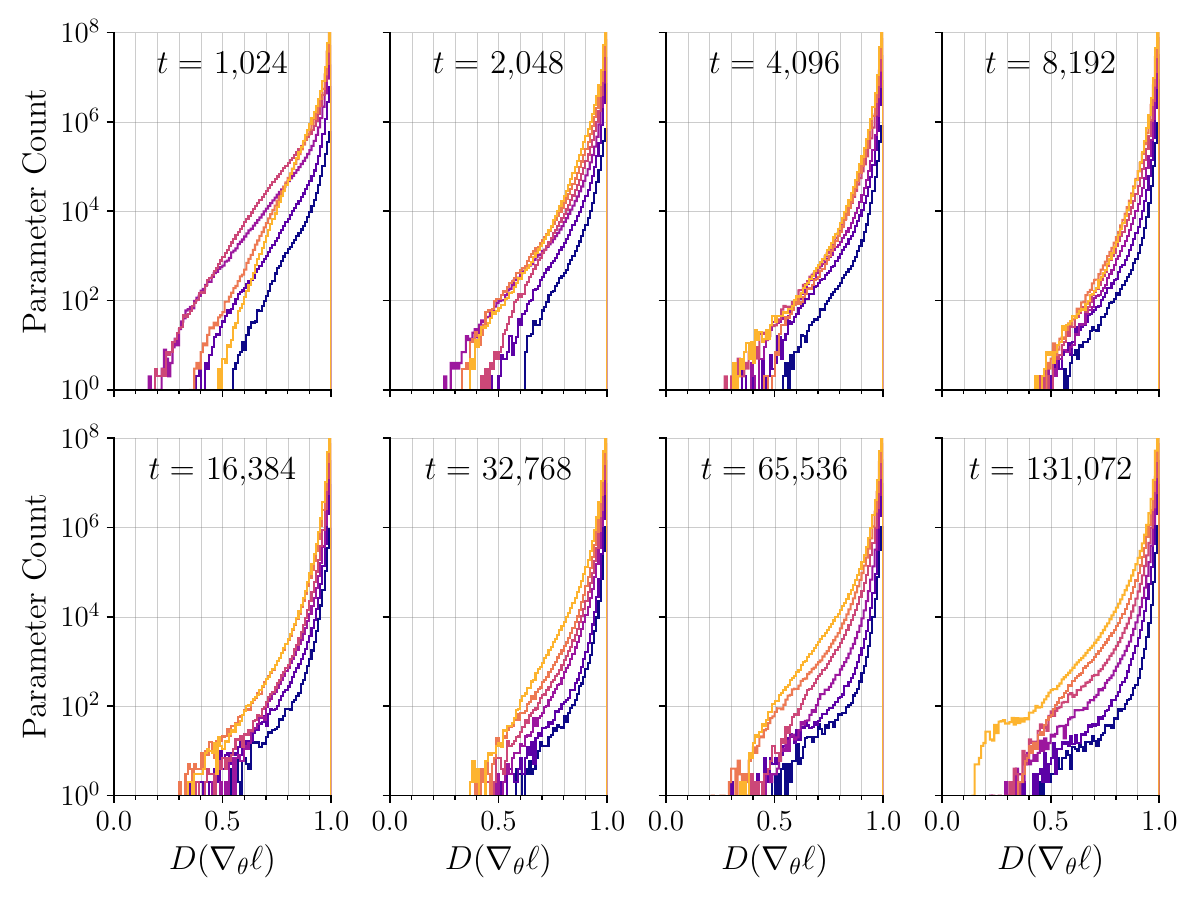}
    \caption{
       Histograms of gradient destructive interference. Deceleration happens between steps 4096 and 8192. Note that model size legend is consistent with other plots but omitted for space.
    }
    \label{fig.d_histo}
\end{figure}

We see in \reffig{fig.sgo.di_grads_avg} that gradient destructive interference (averaged across parameters) actually becomes more pronounced with larger models. However, $99\%$ destructive interference in a $14$M-dimensional gradient does not have the same effect as in a $144$M-dimensional gradient. In particular, the latter will have more degrees of freedom along which shared gradient directions can exist between tokens. Indeed, we find in \reffig{fig.d_histo} that, especially after deceleration, larger models have more parameters with lower destructive interference. 
This can explain why larger models have lower ZSL after deceleration, and thus improved $r_d$.

\section{Conclusion and outlook}
In this work we proposed and validated a mechanistic explanation of scaling laws grounded in training dynamics. 
Specifically, we identified loss deceleration as a novel transition in training dynamics that can explain scaling improvements in quantifiable but interpretable terms, such that an explanation of deceleration becomes an explanation of scaling improvements. 
To this end, we proposed zero-sum learning as the mechanism underlying deceleration, validating these against alternate hypotheses with empirical and theoretical analyses. 
Lastly, we revisit scaling improvements from the perspective of these mechanisms and show how scaling improves loss by mitigating zero-sum learning and more specifically, systematic gradient opposition. 

Our findings suggest that these could potentially be mitigated directly to improve loss independent of scale, laying a foundation for future research.  
Furthermore, studying per-example gradient dynamics as in our work is an under-explored area of research that can shed new light on learning dynamics, scaling, and generalization more broadly.

\section*{Limitations}
\paragraph{Comprehensiveness of experimental settings.}
Scaling laws are a general phenomenon observed across tasks, model architectures, parameters, and evaluation measures. However, this work only considers the scaling of cross-entropy loss with model size in transformer-based language models on typical webscale text. 
Despite replicating our experiments across several ablations in \refapp{app.ablation}, we do not generalize our findings to different settings beyond language modeling. While this lies beyond the scope of our original research question and the prior works on which we build, verifying how our findings generalize across different settings is an important area of future work.

\paragraph{Accounting for gradient opposition in both $\MA{\Dlx{}}$ and $\DI{\Dlx{}}$.}
In \refsec{sec.expl.attrib} and \refeqn{eqn.di_vs_ma} we showed that destructive interference in loss improvements $\DI{\Dlx{}}$ is primarily responsible for deceleration. In contrast, decreases in average per-token loss improvements $\MA{\Dlx{}}$ played an non-negligible but less significant role. However, our analysis of gradient opposition (\refsec{sec.expl.grads}) only considers its effect on $\DI{\Dlx{}}$, while it likely also has an effect on $\MA{\Dlx{}}$ via its effect on optimizer steps $\Dw$. 
However, these effects are likely highly dependent on the optimizer and its configuration, and likely not generalizable in the scope of our research question; hence why we chose to abstract away $\Dw$ in our analysis. Nevertheless, this a salient gap in our analysis that should be further explored. 

\paragraph{Reconciling single step and multi step training dynamics.} 
The connection between the behaviour of gradients and loss can be made more precise. In particular, our gradient analysis only reflects single-step training dynamics, while ZSL and loss improvements appear to depend on interactions across multiple optimization steps (see \refapp{app.zsl_incr_steps}). Understanding the effect of multi step interaction is a natural next step for this research. 

\paragraph{Negative societal impacts or ethical concerns.} 
Our work focuses on understanding existing and well-established methods, and does not meaningfully contribute to any negative societal impacts or ethical concerns beyond what is typically associated with language modeling research. In principle, by focusing our analysis on a single metric (cross-entropy loss), this could lead to over-optimizing that metric at the expense of other real-world concerns. While this work is at too early a stage for this to pose a meaningful risk, it is important to keep in mind as a limitation in interpreting our findings and building new methods on top of them. 

\section*{Acknowledgments}
We would like to acknowledge support from IVADO and the Canada First Research Excellence Fund [E.L.]; from the Canada CIFAR AI Chair Program and the Canada Excellence Research Chairs Program [I.R.]; and from the NSERC post-graduate doctoral (PGS D) scholarship as well as the FRQNT Doctoral (B2X) doctoral scholarship [A.M.]. We would also like to thank
Mila (\url{mila.quebec}) and its IDT team for providing and supporting the computing resources used in this work; as well as the Capital One AGI Foundations Team and in particular Ashwinee Panda for the helpful discussions and feedback.

\clearpage\bibliography{zotero}
\clearpage\appendix
\clearpage
\onecolumn
\section{Methodology}
\label{app.meth}
%

\subsection{Language model pretraining}
For our experiments, we adapt the {OLMo} codebase (licensed under Apache-2.0) and 
train variants of {OLMo} with the publicly available training dataset of {OLMo}-7B-0724  \cite{groeneveld_olmo_2024}. Model dimensions and learning rates are based on \cite{kaplan_scaling_2020} and shown in \reftab{tab:app.model.dims}, labeled with (rounded) total parameter counts. For pretraining, we again adapt the experimental setup of \cite{kaplan_scaling_2020}, training with a batch size of $0.5$M tokens for $2^{18}$ steps. However, instead of a cosine learning rate decay, we adopt the trapezoidal learning rate schedule from \cite{hagele_scaling_2024} with a learning rate warmup to the values in \reftab{tab:app.model.dims} in the first $2\small{,}000$ steps and no cooldown in the $2^{18}$ steps considered. Note that the OLMo-1B and OLMo-7B models are those trained by \cite{groeneveld_olmo_2024} and could not included in our analysis of ZSL because of insufficient checkpointing frequency before deceleration. 

\paragraph{Code and artefacts} Model and optimizer checkpoints, and logs across training are available at \url{https://github.com/mirandrom/zsl} under an Apache-2.0 license to enable future research in this direction. In our experiments, we used a variety of computational resources which are recorded in the logs we make available. Generally, we performed distributed training 4-32 L40 GPUs or 4 H100 GPUs, with smaller models pretraining requiring on the order of 10 GPU hours, and the largest 472M requiring on the order of 1000 GPU hours. 

\paragraph{Language model analyses}
During training, we checkpoint the model and optimizer every $2^i$ steps with $i \in [0,18]$. Our analyses of ZSL and gradient opposition are done on these checkpoints after pretraining. Methodological details regarding e.g. precision or batch size are kept consistent with pretraining to obtain representative results. All of our evaluations are conducted on the \textsc{C4} validation set from \cite{magnusson_paloma_2023}, using the tokenizer from \cite{groeneveld_olmo_2024}, consistent with pretraining.

\begin{table*}[h]
\caption{Model and Optimizer Parameters for Different Runs}
\label{tab:app.model.dims}
\centering
\begin{tabular}{lccccccccc}
\toprule
{Model size}                 
        & \textbf{14M} 
        & \textbf{37M} 
        & \textbf{78M} 
        & \textbf{144M} 
        & \textbf{285M} 
        & \textbf{472M} 
        &
        & \textbf{OLMo-1B} 
        & \textbf{OLMo-7B} 
        \\ 
\midrule
{\texttt{d\_model}}    
        & 256          
        & 512          
        & 768          
        & 1024
        & 1536
        & 2048
        &
        & 2048
        & 4096
        \\ 
{\texttt{mlp\_dim}}    
        & 256          
        & 512          
        & 768          
        & 1024
        & 1536
        & 2048
        &
        & 16384
        & 22016
        \\ 
{\texttt{n\_heads}}    
        & 4            
        & 8            
        & 12           
        & 16
        & 16
        & 16  
        &
        & 16
        & 32
        \\ 
{\texttt{n\_layers}}    
        & 4            
        & 8            
        & 12           
        & 16            
        & 16            
        & 16          
        &
        & 16
        & 32
        \\
{\texttt{peak\_lr}}    
        & 1.3E-3       
        & 9.7E-4       
        & 8.0E-4       
        & 6.8E-4   
        & 5.7E-4   
        & 4.9E-4   
        &
        & 4.0E-4
        & 3.0E-4
        \\ 
{\texttt{warmup}}    
        & \num{2000}       
        & \num{2000}       
        & \num{2000}       
        & \num{2000}       
        & \num{2000}
        & \num{2000}       
        &
        & \num{2000}       
        & \num{5000}       
        \\ 
\bottomrule
\end{tabular}
\end{table*}

\clearpage
\subsection{Additional Details on Fitting BNSL}
\label{app.bnsl}
\paragraph{Fitting}

We adapt the methodology for fitting \refeqn{eqn.bnsl} published by \citet{caballero_broken_2023} at \url{https://github.com/ethancaballero/broken_neural_scaling_laws}. 
We include the code implementation below. Empirically, we had to implement the following changes to improve stability: 
\begin{itemize}
    \item Assume $a = 0$ and remove it from the optimization procedure. 
    \item Fit the function in log-log space instead of manually scaling $b$ and $d_1$. Note that datapoints sampled uniformly along $x$ will result in a data imbalance when fitting in log-log space; to mitigate this we also subsample datapoints uniformly in log space.
    \item Estimate initial parameters instead of running a bruteforce gridsearch.
\end{itemize}

\begin{lstlisting}[
    language=Python, 
    caption={
        Code for fitting one-break BNSL.
    }
]

    import numpy as np
    import scipy

    def log_1b_bnsl(xlog, b, c0, c1, d1log, f1):
        ylog_pred = np.log(b) - c0*xlog - (c1*f1)*np.log(1+np.exp((xlog-d1log)/f1))
        return ylog_pred
    
    def fit_1b_bnsl(x: np.ndarray, y: np.ndarray, d1_est: float = 6000):
        # initialize parameters with reasonable values (for stability)
        d1log = np.log(d1_est)
        xlog = np.log(x)
        ylog = np.log(y)
        d1_idx = np.argmin(np.abs(xlog - d1log))
        c0 = -np.mean((ylog[0:d1_idx] - ylog[1:d1_idx+1]) \
                / (xlog[0:d1_idx] - xlog[1:d1_idx+1]))
        c1 = -np.mean((ylog[d1_idx:-2] - ylog[d1_idx+1:-1]) \
                / (xlog[d1_idx:-2] - xlog[d1_idx+1:-1]))
        c1 = c1 - c0
        b = ylog[0] + c0*xlog[0]
        
        # fit parameters with scipy
        p0 = [b, c0, c1, d1log, 0.3]
        popt, pcov = scipy.optimize.curve_fit(
            log_1b_bnsl,
            xlog, ylog, 
            p0=p0,
            method='dogbox',
        )
    
        return popt, pcov
        
\end{lstlisting}

\begin{figure*}[h]
     \centering 
    \includegraphics[width=0.8\linewidth]{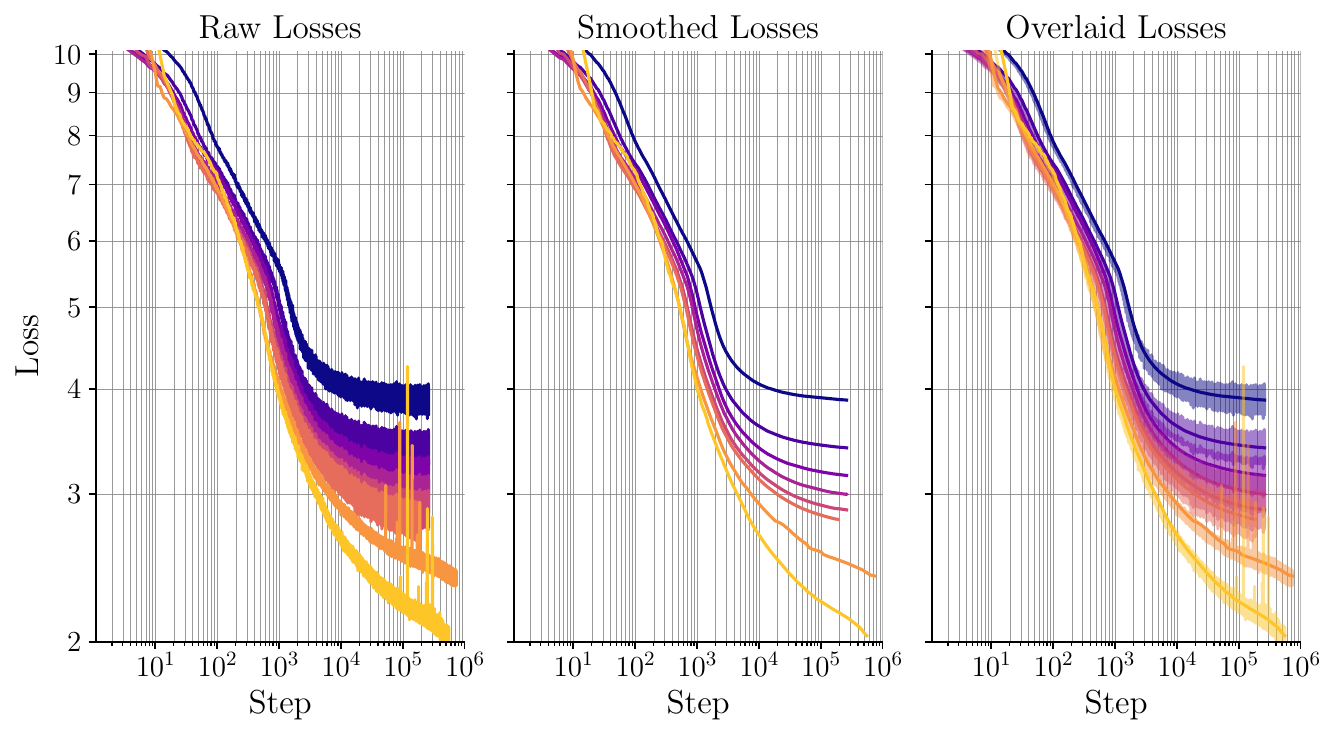}
    \vspace{-1em}
    \caption{
        $\operatorname{LMSA}_k\pp{L_t}$ smoothing with $k=1.2$.
    }
    \label{fig.app.loss_smoothing}
\end{figure*}

\clearpage
\paragraph{Smoothing}
The loss curves we fit are batch losses logged at every step during training. 
Because training is single-epoch, i.e. online, these losses are effectively a noisy measurement of the true validation loss. 
However, we found that this noise (characterized by oscillations in loss at too-small timescales) leads to severe instability with the original methodology published by \cite{caballero_broken_2023}. 
To smooth these curves, we use $\operatorname{LSMA}_k$, a logarithmic variant of the simple moving average that we found to work well for fitting noisy log-log loss curves with high fidelity. Notably, $\operatorname{LSMA}$ naturally handles the increasing timescales at which loss oscillations occur as number of training steps increase. 
We found $k=1.2$ to work sufficiently well as shown in \reffig{fig.app.loss_smoothing}.

\begin{equation}
    \operatorname{LSMA}_k\pp{L_t} = \frac{1}{t - p(t)} \sum_{s=p(t)}^{t} L_s
    \quad,\quad\quad
    p(t) = \operatorname{floor}(t/k)
    \notag
\end{equation}

\paragraph{Results and validation} 

We report the resulting parameters and error measurements from fitting \refeqn{eqn.bnsl} in \reftab{tab.app.bnsl_params}, finding that parameter standard deviation is typically on the order of $1\%$, while root standard log error (RSLE) is on the order of $0.01$, comparable with values reported by \citet{caballero_broken_2023}. These results suggest that loss deceleration is reliably measurable with BNSL. 

\begin{table}[!h]
  \caption{BNSL parameters and root-standard log error of resulting fit (RSLE).}
  \label{tab.app.bnsl_params}
  \centering
  \resizebox{\columnwidth}{!}{
    \begin{tabular}{lcccccc}
    \toprule
    Model & $b$ & $c_0$ & $c_1$ & $\log(d_1)$ & $f_1$ & RSLE  \\
    \midrule
    14M & $18.42 \pm 0.16$ & $0.17 \pm 0.00$ & $-0.16 \pm 0.00$ & $8.68 \pm 0.02$ & $0.20 \pm 0.03$ & $0.011$ \\ 
    37M & $19.64 \pm 0.23$ & $0.20 \pm 0.00$ & $-0.18 \pm 0.00$ & $8.68 \pm 0.03$ & $0.24 \pm 0.03$ & $0.015$ \\ 
    78M & $20.66 \pm 0.25$ & $0.21 \pm 0.00$ & $-0.19 \pm 0.00$ & $8.69 \pm 0.03$ & $0.29 \pm 0.03$ & $0.014$ \\ 
    144M & $20.31 \pm 0.26$ & $0.21 \pm 0.00$ & $-0.19 \pm 0.00$ & $8.71 \pm 0.03$ & $0.34 \pm 0.03$ & $0.015$ \\ 
    285M & $20.85 \pm 0.30$ & $0.22 \pm 0.00$ & $-0.20 \pm 0.00$ & $8.57 \pm 0.03$ & $0.44 \pm 0.03$ & $0.013$ \\ 
    472M & $21.16 \pm 0.32$ & $0.23 \pm 0.00$ & $-0.19 \pm 0.00$ & $8.44 \pm 0.03$ & $0.39 \pm 0.04$ & $0.014$ \\ 
    \midrule
    OLMo-1B & $25.97 \pm 0.38$ & $0.27 \pm 0.00$ & $-0.23 \pm 0.00$ & $8.22 \pm 0.03$ & $0.76 \pm 0.02$ & $0.008$ \\ 
    OLMo-7B & $27.49 \pm 0.48$ & $0.28 \pm 0.00$ & $-0.22 \pm 0.00$ & $8.44 \pm 0.04$ & $0.76 \pm 0.03$ & $0.008$ \\    
    \bottomrule
  \end{tabular}
  }
\end{table}

\clearpage
\subsection{Additional Details on Computing Gradient Opposition}
\label{app.measure_ptg}
\parnl{Measuring a tractable proxy for per-token gradient destructive interference}
While computing true per-token gradients is typically intractable, we can tractably compute destructive interference between gradients for each token \textit{features} rather than for each token \textit{loss}. Concretely, for any module $\mathcal{M}(x) = y, \ \mathcal{M} : S \times D_1 \mapsto S \times D_2$ with sequence length $S$ and hidden dimensions $D_1, D_2$; we define $\nabla_{\theta} \lossh{i} = \sum_{\mathcal{M}} (\delta \Loss / \delta y_{i}) (\delta y_{i} / \delta \theta_{\mathcal{M}})$ for the $i^{th}$ token in a sequence.
In the PyTorch code block below, we illustrate how the backward pass of a linear layer with weights $W$ is modified to compute gradient destructive interference across samples 
$\tfrac{\delta L}{\delta W}= \sum_{i} \tfrac{\delta L}{\delta y_{i}} \tfrac{\delta y_{i}}{\delta W}$ 
, similar to \citet{yousefpour_opacus_2021}: 

\begin{lstlisting}[
    language=Python, 
    caption={
        Illustrative example computing gradient destructive interference in PyTorch
    }
]

  import torch
  from torch import nn, autograd, functional as F

  def compute_gdi(W: nn.Parameter):
    gdi = 1 - W.sum_grads.abs()/W.sum_abs_grads
    return gdi.mean()

  class GDILinearFunction(autograd.Function):
    @staticmethod
    def forward(ctx, x, W):
      ctx.save_for_backward(x, W)
      y = F.linear(x,W)
      return y

    @staticmethod
    def backward(ctx, dLdy): 
      x, W = ctx.saved_tensors   
      if ctx.needs_input_grad[1]:

        # instantiate metrics if not present
        if not hasattr(W, 'sum_grads'):
          W.sum_grads = torch.zeros_like(W)
          W.sum_abs_grads = torch.zeros_like(W)

        # accumulate sum of gradients
        W.sum_grads.add_(
          torch.einsum(
            'B...d,B...p->pd', x, dLdy
          )
        )
        # accumulate sum of absolute gradients
        W.sum_abs_grads.add_(
          torch.einsum(
            'B...d,B...p->pd', x.abs(), dLdy.abs()
          )
        )
      # compute and return input gradient for backprop
      if ctx.needs_input_grad[0]:
        dLdx = torch.einsum(
          'B...p,pd->B...d', dLdy, W
        )
      else:
        dLdx = None

      return dLdx, None

\end{lstlisting}

\clearpage
\parnl{Consistency of findings between proxy and true measure of gradient destructive interference}
Computing true per-token gradients is feasible, just intractable in most cases. In particular, it requires doing one backward pass per token in a sequence, effectively resulting in 1000x increase in costs. We were nevertheless able to measure true gradient destructive interference for models up to 144M so as to verify the validity of the results obtained in \reffig{fig.sgo.di_grads_avg} with out proxy measure.  We find that qualitatively, both our proxy and true measure behave consistently, converging towards complete destructive interference during deceleration for non-embedding parameters. However, we find that our proxy systematically \textit{underestimates} gradient destructive interference. Understanding the discrepancies between the proxy and true measures of gradient destructive interference, especially in terms of how they relate to zero-sum learning, remains an open question.

\begin{figure}[h]
     \centering 
    \includegraphics[width=0.4\linewidth]{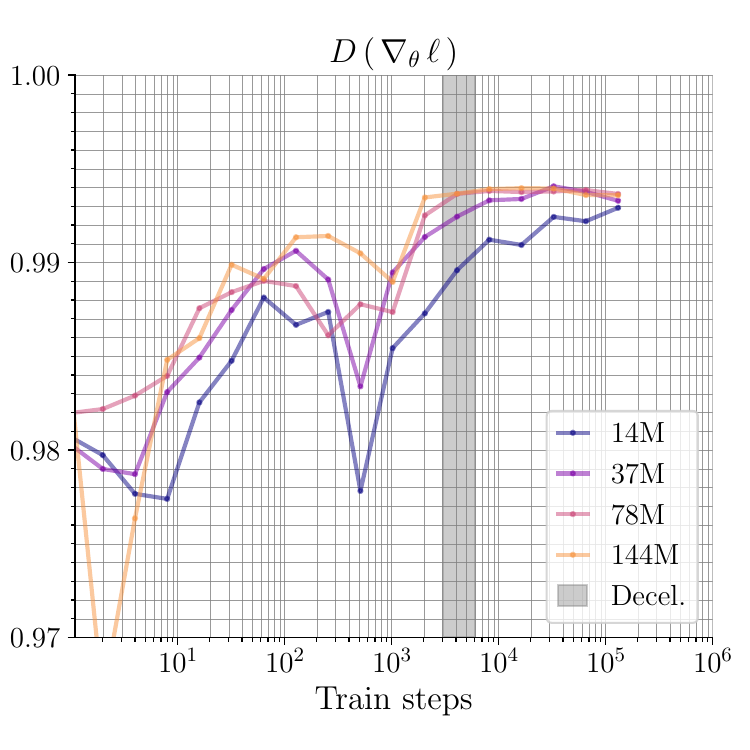}
    \caption{Actual destructive interference in per-example gradient behaves consistently with our proxy in \reffig{fig.sgo.di_grads_avg}, but is systematically underestimated by it.}
    \label{fig.app.gdi_vs_ldi_true}
\end{figure}

\clearpage
\onecolumn
\section{Additional Results (Analyses)}
\label{app.analyses}

\subsection{Decomposing First-Order Training Dynamics}\label{app.cucg}
\parnl{Notation and goal}
Generalizing Eqns. \ref{eqn.di},\ref{eqn.ma},\ref{eqn.di_vs_ma} to any empirical average $X = \tsum_i^N x_i / N$, we have:

\begin{alignat}{3}
    &\text{Destructive interference: }& 
    \zslm{D}{i}{N}{x_i}
    &= 1 - \frac{|\tsum_i^N{x_i}|}{\tsum_i^N |x_i|} 
    \in [0,1]
\label{eqn.zsl.Di}\\[2ex]
    &\text{Constructive interference: }&
    \zslm{C}{i}{N}{x_i}
    &= \frac{|\tsum_i^N{x_i}|}{\tsum_i^N |x_i|} = 1 - \zslm{D}{i}{N}{x_i} 
    \in [0,1]
\label{eqn.zsl.Ci}\\[2ex]
    &\text{Average magnitude: }&
    \zslm{M}{i}{N}{x_i}
    &= \frac{1}{N} \tsum_i^N |x_i|
\label{eqn.zsl.Mi}\\[2ex]
    &\text{Absolute empirical average: }&
    |X|
    &= \zslm{M}{i}{N}{x_i} \: \zslm{C}{i}{N}{x_i}
\label{eqn.zsl.X}
\end{alignat}
\\

\noindent
Our goal is to understand and quantify how destructive interference between per-example loss improvements $\zslm{D}{i}{N}{\Dlx{i}}$ (\refeqn{eqn.zsl.D.Dl})
is affected by gradient opposition 
(i.e. destructive interference between per-example gradients $\zslm{D}{i}{N}{\vg_i}$, \refeqn{eqn.zsl.D.g_vec}). 
Recall that the overall change in loss for examples $i \in [1 \ldots N]$ is $\DL = \tfrac{1}{N}\tsum_i^{N} \Dlx{i}$. 
Similarly, the overall gradient is $\dLdw = \tfrac{1}{N} \tsum_i \dldwx{i} \in \R{M}$ where $M$ is number of parameters. 
For compactness, we denote $\dLdw$ as $\vG$ and $\dldwx{i}$ as $\vg_i$, using $\vidx{\vG}{j}$ and $\vidx{\vg_i}{j}$ to indicate the scalar value of a gradient at coordinate $j$. 

\begin{equation}\label{eqn.zsl.D.Dl}
    \zslm{D}{i}{N}{\dldwx{i}} 
        = 1 - 
            {|\tsum_i^N{\dldwx{i}}|} 
            \:/\:
            {\tsum_i^N |\dldwx{i}|} 
\end{equation}

\begin{equation}\label{eqn.zsl.D.g_vec}
    \zslm{D}{i}{N}{\vg_i} 
        = 1 - 
            {|\tsum_i^N{\vg_i}|} 
            \:/\:
            {\tsum_i^N |\vg_i|} 
\end{equation}

\begin{equation}\label{eqn.zsl.D.g_idx}
    \vidx{\zslm{D}{i}{N}{\vg_i}}{j} 
        = \zslm{D}{i}{N}{\vidx{\vg_i}{j}} 
        = 1 - 
            {|\tsum_i^N \vidx{\vg_i}{j}|} 
            \:/\: 
            {\tsum_i^N |\vidx{\vg_i}{j}|}
\end{equation}
\\

\noindent
Unless otherwise stated, moving forward $i$ indexes the $N$ examples used in computing the empirical average change in loss or gradient, and $j$ indexes the $M$ learnable model parameters flattened into a vector.

\paragraph{Change in loss under first-order training dynamics}~\\
Under first-order training dynamics, weight updates $\Dw$ are sufficiently small such that changes in loss (per-example $\Dlx{i}$, and overall $\DL$) are approximable by first-order Taylor expansions $\Dlxfote{i}$, $\DLfote$:
\begin{align}\label{eqn.zsl.fote}
    \Dlxfote{i} 
        &= \dotpB{\Dw}{\vg_{i}}
        = \tsum_{j}^{M} \vidx{\Dw}{j} \cdot \vidx{\vg_i}{j}
    \\[2ex]
    \DLfote 
        &= \dotpb{\Dw}{\vG} 
        = \tsum_{j}^{M} \vidx{\Dw}{j} \cdot \vidx{\vG}{j}
    \\
        &= \dotpb{\Dw}{\tfrac{1}{N}\tsum_i^{N}\vg_i} 
        = \tsum_{j}^{M} \tsum_{i}^{N} 
            \tfrac{1}{N} \cdot \vidx{\Dw}{j} \cdot \vidx{\vg_i}{j}
    \notag \\
        &= \tfrac{1}{N} \tsum_i^N \Dlxfote{i}
        = \tsum_{i}^{N} \tsum_{j}^{M} 
            \tfrac{1}{N} \cdot \vidx{\Dw}{j} \cdot \vidx{\vg_i}{j}
\end{align}

In such cases, ZSL is intrinsically a result of destructive interference in $\Dlxfote{i} = \dotpb{\Dw}{\dldwx{i}}$:

\begin{equation}\label{eqn.zsl.D.Dl_fote}
    \zslm{D}{i}{N}{\Dlxfote{i}}
        = 1 - 
        \frac
            {\abs{\tsum_{i}^{N} \dotpB{\Dw}{\vg_{i}}}}
            {\tsum_i^N \abs{\dotpB{\Dw}{\vg_{i}}}}
        = 1 - 
        \frac
            {\abs{\tsum_{i}^{N} \tsum_j^M 
                \vidx{\Dw}{j} \cdot \vidx{\vg_{i}}{j}}
            }
            {\tsum_i^N \abs{\tsum_j^M 
                \ \vidx{\Dw}{j} \cdot \vidx{\vg_{i}}{j}}
            }
\end{equation}

Note that \refeqn{eqn.zsl.D.Dl_fote} does not imply that gradient opposition necessarily results in ZSL. 
For instance, directions of high opposition in per-token gradients $\vg_{i}$ may be orthogonal to a weight update $\Dw$, such that they are nullified when $\vg_{i}$ is projected onto $\Dw$. 
Conversely, two gradients $\vg_{a}$, $\vg_{b}$ with no destructive interference may result in ZSL if e.g. $\Dw$ is aligned with $\vg_{a} - \vg_{b}$. 
In light of this, we want to disentangle ZSL in \refeqn{eqn.c2.fote_ldi} that is attributable to update-gradient alignment independent of gradient opposition, from ZSL attributable to gradient opposition specifically. 

\parnl{Isolating the role of gradient opposition}
Instead of destructive interference (\refeqn{eqn.zsl.D.Dl_fote}), we can equivalently consider and isolate the effect of gradient opposition in constructive interference (\refeqn{eqn.zsl.C.Dl_fote}) based on its identity as $1 - D$ (\refeqn{eqn.zsl.Ci}).

\begin{equation}\label{eqn.zsl.C.Dl_fote}
    \zslm{C}{i}{N}{\Dlxfote{i}}
        = 1 - \zslm{D}{i}{N}{\Dlxfote{i}} 
        = \frac
            {\abs{\tsum_{i}^{N} \tsum_j^M 
                \vidx{\Dw}{j} \cdot \vidx{\vg_{i}}{j}}
            }
            {\tsum_i^N \abs{\tsum_j^M 
                \ \vidx{\Dw}{j} \cdot \vidx{\vg_{i}}{j}}
            }
\end{equation}
\\

\noindent For compactness, we will denote $\vidx{p_i}{j} = \vidx{\Dw}{j} \cdot \vidx{\vg_i}{j}$ and $\vidx{q}{j} = \vidx{\Dw}{j} \cdot \vidx{\vG}{j}$. Intuitively, $\vidx{p_i}{j}$ captures the contribution of example $i$ and parameter $j$ to the first-order change in loss $\DLfote = \tfrac{1}{N} \tsum_i^N \tsum_j^M \vidx{p_i}{j}$. Note that $\vg$ is not normalized by number of examples, i.e. $\vidx{\vG}{j} = \tfrac{1}{N} \tsum_{i}^N \vidx{\vg_i}{j}$ and $\vidx{q}{j} = \tfrac{1}{N} \tsum_{i}^N \vidx{p_i}{j}$. 
\\

\noindent Our goal is to isolate the effect of gradient opposition in \refeqn{eqn.zsl.C.Dl_fote}. To this end, we rewrite the numerator (\refeqn{eqn.zsl.C.Dl_fote.numer}) and denominator (\refeqn{eqn.zsl.C.Dl_fote.denom}) in terms of constructive interference in $\vidx{p_i}{j}$ and $\vidx{q}{j}$, leveraging the fact that 
$\abstt{\tsum_{i}^{N} x_i} = \zslm{C}{i}{N}{x_i} \cdot \tsum_{i}^{N} \abstt{x_i}$. 
Crucially, $\zslm{C}{i}{N}{\vg_i} = \zslm{C}{i}{N}{p_i}$, allowing us to finally isolate the effect of gradient opposition in \refeqn{eqn.zsl.C.Dl_fote.v2}, with all other terms being independent of gradient opposition. 

\begin{align}
    \abs{\tsum_{i}^{N} \tsum_j^M \vidx{\Dw}{j} \cdot \vidx{\vg_{i}}{j}}
        &= \abs{\tsum_j^M 
           \vidx{\Dw}{j} \cdot N \vidx{\vG}{j}}
    \label{eqn.zsl.C.Dl_fote.numer} 
    \\
    &= \tsum_j^M 
        N \cdot \abs{\vidx{q}{j}} 
        \cdot \zslm{C}{j'}{M}{\vidx{q}{j'}}
    \notag \\
    &= \tsum_i^N \tsum_j^M 
        \abs{\vidx{p_i}{j}} \cdot \zslm{C}{i'}{N}{\vidx{p_{i'}}{j}}
        \cdot \zslm{C}{j'}{M}{\vidx{q}{j'}}
    \notag \\
    &= \tsum_i^N \tsum_j^M 
        \abs{\vidx{p_i}{j}} \cdot \zslm{C}{i'}{N}{\vidx{\vg_{i'}}{j}}
        \cdot \zslm{C}{j'}{M}{\vidx{q}{j'}}
    \notag 
    \\[3ex]
    \tsum_i^N \abs{\tsum_j^M \vidx{\Dw}{j} \cdot \vidx{\vg_{i}}{j}}
        &= \tsum_i^N \abs{\tsum_j^M \vidx{p_i}{j}} 
        \label{eqn.zsl.C.Dl_fote.denom} 
        \\
        &=\tsum_i^N \tsum_j^M 
            \abs{\vidx{p_i}{j}}
            \cdot \zslm{C}{j'}{M}{\vidx{p_i}{j'}}
    \notag 
    \\[3ex]
    \zslm{C}{i}{N}{\Dlxfote{i}}
        &= \frac
            {
                \tsum_i^N \tsum_j^M 
                    \abs{\vidx{p_i}{j}} 
                    \cdot 
                    \zslm{C}{i'}{N}{\vidx{\vg_{i'}}{j}}
            }
            {
                \tsum_i^N \tsum_j^M 
                    \abs{\vidx{p_i}{j}} 
                    \cdot 
                    \zslm{C}{j'}{M}{\vidx{p_i}{j'}}
            }
            \cdot \zslm{C}{j'}{M}{\vidx{q}{j'}}
        \label{eqn.zsl.C.Dl_fote.v2}
\end{align}

\noindent Intuitively, the constructive interference terms in \refeqn{eqn.zsl.C.Dl_fote.v2} can be interpreted as: 

\begin{align}
   \zslm{C}{i'}{N}{\vidx{\vg_{i'}}{j}}
        &\quad\text{ (lack of) gradient opposition across examples $i'$,  independent of $\Dw$}
    \notag \\
    \zslm{C}{j'}{M}{\vidx{p_i}{j'}}
        &\quad\text{ alignment between $\Dw, \vg_i$, independent of gradient opposition}
        \notag \\
    \zslm{C}{j'}{M}{\vidx{q}{j'}}
        &\quad\text{ alignment between $\Dw, \vG$, independent of gradient opposition}
    \notag
\end{align}

\clearpage
\parnl{Quantifying the role of gradient opposition}
To disentangle and quantify the role of gradient opposition in $\zslm{C}{i}{N}{\Dlxfote{i}}$, we can rewrite \refeqn{eqn.zsl.C.Dl_fote.v2} as: 

\begin{equation}
    \zslm{C}{i}{N}{\Dlxfote{i}} = 
        \cg \cdot \frac{\cug}{\cu}
\end{equation}

\noindent $\cg \in [0,1]$ captures constructive interference in $\DLfote = \tfrac{1}{N} \tsum_i^N \tsum_j^M \vidx{p_i}{j}$ attributable to (lack of) gradient opposition, weighted by the relative magnitude of the update at coordinate $j$.

\begin{equation}
    \cg = \frac
    {\tsum_i^N \tsum_j^M \abs{\vidx{p_i}{j}} \cdot \zslm{C}{i'}{N}{\vidx{\vg_{i'}}{j}}}
    {\tsum_i^N \tsum_j^M \abs{\vidx{p_i}{j}}}
    = \frac
        {\tsum_j^M \abs{\tsum_i^N \vidx{p_i}{j}}}
        {\tsum_i^N \tsum_j^M \abs{\vidx{p_i}{j}}}
\end{equation}

\noindent $\cu \in [0,1]$ captures constructive interference in $\DLfote = \tfrac{1}{N} \tsum_i^N \tsum_j^M \vidx{p_i}{j}$ attributable solely to update-gradient alignment, aggregated over all examples $i$ and independent of gradient opposition.

\begin{equation}
    \cu = \frac
    {\tsum_i^N \tsum_j^M \abs{\vidx{p_i}{j}} \cdot \zslm{C}{j'}{M}{\vidx{p_i}{j'}}}
    {\tsum_i^N \tsum_j^M \abs{\vidx{p_i}{j}}}
    = \frac
        {\tsum_i^N \abs{\tsum_j^M \vidx{p_i}{j}}}
        {\tsum_i^N \tsum_j^M \abs{\vidx{p_i}{j}}}
\end{equation}

\noindent $\cug \in [0,1]$ captures constructive interference in $\DLfote = \tfrac{1}{N} \tsum_i^N \tsum_j^M \vidx{p_i}{j}$ attributable solely to update-gradient alignment for the overall gradient, independent of gradient opposition.

\begin{equation}\label{eqn.zsl.C.Dl_fote.v3}
    \cug = \zslm{C}{j'}{M}{\vidx{q}{j'}}
    = \frac
        {\abs{{\tsum_j^M \vidx{q}{j}}}}
        {\tsum_j^M \abs{\vidx{q}{j}}}
    = \frac
        {\abs{{\tsum_i^N \tsum_j^M \vidx{p_i}{j}}}}
        {\tsum_j^M \abs{\tsum_i^N \vidx{p_i}{j}}}
\end{equation}

\clearpage
\subsection{Effect of Increasing Steps on ZSL}
\label{app.zsl_incr_steps}
\begin{figure}[h]
    \centering
     \captionsetup{width=\linewidth}
     \begin{subfigure}[b]{0.35\linewidth}
         \centering
         \includegraphics[width=\linewidth]{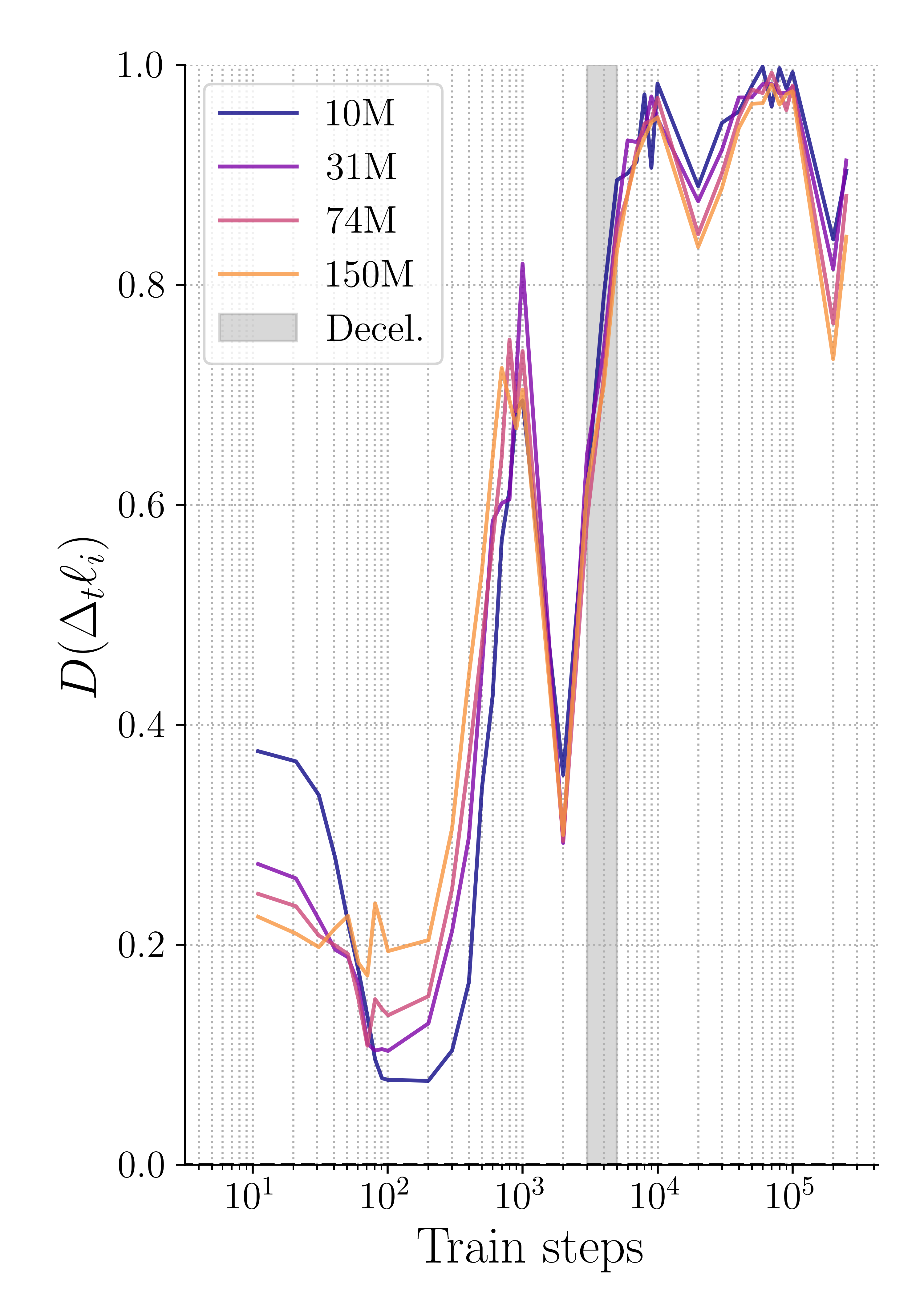}
        \subcaption[]{}
     \end{subfigure}
     \begin{subfigure}[b]{0.25\linewidth}
         \centering
         \includegraphics[width=\linewidth]{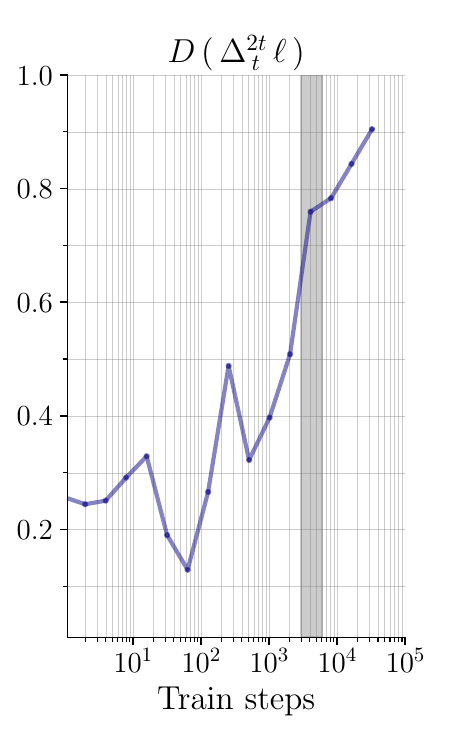}
        \subcaption[]{}
     \end{subfigure}
     \begin{subfigure}[b]{0.25\linewidth}
         \centering
         \includegraphics[width=\linewidth]{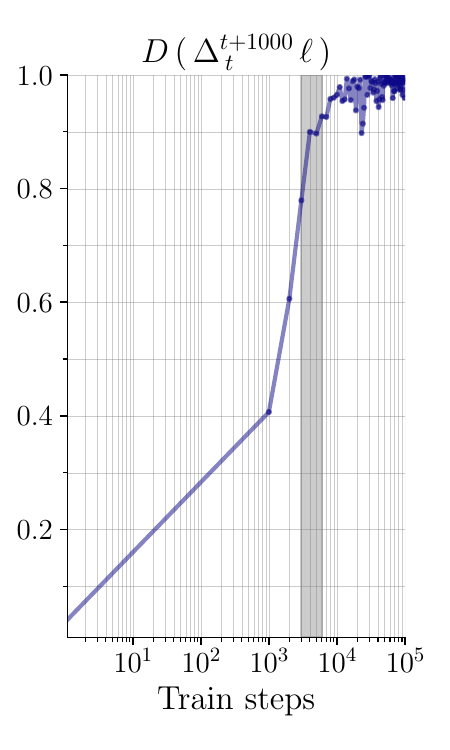}
    \subcaption[]{}
     \end{subfigure}
    \caption{Effect of number of steps (between changes in loss) on destructive interference in per-example loss improvements, across several experiments. 
    \textbf{(a)} Initial experiments based on checkpoints for steps $[1,2,\dots,10,20,\dots,100,200,\dots,1000,2000,\dots]$, i.e. varying $\Delta t$. 
    \textbf{(b)} The nanoGPT baseline from \refapp{app.ablation_optimizer} based on checkpoints for steps $[1,2,4,8,\dots]$ as in our main results. 
    \textbf{(c)} The same nanoGPT baseline but for checkpoints $[0,1000,2000,\dots]$, i.e. $\Delta t = 1000$. These different results point to the fact that increasing the number of steps for measuring loss improvements mitigates destructive interference in loss improvements.}
    \label{fig.app.zsl_incr_steps}
\end{figure}

%
\parnl{Destructive interference is mitigated by increasing number of steps.}
While the experiments and results in \refsec{sec.expl.cooccur} consider the change in loss between steps $t$ and $2t$, our initial experiments were based on checkpoints for steps $[1,2,\dots,10,20,\dots,100,200,\dots,1000]$ and so on. 
When plotting $\DI{\Dlx{}}$ between these checkpoints in \reffig{fig.app.zsl_incr_steps}, we can see that $\DI{\Dlx{}}$ increases much more rapidly leading up to deceleration, when compared to \reffig{fig.zsl.zsl_multistep_log2_DI}. 
However, we also observe abrupt drops and subsequent rises in $\DI{\Dlx{}}$ after the number of steps between checkpoints is increased by a factor of $10$. 
These results highlight that ZSL actually increases leading up to (rather than during) deceleration, but is mitigated by increasing number of steps between loss measurements.
Indeed, comparing the baseline nanoGPT model from \refapp{app.ablation_optimizer} in \reffig{fig.app.zsl_incr_steps}, we see that ZSL rises faster and approaches a maximum \textit{at} deceleration when $\Delta t = 1000$. In contrast, when $\Delta t = t$ as in our main results, ZSL rises slower and approaches a maximum \textit{after} deceleration.

\clearpage
\parnl{Destructive interference in loss improvements after one step}
In the extreme, we can consider loss improvements after only one optimizer step, although this can be quite noisy and vary significantly between consecutive steps. 
Nevertheless, this allows us to compare "Train" and "Eval"  batches, i.e. the training batch for the given optimizer step and a withheld validation batch. We observe several surprising trends that are distinct from our main results (where we considered loss improvements over multiple steps to avoid issues with noise). Notably, we observe:

\setlist[enumerate,1]{leftmargin=3em}
\begin{enumerate}[
    align=parleft, topsep=0ex,itemsep=0ex,partopsep=1ex,parsep=0.5ex, labelwidth=1em
]
    \item[$\bm{(1)}$] In \texttt{Eval.} examples, destructive interference in loss improvements approaches it's maximum well before deceleration. Note that the dip before deceleration actually corresponds to a period where loss degrades after one step, suggesting overfitting of the training data that ends near deceleration, when destructive interference stabilizes around $1.0$. 
    \item[$\bm{(2)}$] In contrast, destructive interference in \texttt{Train.} examples increases until deceleration, but actually decreases after deceleration. In particular, there is a clear trend where increased model size correlates with decreased destructive interference after deceleration. 
\end{enumerate}

\begin{figure}[!h]
     \centering
     \captionsetup{width=1\linewidth}
     \begin{subfigure}[b]{0.3\linewidth}
         \centering
         \includegraphics[width=\linewidth]{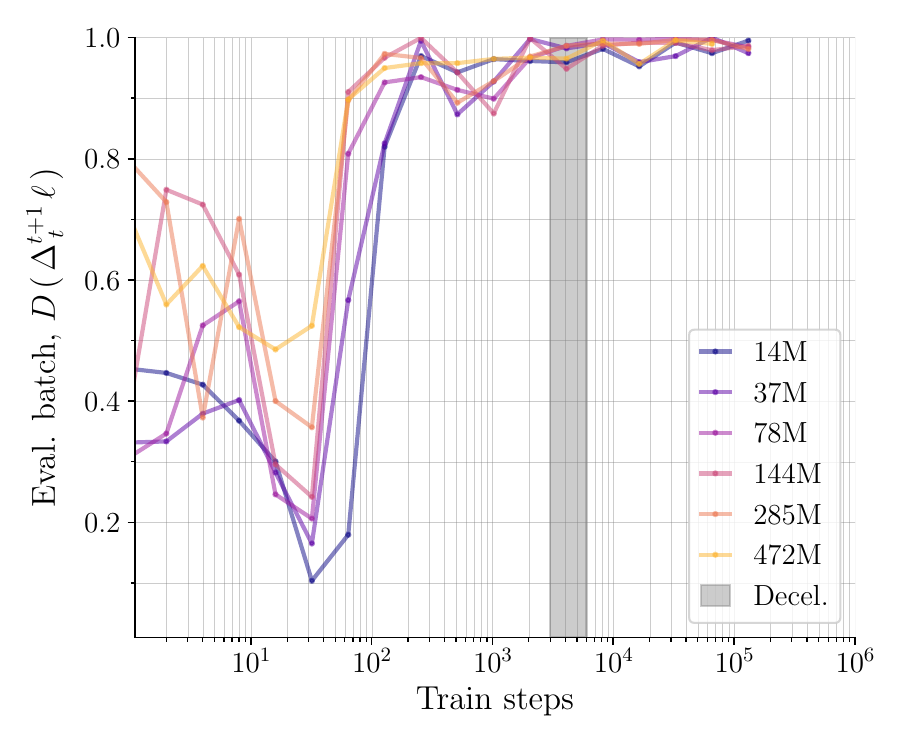}
     \end{subfigure}
     \begin{subfigure}[b]{0.3\linewidth}
         \centering
         \includegraphics[width=\linewidth]{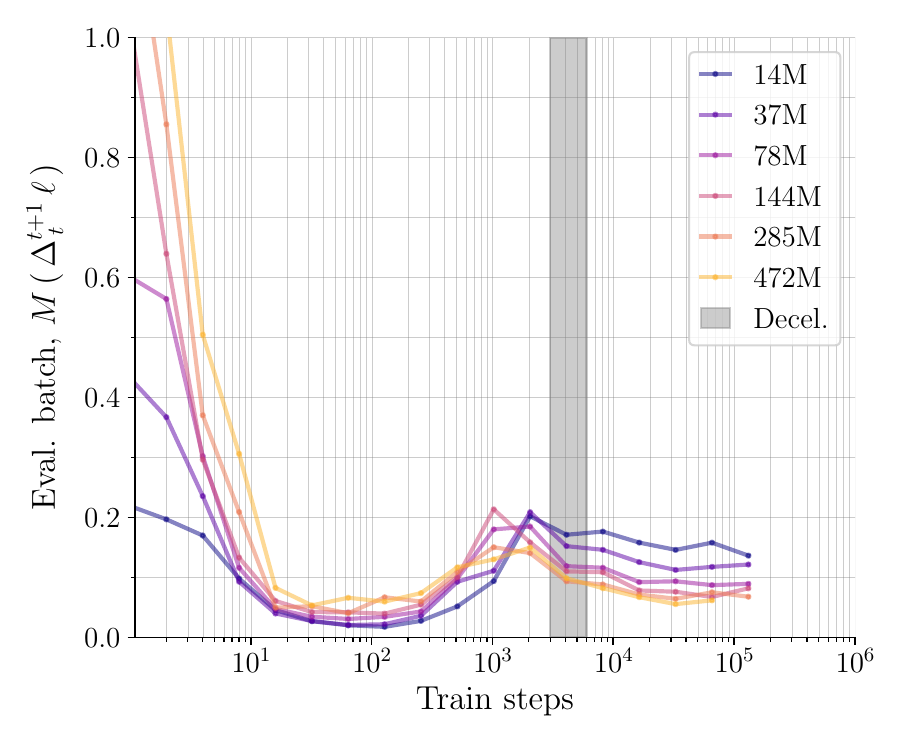}
     \end{subfigure}
     \begin{subfigure}[b]{0.3\linewidth}
         \centering
         \includegraphics[width=\linewidth]{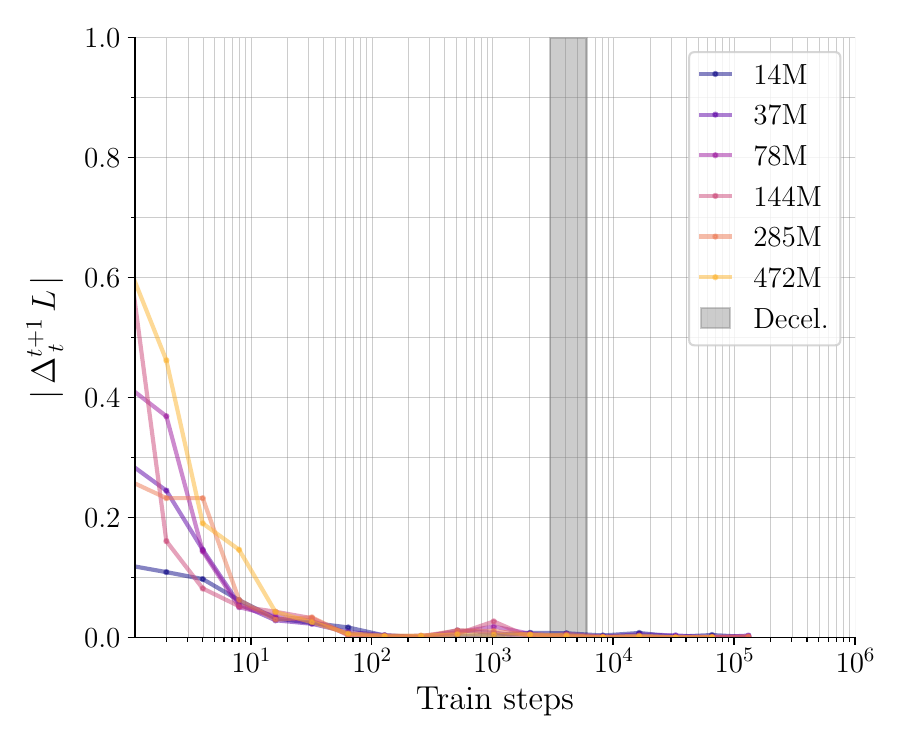}
     \end{subfigure}
      \begin{subfigure}[b]{0.3\linewidth}
     \centering
     \includegraphics[width=\linewidth]{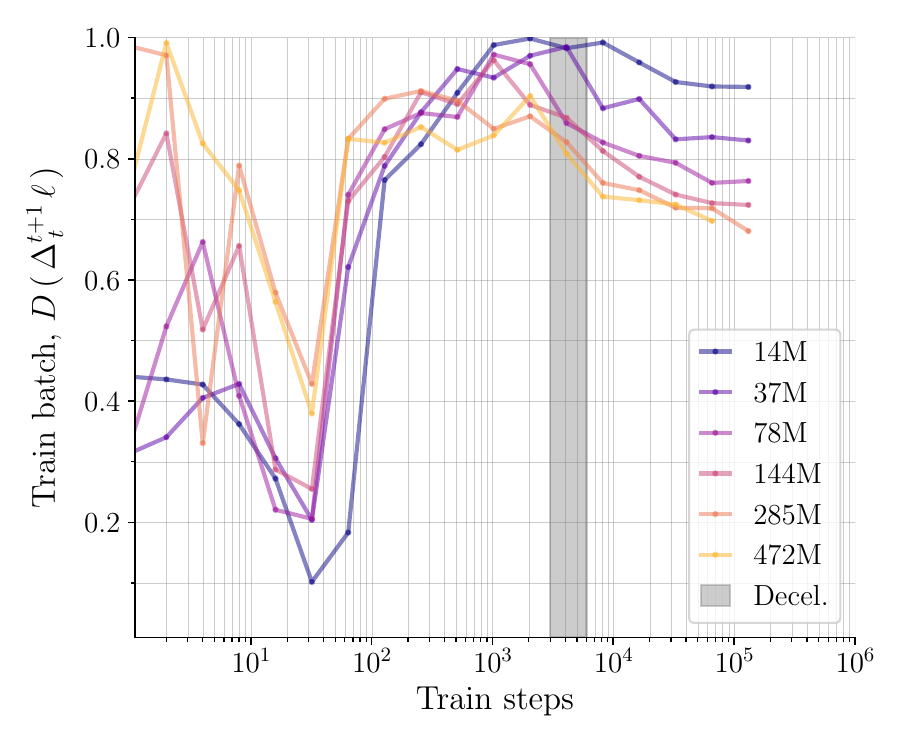}
 \end{subfigure}
 \begin{subfigure}[b]{0.3\linewidth}
     \centering
     \includegraphics[width=\linewidth]{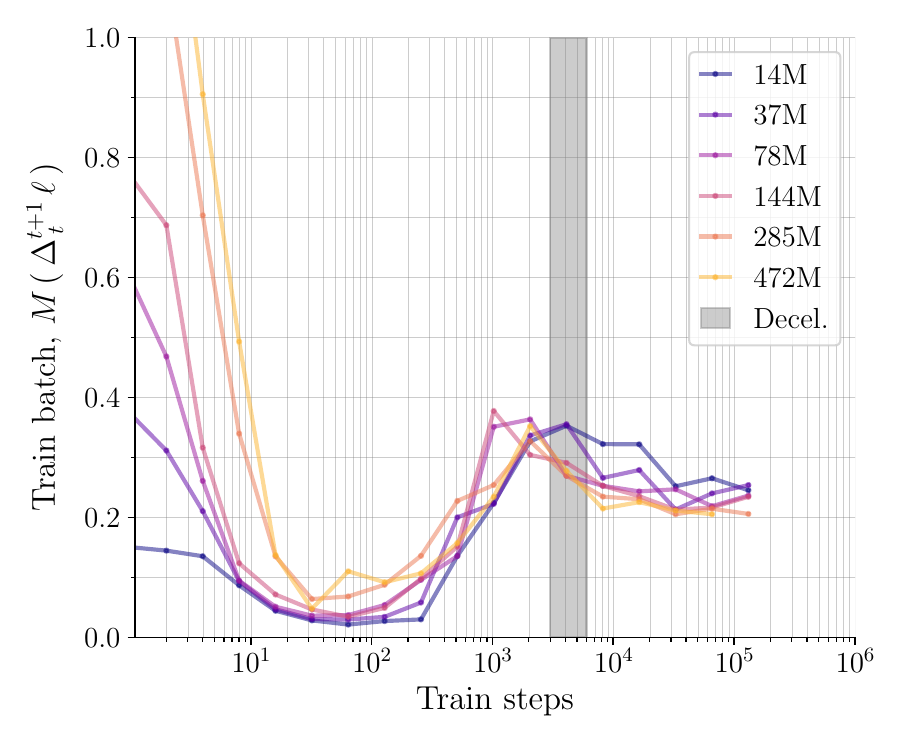}
 \end{subfigure}
 \begin{subfigure}[b]{0.3\linewidth}
     \centering
     \includegraphics[width=\linewidth]{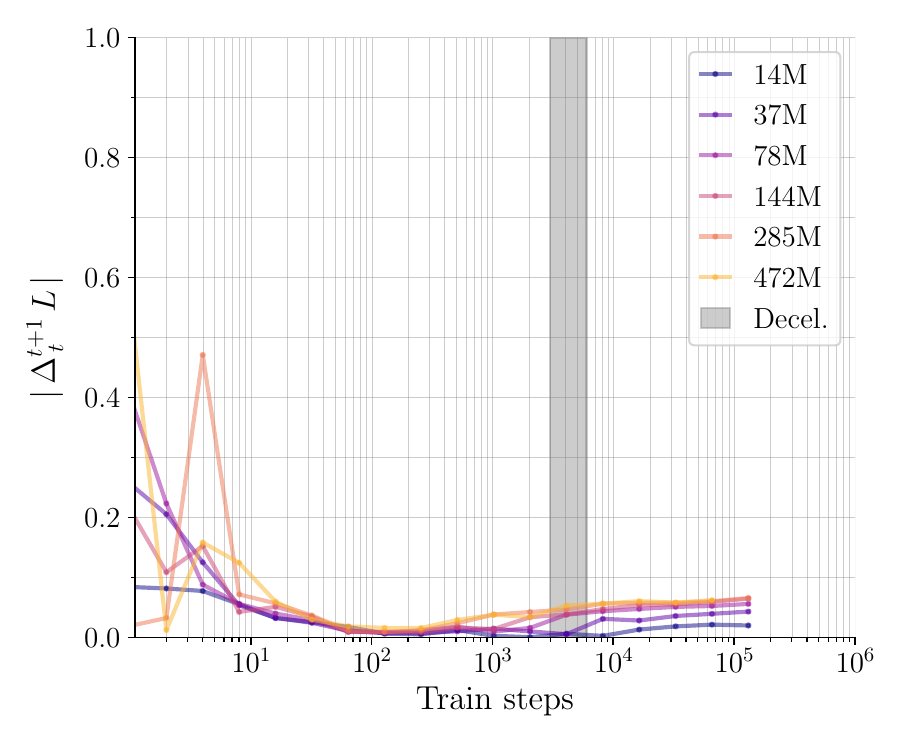}
 \end{subfigure}
\caption{Single-step ZSL in Train and Eval. batches.}
\label{fig.app.zsl_onestep}
\end{figure}

How these two observations relate to one-another and the multi-step behaviour seen in our main results is not clear. Counter-intuitively, it seems the only way in which larger models are markedly better is in their ability to overfit examples from a given training batch in one step, without generalization to other examples. However, despite this, our multi-step results suggest that larger models improve generalization, given the single epoch training on which our results are based. 
This implies that while overfitting occurs in any one step, there is generalization that occurs over many steps. One informal explanation for this phenomenon could be that larger models are better able to find common gradient directions for a given batch of data, leading to overfitting when considering one step, but better generalization when considering multiple steps. This interpretation is consistent with several other observations that learning after deceleration is more strongly associated with generalization (notably in-context learning ability \cite{olsson_-context_2022} as measured by improvements in later tokens, \refapp{app.ablation_bsz_seq}; and greater improvements on downstream tasks compared to a model with the same loss but before deceleration \refapp{app.res.olmo_downstream}).
Making this admittedly vague notion more precise and exploring it more rigorously is something we leave for future work.

\clearpage
\subsection{Loss Deceleration and Downstream Performance}
\label{app.res.olmo_downstream}
We compare the dowsntream performance of OLMo-1B and OLMo-7B on several downstream tasks as reported by \citet{groeneveld_olmo_2024}. Surprisingly, we find that OLMo-1B checkpoints after deceleration typically outperform OLMo-7B checkpoints with the same loss before and during deceleration. 
This suggests that zero-sum learning plays an important role in generalization and that loss improvements after deceleration result in more significant generalization capabilities than equivalent loss improvements before deceleration. 

\begin{figure}[!h]
     \centering
     \captionsetup{width=0.9\linewidth}
     \begin{subfigure}[b]{0.45\linewidth}
         \centering
         \includegraphics[width=\linewidth]{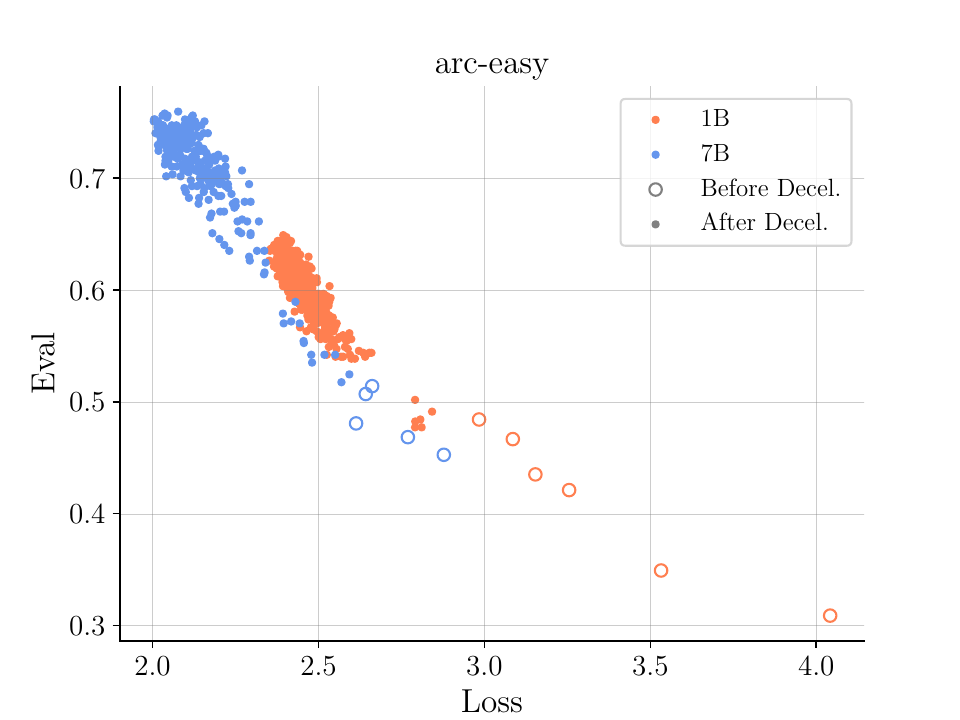}
     \end{subfigure}
     \begin{subfigure}[b]{0.45\linewidth}
         \centering
         \includegraphics[width=\linewidth]{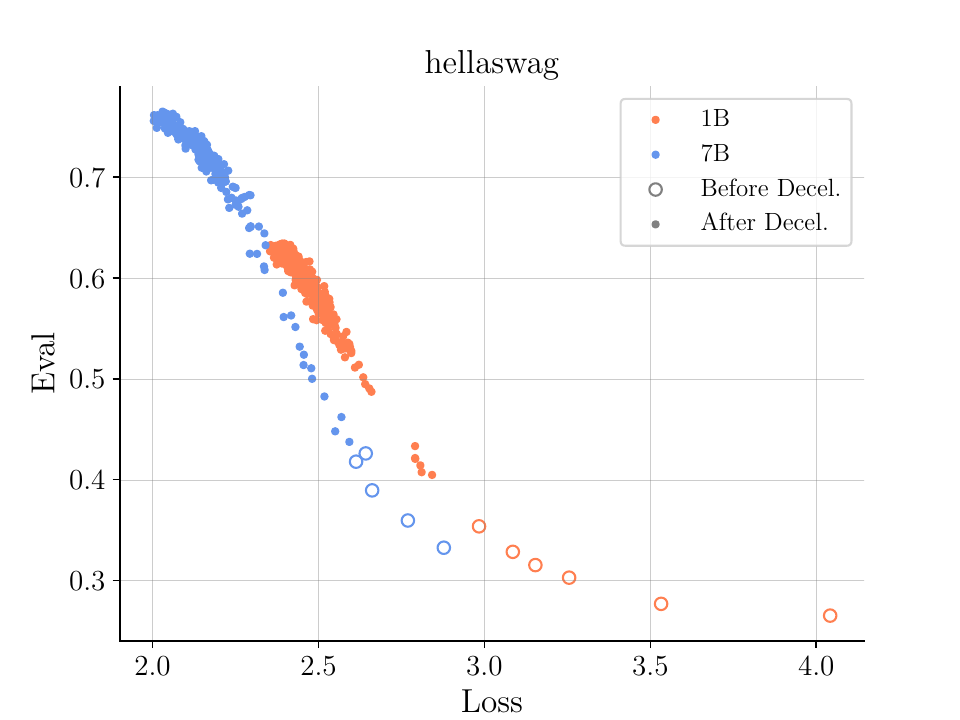}
     \end{subfigure}
     \begin{subfigure}[b]{0.45\linewidth}
         \centering
         \includegraphics[width=\linewidth]{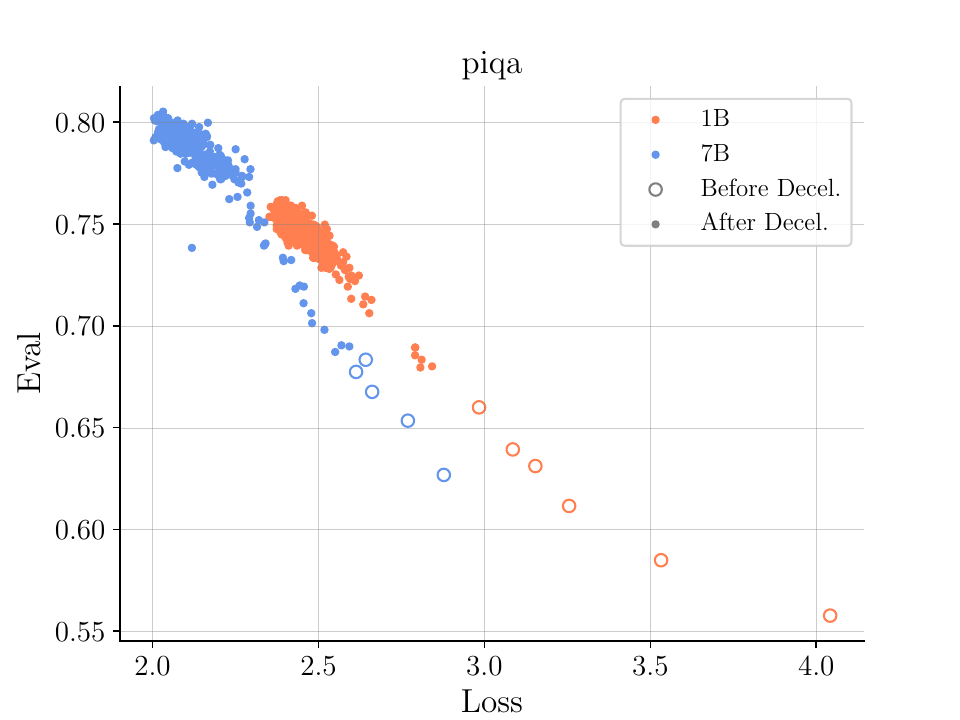}
     \end{subfigure}
      \begin{subfigure}[b]{0.45\linewidth}
     \centering
         \includegraphics[width=\linewidth]{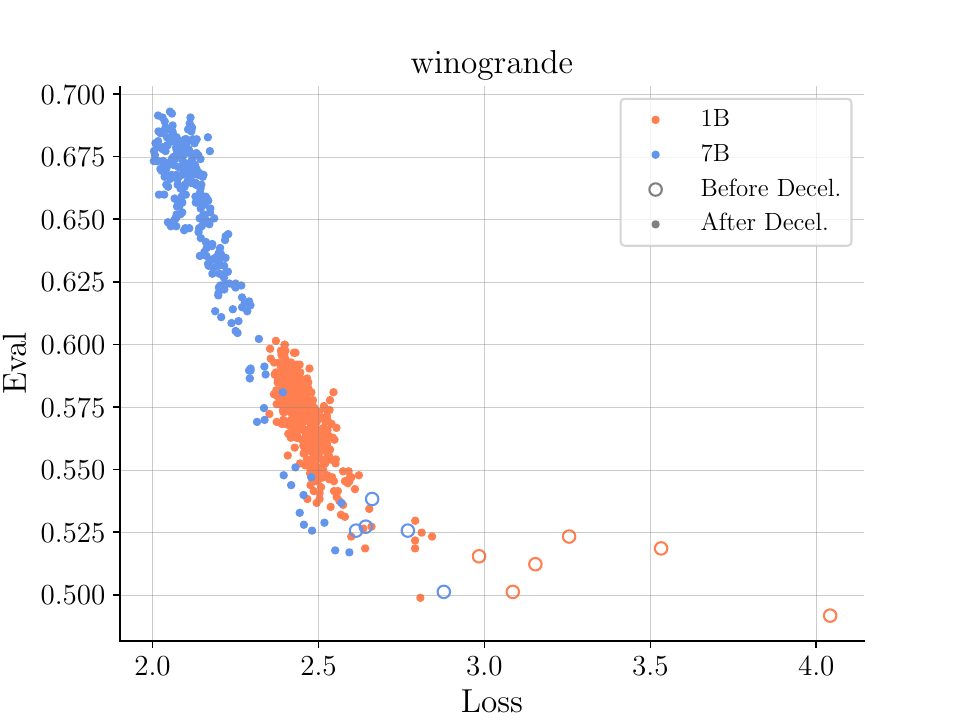}
 \end{subfigure}
\caption{Downstream performance of OLMo-1B and OLMo-7B on different tasks. Before and during deceleration, OLMo-7B checkpoints underperform OLMo-1B checkpoints with the same loss but after deceleration. This suggests zero-sum learning after deceleration plays an important role in generalization.}
\label{fig.app.olmo_downstream}
\end{figure}

\clearpage
\onecolumn
\subsection{Per-token loss landscape cross-sections}
\label{app.res.xsections}
\begin{figure}[h]
\centering
    \captionsetup{width=\linewidth}
    \includegraphics[width=0.8\linewidth]{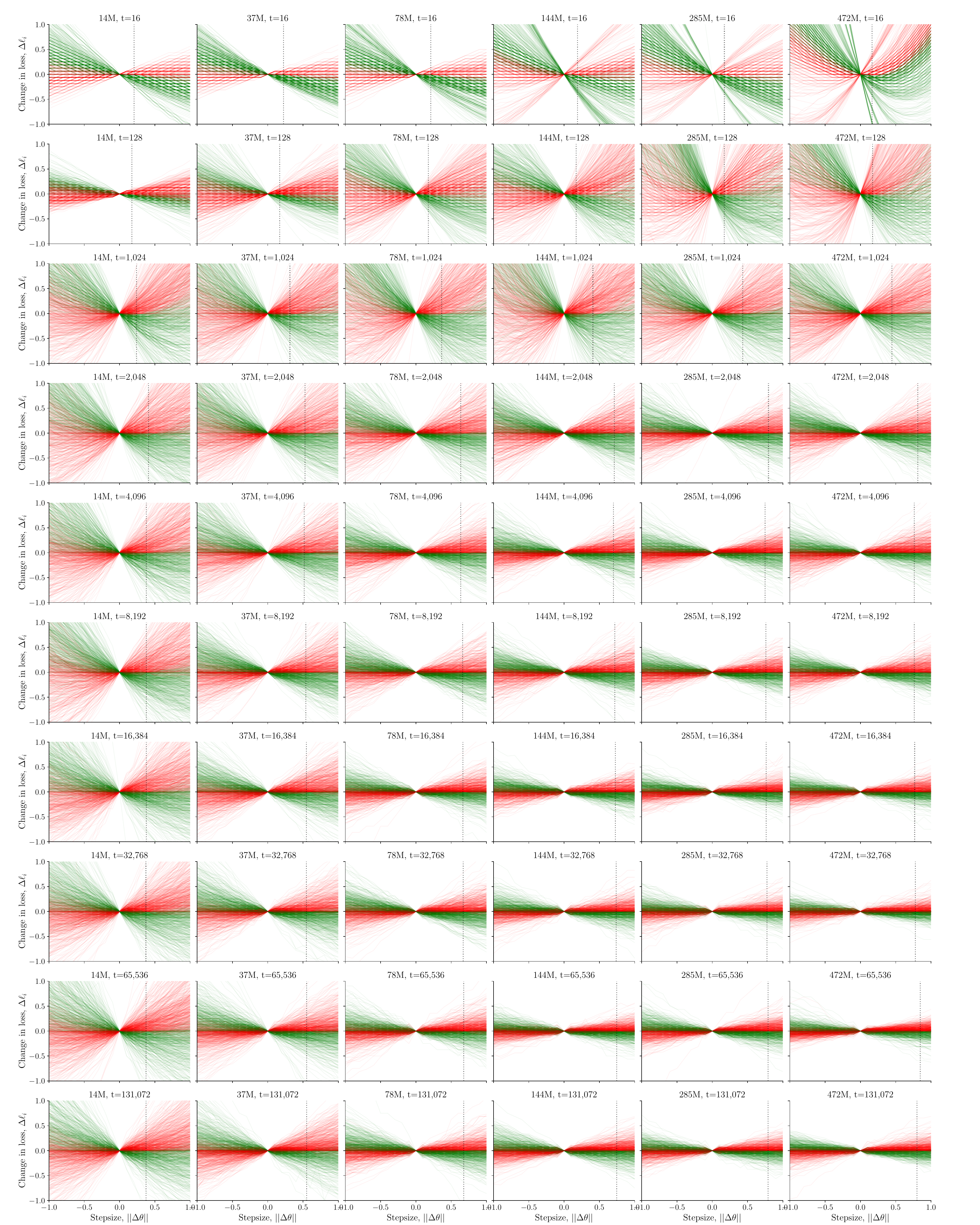}
    \caption{
    \textbf{Sampled per-token loss landscape cross-sections across model sizes and train steps}  \hfill \\
    Across model sizes (columns) and train steps (rows), we plot loss landscape cross-sections along increments of the weight update $\Dw$ at step $t$. The actual stepsize is indicated with a dotted vertical line. We plot $\DL$ rather than $L$, which has the same geometry but allows more easily distinguishing loss improvements from degradations.  Lines are colored in green or red depending on whether the loss (respectively) improved or deteriorated at the actual stepsize.
     }
    \label{fig.app.xsections}
\end{figure}

\begin{figure}[h]
\centering
    \captionsetup{width=\linewidth}
    \includegraphics[width=0.8\linewidth]{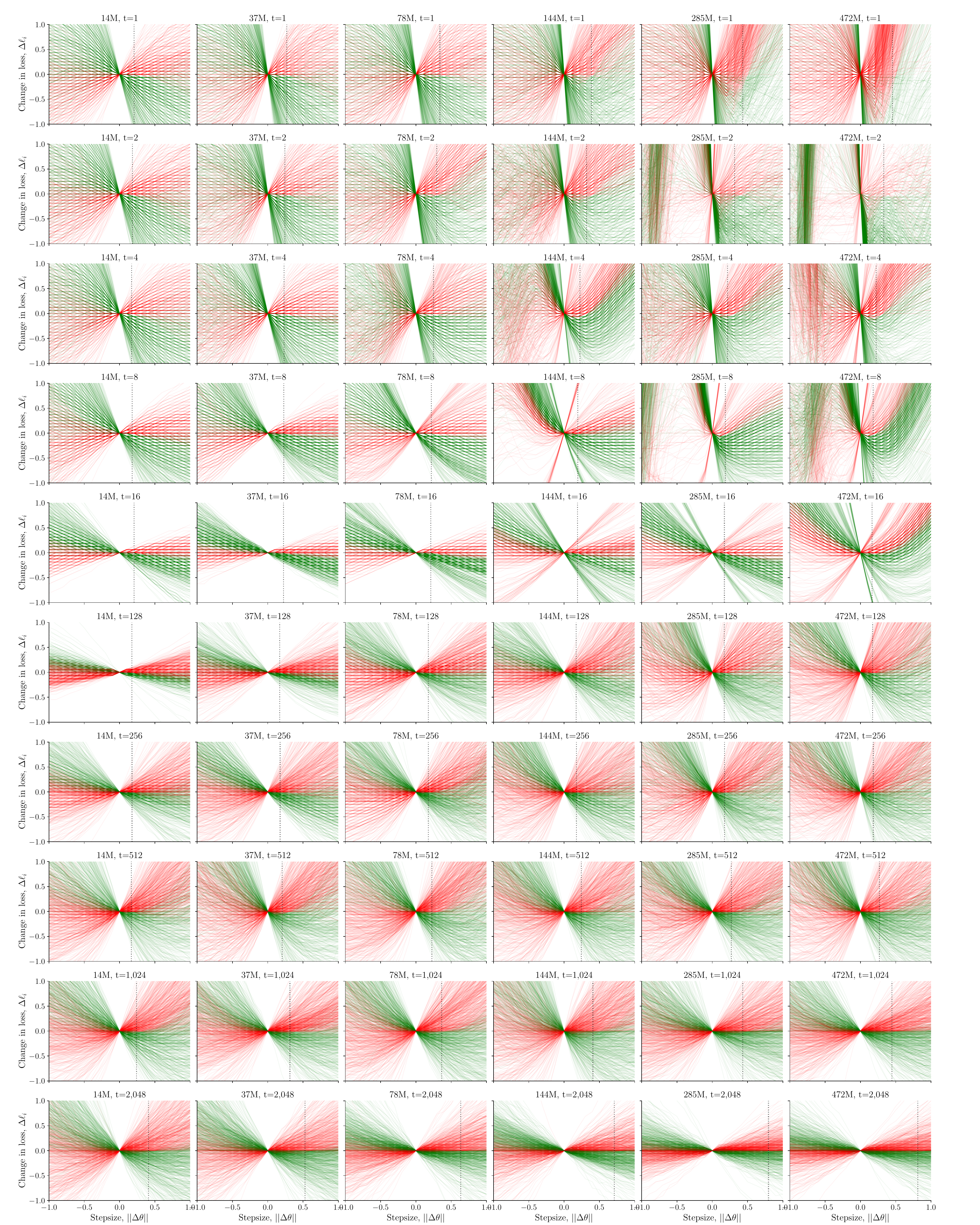}
    \caption{
    \textbf{Sampled per-token loss landscape cross-sections across model sizes at the start of training}  \hfill \\
      We plot the same data as in \reffig{fig.app.xsections}, but focused on the beginning of training (before deceleration).   
     }
    \label{fig.app.xsections_predecel}
\end{figure}

\clearpage
\subsection{Overall loss landscape cross-sections throughout training}
\label{app.res.landscapes}
\begin{figure}[!h]
     \centering
     \captionsetup{width=\textwidth}
     \begin{subfigure}[b]{0.3\textwidth}
         \centering
         \includegraphics[trim={3.8cm 4cm 4cm 4cm},clip,width=\linewidth]{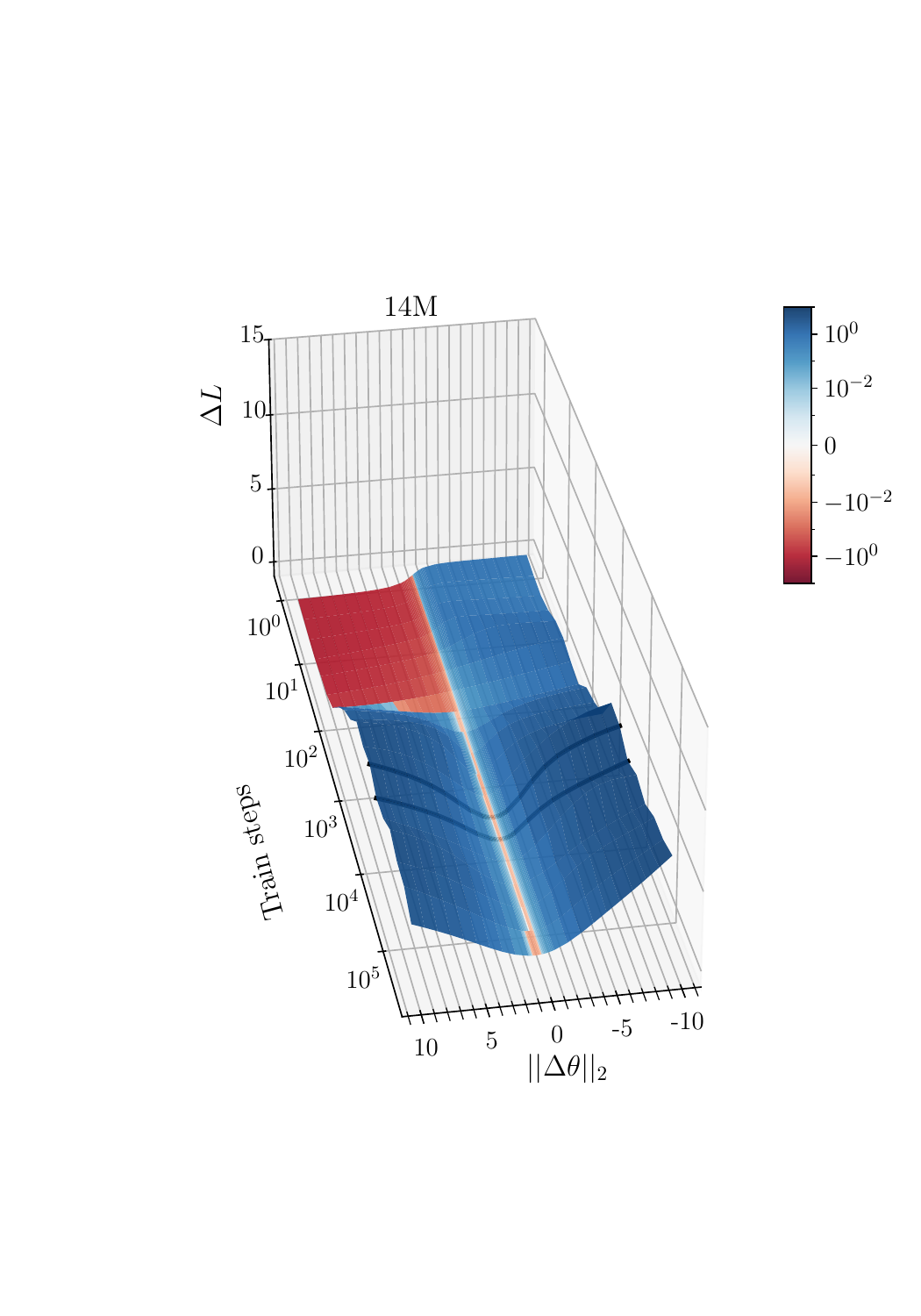}
     \end{subfigure}
     \hfill
     \begin{subfigure}[b]{0.3\textwidth}
         \centering
         \includegraphics[trim={3.8cm 4cm 4cm 4cm},clip,width=\linewidth]{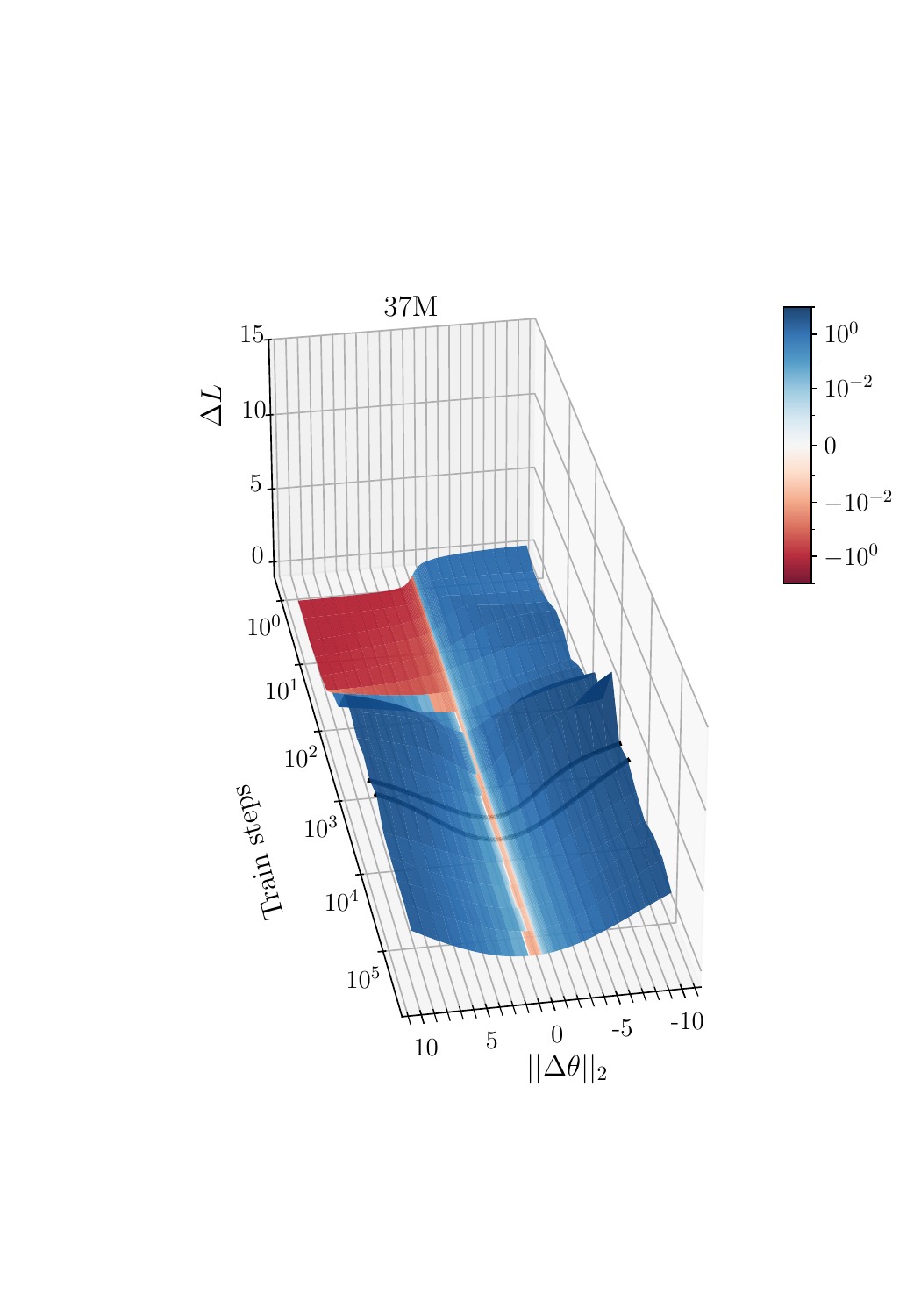}
     \end{subfigure}
     \hfill
      \begin{subfigure}[b]{0.3\textwidth}
         \centering
         \includegraphics[trim={3.8cm 4cm 4cm 4cm},clip,width=\linewidth]{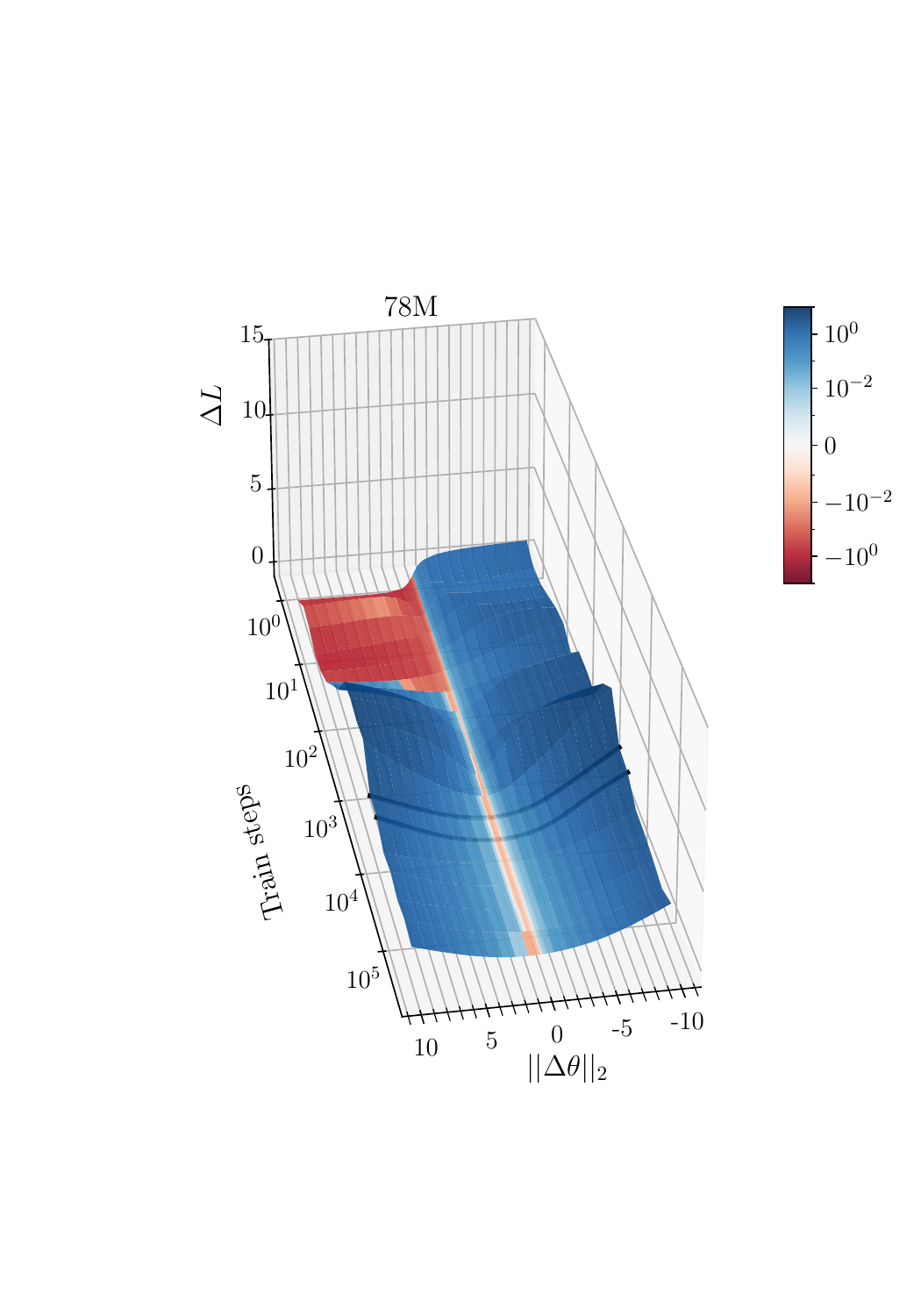}
     \end{subfigure}
     \begin{subfigure}[b]{0.31\textwidth}
         \centering
         \includegraphics[trim={3.8cm 4cm 4cm 4cm},clip,width=\linewidth]{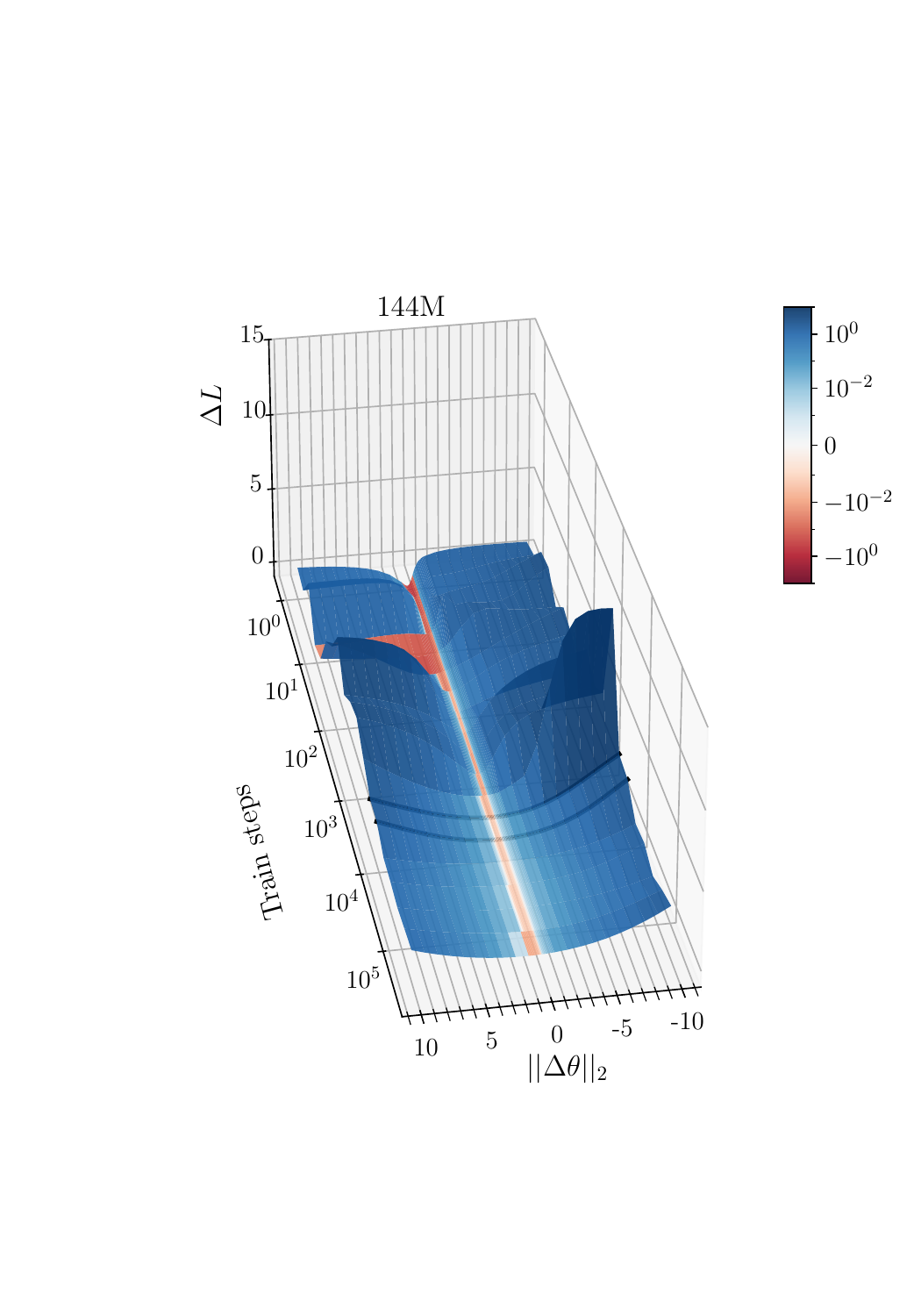}
     \end{subfigure}
     \hfill
     \begin{subfigure}[b]{0.31\textwidth}
         \centering
         \includegraphics[trim={3.8cm 4cm 4cm 4cm},clip,width=\linewidth]{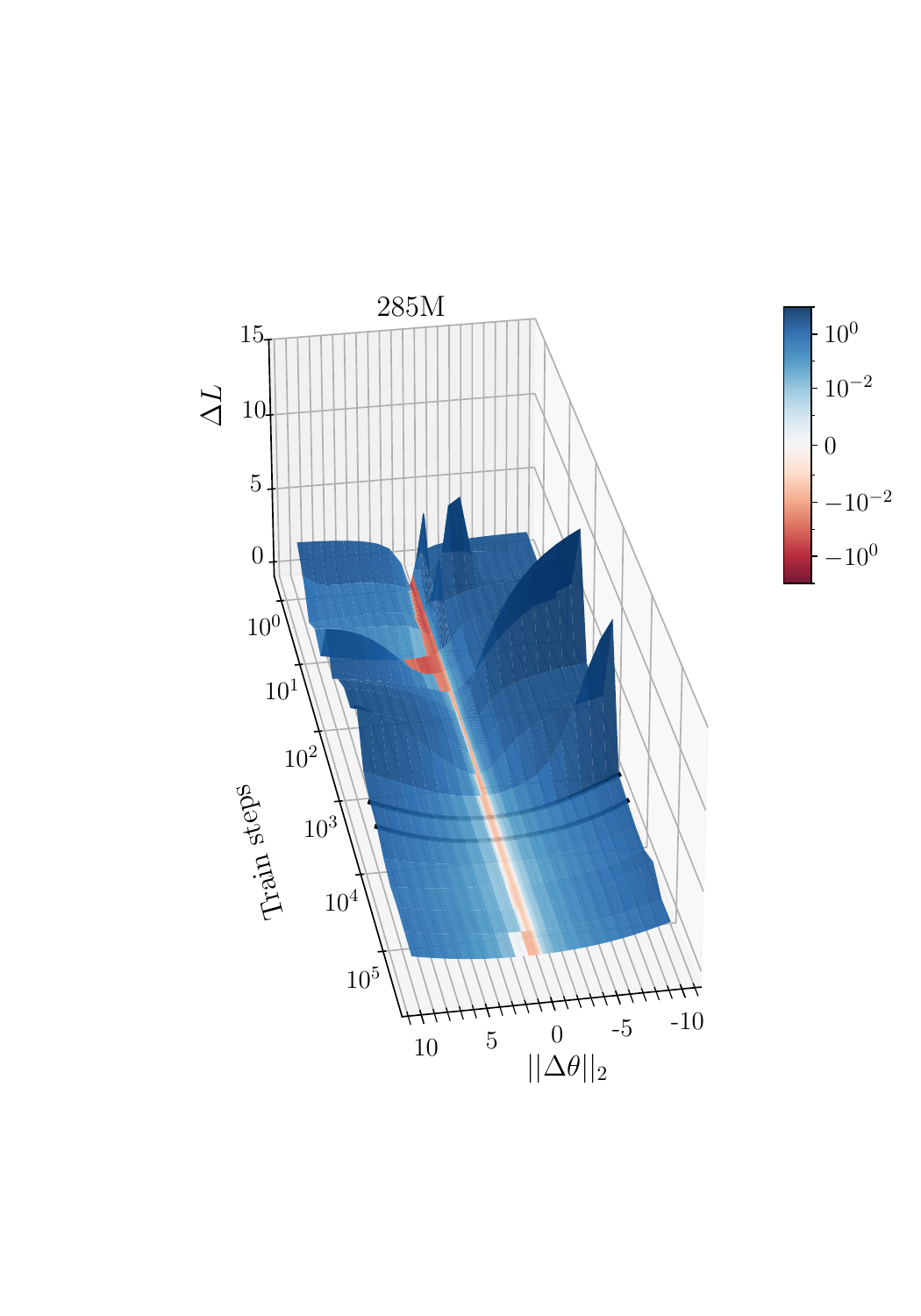}
     \end{subfigure}
     \hfill
      \begin{subfigure}[b]{0.31\textwidth}
         \centering
         \includegraphics[trim={3.8cm 4cm 4cm 4cm},clip,width=\linewidth]{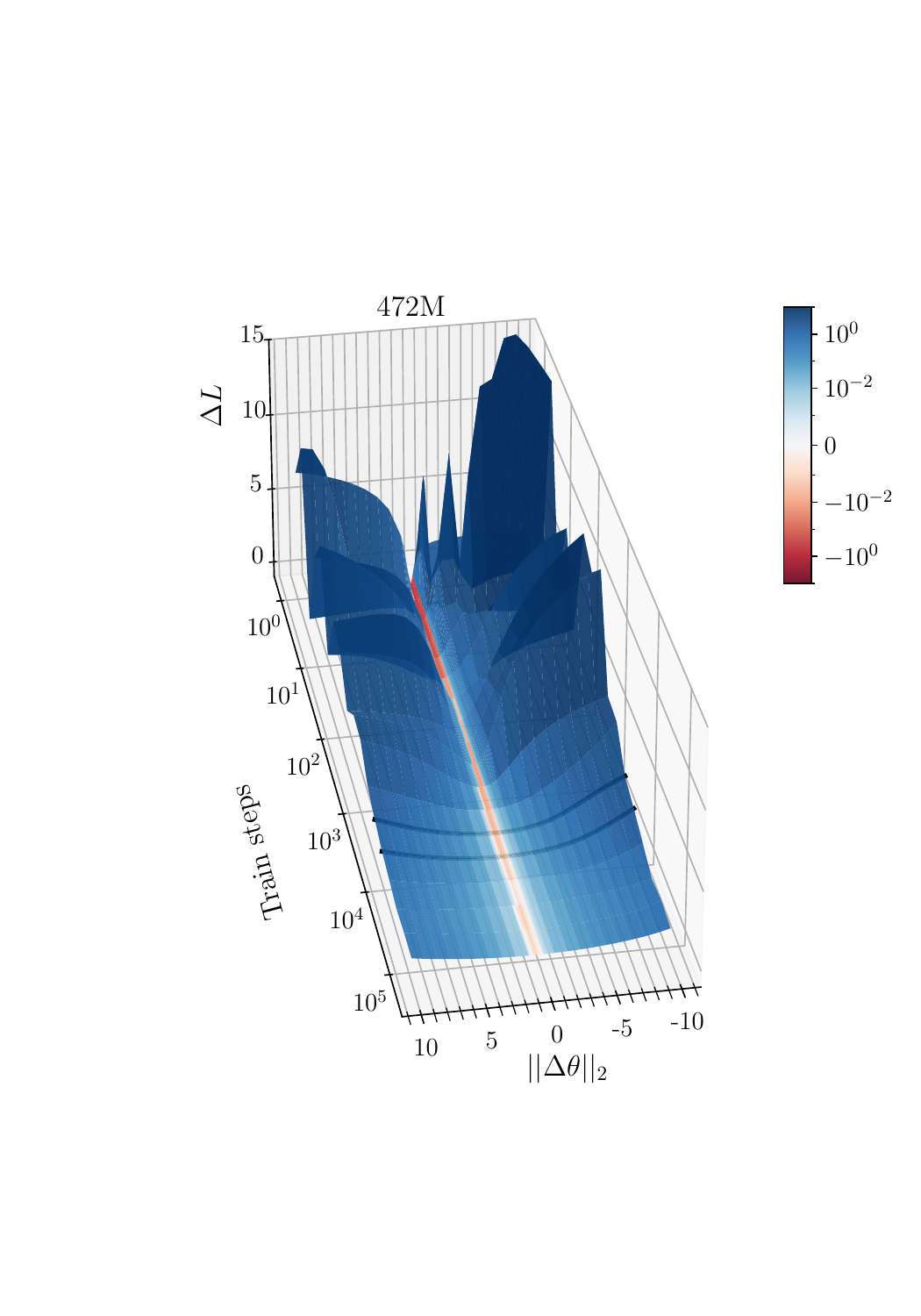}
     \end{subfigure}
     
    \caption{
    \textbf{Overall loss landscapes (cross section along $\Dw$), visualized throughout training}  \hfill \\
    We plot overall loss landscape cross sections across model sizes and train steps. Similar to \refapp{app.res.xsections}, we plot $\DL$ which has equivalent geometry to $L$ but allows better distinguishing loss improvements from loss degradations. $\DL$ is additionally indicated with a symlog colorscale, with loss improvements being red. Loss deceleration is approximately indicated with two lines at $t=4096$ and $t=8192$. We observe that loss landscapes sharpen leading up to deceleration, but flatten significantly afterwards; with this trend being more pronounced in larger models. Furthermore, loss landscapes along $\Dw$ appear much sharper in the beginning of training for larger models.
    }
    \label{fig.app.landscapes}
\end{figure}

\begin{figure}[h]
     \centering
     \captionsetup{width=\textwidth}
     \begin{subfigure}[b]{0.3\textwidth}
         \centering
         \includegraphics[trim={3.8cm 4cm 4cm 4cm},clip,width=\linewidth]{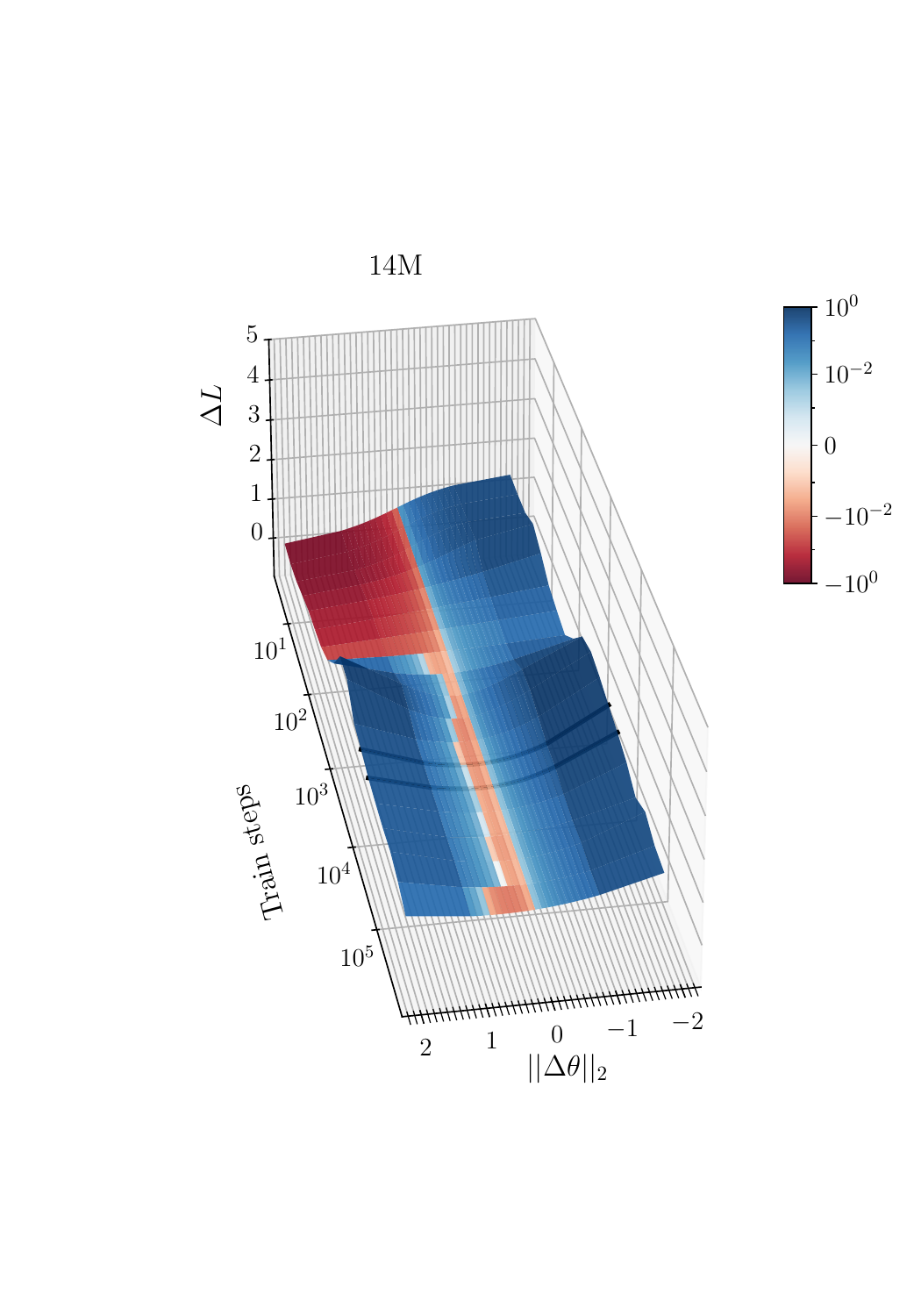}
     \end{subfigure}
     \hfill
     \begin{subfigure}[b]{0.3\textwidth}
         \centering
         \includegraphics[trim={3.8cm 4cm 4cm 4cm},clip,width=\linewidth]{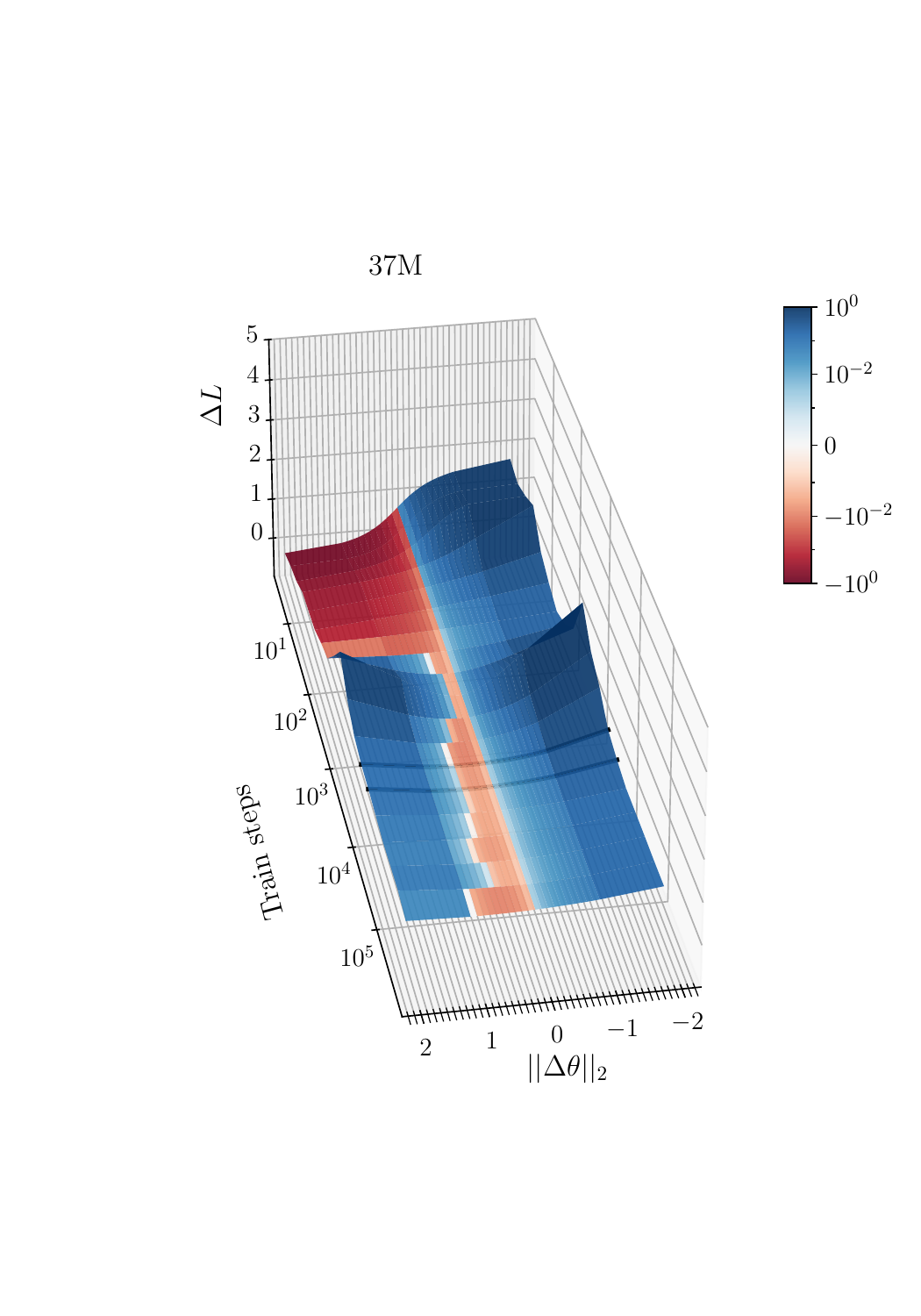}
     \end{subfigure}
     \hfill
      \begin{subfigure}[b]{0.3\textwidth}
         \centering
         \includegraphics[trim={3.8cm 4cm 4cm 4cm},clip,width=\linewidth]{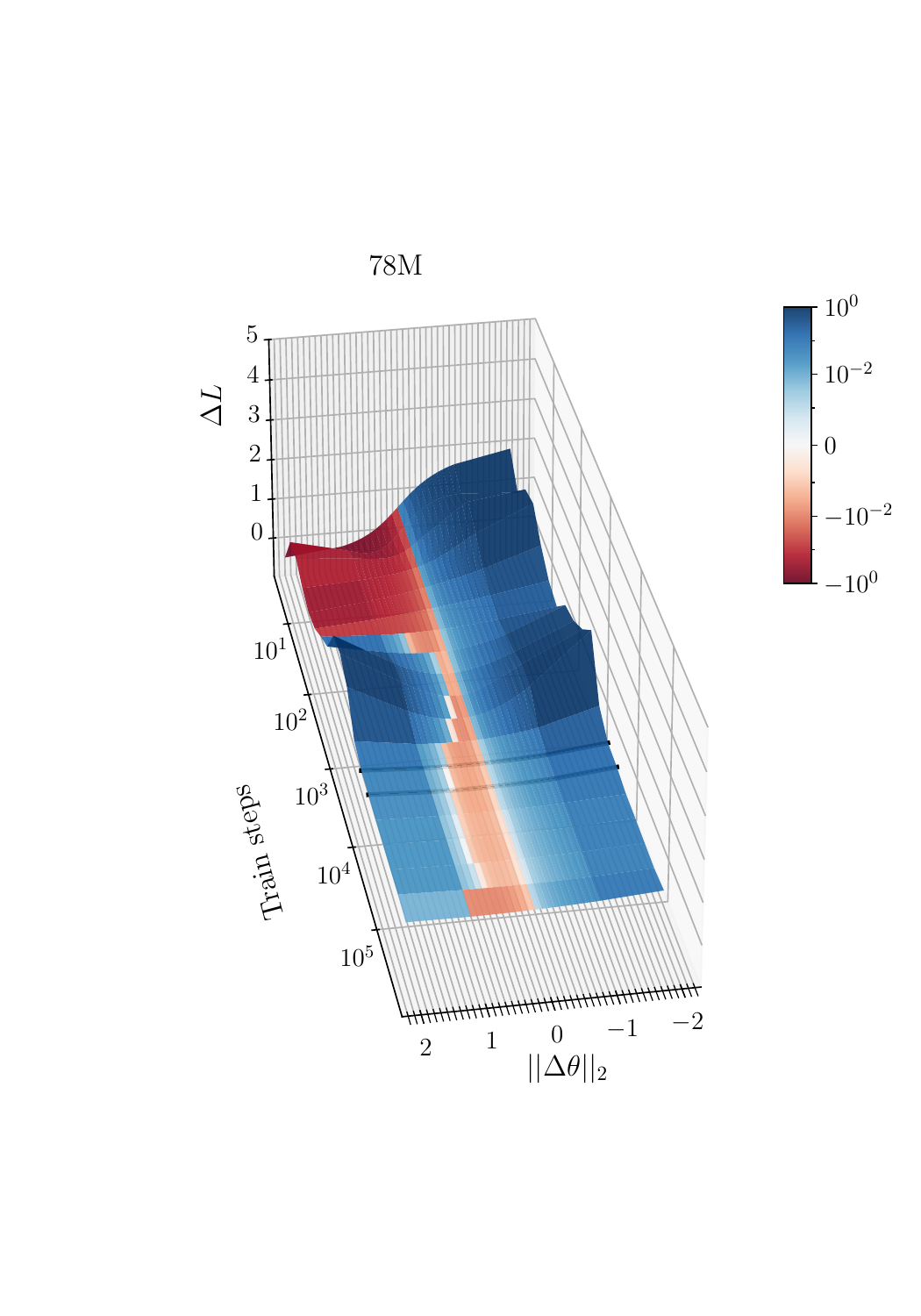}
     \end{subfigure}
     \begin{subfigure}[b]{0.31\textwidth}
         \centering
         \includegraphics[trim={3.8cm 4cm 4cm 4cm},clip,width=\linewidth]{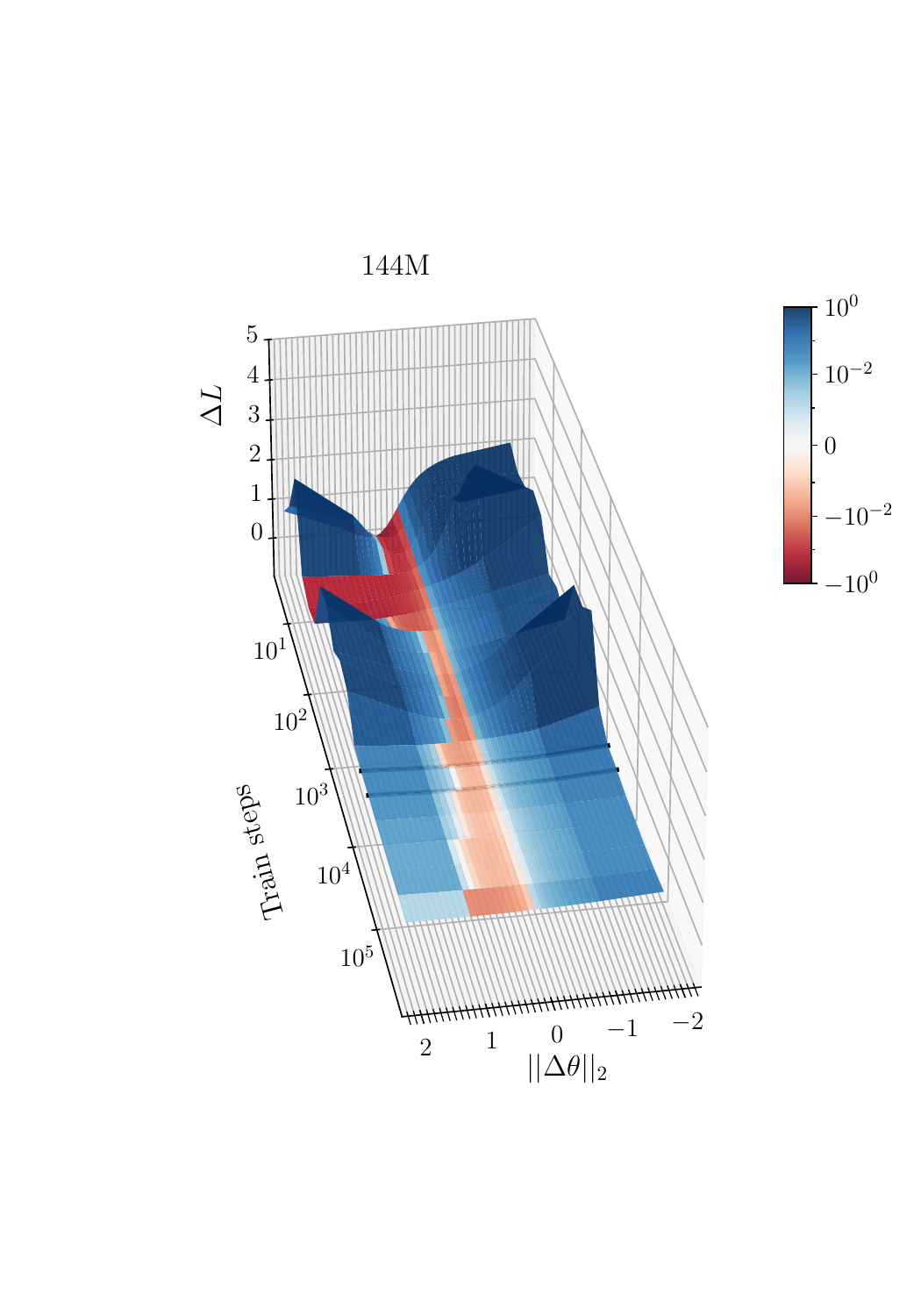}
     \end{subfigure}
     \hfill
     \begin{subfigure}[b]{0.31\textwidth}
         \centering
         \includegraphics[trim={3.8cm 4cm 4cm 4cm},clip,width=\linewidth]{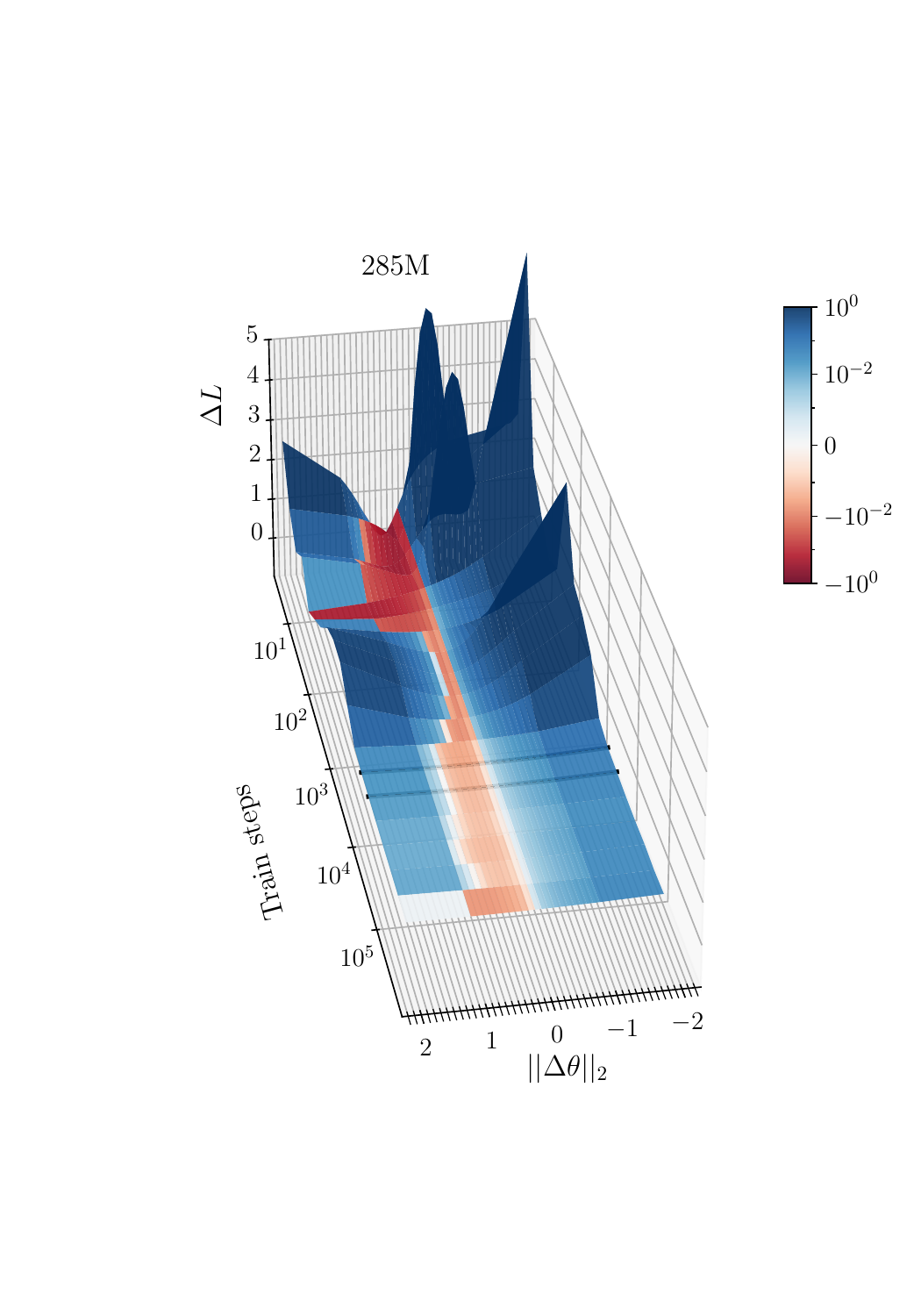}
     \end{subfigure}
     \hfill
      \begin{subfigure}[b]{0.31\textwidth}
         \centering
         \includegraphics[trim={3.8cm 4cm 4cm 4cm},clip,width=\linewidth]{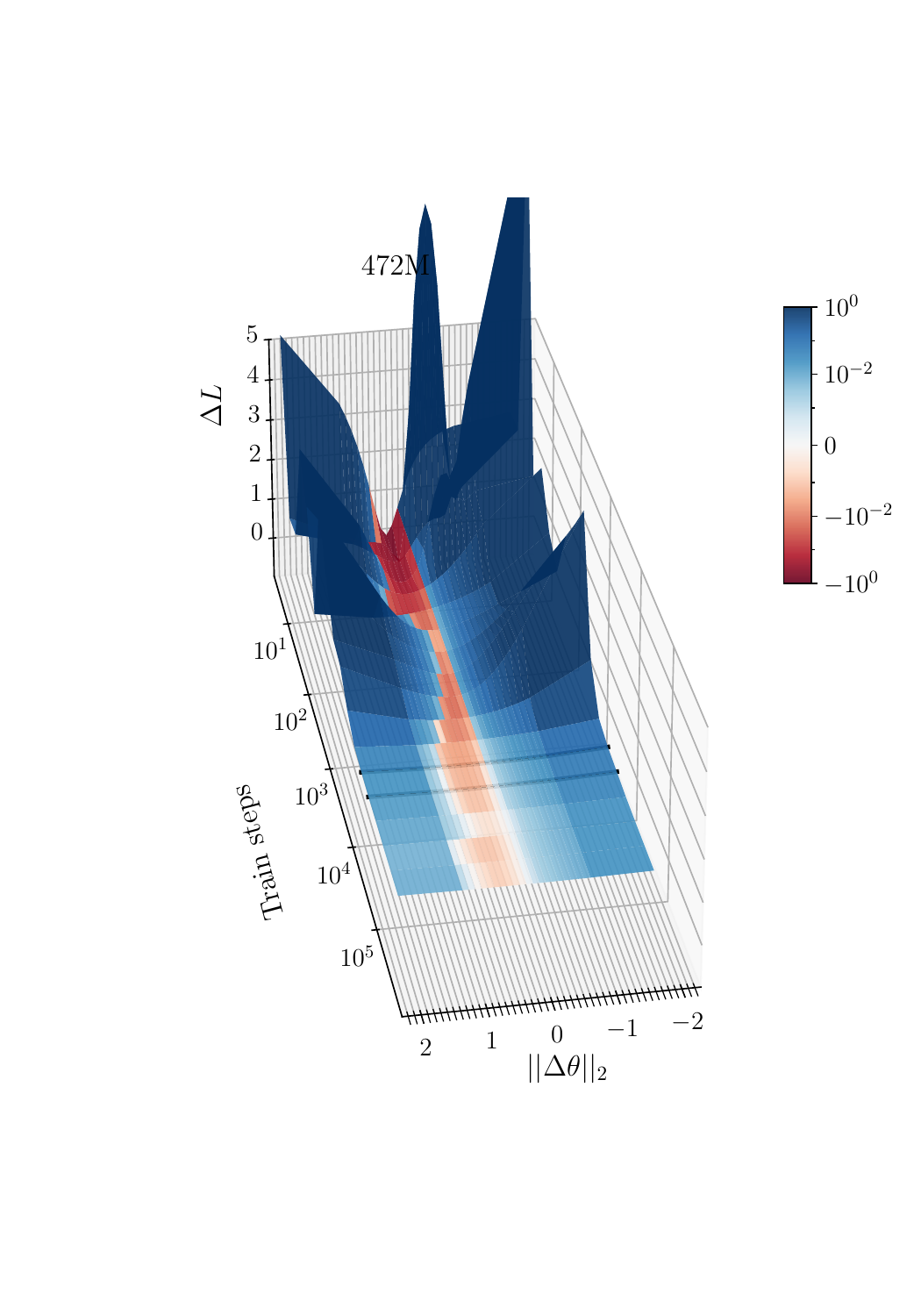}
     \end{subfigure}
     
    \caption{
    \textbf{Overall loss landscapes (cross section along $\Dw$), visualized throughout training (zoomed in)}  \hfill
    We plot the same data as in \reffig{fig.app.landscapes}, but zoomed into a narrower range. 
    }
    \label{fig.app.landscapes_zoomed}
\end{figure}

\clearpage
\subsection{Language Model Scaling Law Grounded in Loss Deceleration}
\label{app.decel_scaling_laws}
\begin{figure}[!h]
     \centering
     \captionsetup{width=0.7\linewidth}
     \begin{subfigure}[b]{0.25\linewidth}
         \centering
         \includegraphics[width=\linewidth]{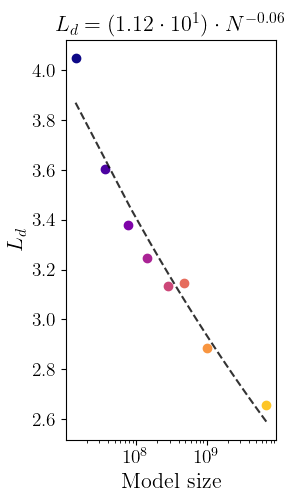}
     \end{subfigure}
     \begin{subfigure}[b]{0.25\linewidth}
         \centering
         \includegraphics[width=\linewidth]{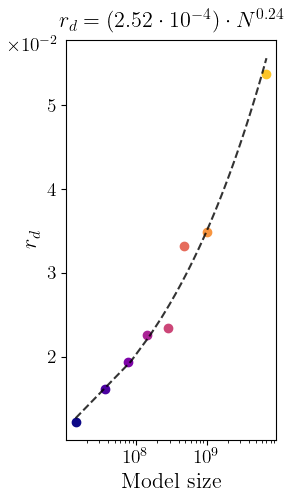}
     \end{subfigure}
     \begin{subfigure}[b]{0.25\linewidth}
         \centering
         \includegraphics[width=\linewidth]{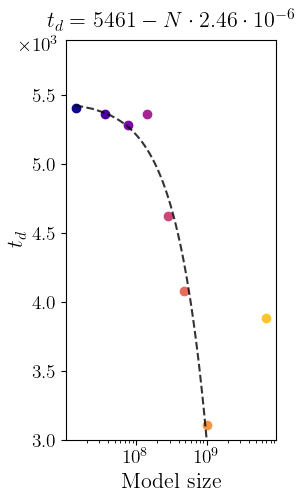}
     \end{subfigure}
    \caption{Power law and linear scaling in deceleration parameters.}
    \label{fig.app.decel_powerlaw}
\end{figure}

\parnl{Defining and fitting a scaling law grounded in loss deceleration}
Let $L\pp{N,T}$ be a scaling law for language model loss $L$, where $N$ is the number of model parameters and $T$ is number of training steps (with dataset size $D=T \cdot B$ for batch size $B$, i.e. single-epoch training).
Recall from \refeqn{eqn.loss_estimate_decel} that an estimate of the loss $L$ can be expressed in terms of the following parameters: \textbf{(1)} the number of steps at which deceleration occurs $t_d$; \textbf{(2)} the loss at which deceleration occurs $L_d$; and \textbf{(3)} the log-log rate of loss improvement after deceleration $r_d$. These parameters, shown in \reftab{tab.decel_measurements}, are dependent on $N$, such that we can define a scaling law $L\pp{N,T}$ grounded in loss deceleration as follows:
%
\begin{equation}\label{eqn.app.decel_scalinglaw}
    L\pp{N,T} = L_d\pp{N} \cdot t_d\pp{N}^{r_d\pp{N}} \cdot {T}^{\smin r_d\pp{N}}
\end{equation}
%
In \reffig{fig.app.decel_powerlaw}, we observe, with the admittedly limited datapoints from our experiments, that $L_d$ and $r_d$ seem to exhibit power law scaling. In contrast, $t_d$ appears to scale linearly if the outlier value for OLMo-$7$B, which is likely a result of being trained with \num{5000} warmup steps instead of \num{2000}, is omitted. This suggests that warmup steps, among potentially other hyperparameters, have an important role not accounted for here. However, these results are preliminary and intended more as an exploratory proof of concept, included here for completeness, rather than a key result or claim of the paper. We leave the costly task of conducting sufficient training runs to more adequately validate this functional form for future work.

\clearpage
\section{Additional Results (Ablations)}
\label{app.ablation}
\subsection{Effect of Batch Size and Sequence Length}
\label{app.ablation_bsz_seq}
\begin{figure}[!h]
     \centering
     \captionsetup{width=1\linewidth}
     \begin{subfigure}[b]{0.35\linewidth}
         \centering
         \includegraphics[width=\linewidth]{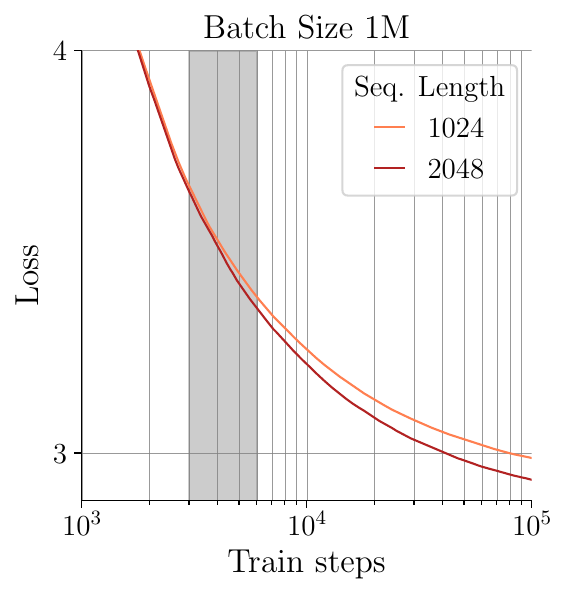}
     \end{subfigure}
     \begin{subfigure}[b]{0.35\linewidth}
         \centering
         \includegraphics[width=\linewidth]{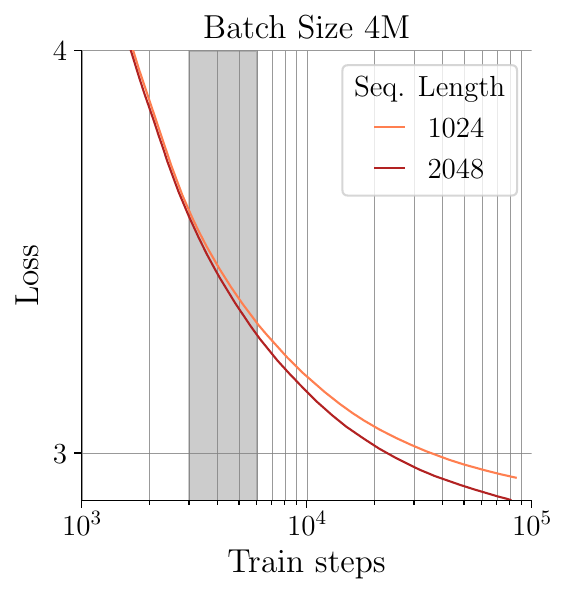}
     \end{subfigure}
     \begin{subfigure}[b]{0.35\linewidth}
         \centering
         \includegraphics[width=\linewidth]{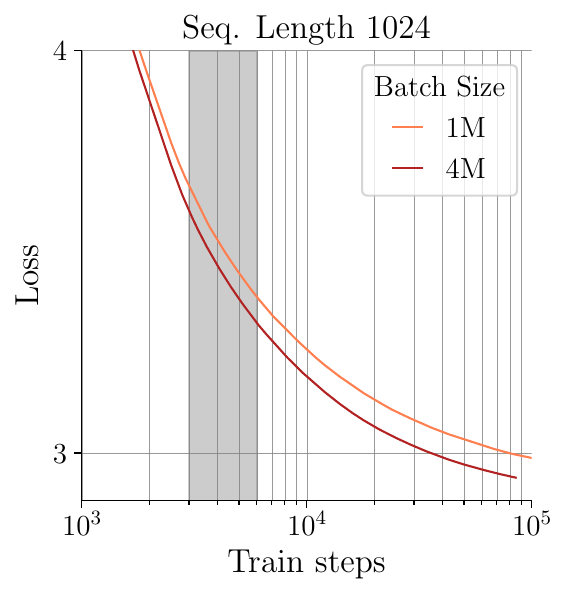}
     \end{subfigure}
     \begin{subfigure}[b]{0.35\linewidth}
         \centering
         \includegraphics[width=\linewidth]{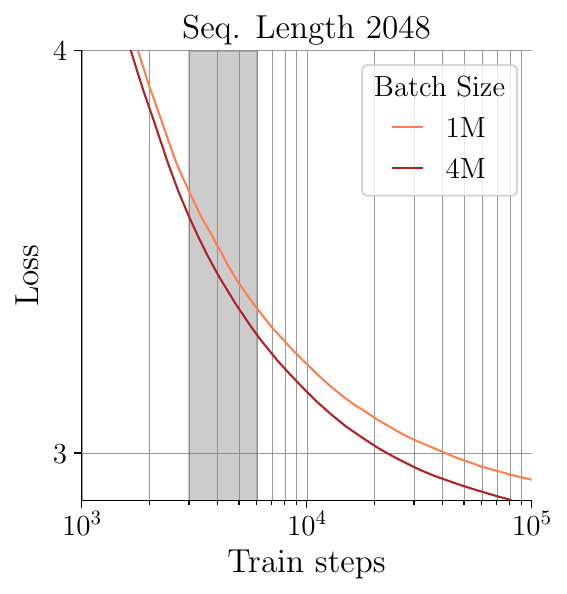}
     \end{subfigure}
\caption{Increasing batchsize improves loss at deceleration $L_d$, but shows similar post-deceleration rate of loss improvement $r_d$. In contrast, increasing sequence length results in improved $r_d$ and similar $L_d$.}
\label{fig.app.bszseq}
\end{figure}

\paragraph{Effect on deceleration}
In this set of experiments, we vary the batch size (1M and 4M tokens) and the sequence length (1024 and 2048 tokens) for our 144M model. The results are presented in \reffig{fig.app.bszseq} and \reftab{tab.app.bnsl_params_bsz_seq}.
In terms of loss deceleration, increasing the batch results in improved loss at deceleration $L_d$ but similar post-deceleration rate of loss improvement $r_d$. In contrast, increasing sequence length results in improved $r_d$ and similar $L_d$. 

Note that these results are obtained from training curves, i.e. losses are computed on training batches that differ in batch size and sequence length. Hence, the differences in results can be partially attributed to differences in the data used to compute the losses. Similarly, the results for OLMo-1B and OLMo-7B in \reffig{fig.bnsl_loss_fit} were obtained from training curves due to limited checkpoint availability. The apparent performance gap between them and smaller models can also be partially attributed to increases in batch size and sequence length. 

\begin{table}[!h]
  \caption{Deceleration measurements, BNSL fit parameters and errors for different batch sizes and sequence lengths.}
  \label{tab.app.bnsl_params_bsz_seq}
  \centering
  \resizebox{\columnwidth}{!}{
    \begin{tabular}{llcccccccccccc}
    \toprule
    {Batch Size} & {Seq. Len} 
    && $L_d$  & $t_d$  & $r_d$ 
    && $b$ & $c_0$ & $c_1$ & $\log(d_1)$ & $f_1$ & RSLE \\
    \midrule
    1M & 1024  
    && $3.28$ & $5120$ & $0.033$
    && $21.39 \pm 0.30$ & $0.22 \pm 0.00$ & $-0.19 \pm 0.00$ & $8.54 \pm 0.03$ & $0.25 \pm 0.04$ & $0.016$ 
    \\
     & 2048  
    && $3.24$ & $5331$ & $0.034$
    && $22.06 \pm 0.34$ & $0.22 \pm 0.00$ & $-0.19 \pm 0.00$ & $8.58 \pm 0.03$ & $0.25 \pm 0.04$ & $0.018$ 
    \\ \midrule
    4M & 1024  
    && $3.26$ & $4893$ & $0.038$
    && $21.56 \pm 0.32$ & $0.22 \pm 0.00$ & $-0.18 \pm 0.00$ & $8.50 \pm 0.03$ & $0.20 \pm 0.05$ & $0.017$ 
    \\
     & 2048  
    && $3.23$ & $4968$ & $0.041$
    && $22.11 \pm 0.35$ & $0.23 \pm 0.00$ & $-0.18 \pm 0.00$ & $8.51 \pm 0.04$ & $0.21 \pm 0.05$ & $0.019$ 
    \\
    \bottomrule
  \end{tabular}
  }
\end{table}

\begin{figure}[!h]
     \centering
     \captionsetup{width=1\linewidth}
     \begin{subfigure}[b]{0.23\linewidth}
         \centering
         \includegraphics[width=\linewidth]{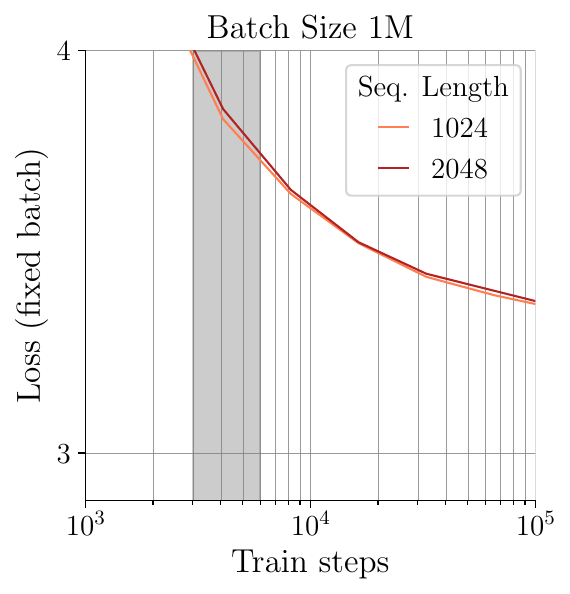}
     \end{subfigure}
     \begin{subfigure}[b]{0.23\linewidth}
         \centering
         \includegraphics[width=\linewidth]{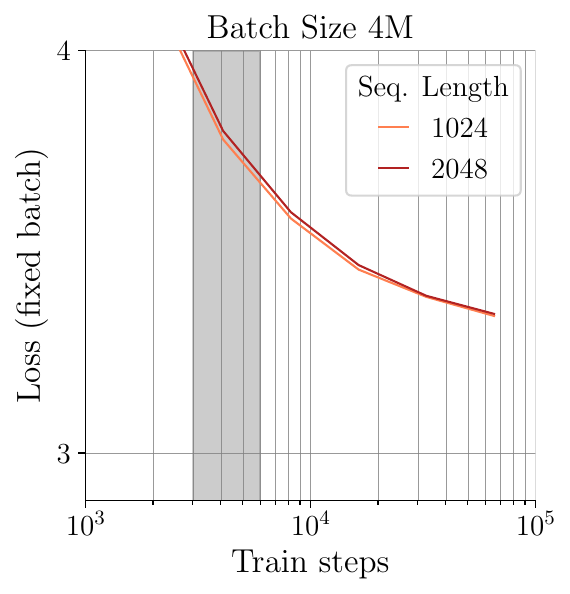}
     \end{subfigure}
     \begin{subfigure}[b]{0.23\linewidth}
         \centering
         \includegraphics[width=\linewidth]{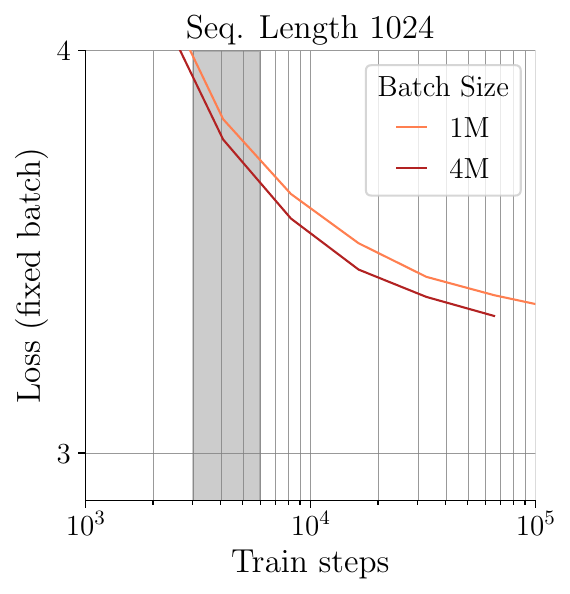}
     \end{subfigure}
     \begin{subfigure}[b]{0.23\linewidth}
         \centering
         \includegraphics[width=\linewidth]{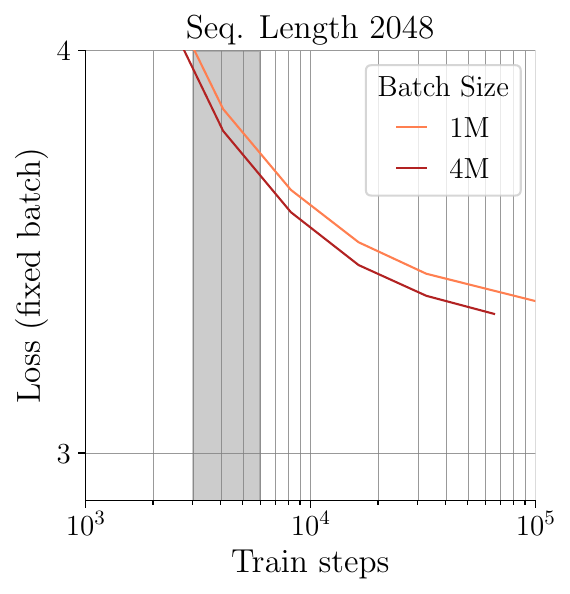}
     \end{subfigure}
\caption{When measuring loss on withheld validation batches with fixed sequence length and batch size, we find that the improvements from increased sequence length disappear (left) while those from increased batch size remain (right). }
\label{fig.app.bszseq2}
\end{figure}

\clearpage
\paragraph{Note on sequence length} We reproduce \reffig{fig.app.bszseq} with fixed batches of size 0.5M and sequence length 1024 on saved checkpoints and show the results in \reffig{fig.app.bszseq2}. We find that improvements from increasing sequence length do not transfer, suggesting they are limited to tokens from longer sequences. This is not particularly surprising given that tokens with more context are typically easier to predict, which would account for the loss improvements seen before. However, what is surprising is that these improvements only appear after deceleration, suggesting that the ability to leverage this extra context to better predict later tokens emerges with ZSL.

\begin{figure}[!h]
     \centering
     \captionsetup{width=1\linewidth}
     \begin{subfigure}[b]{0.24\linewidth}
         \centering
         \includegraphics[width=\linewidth]{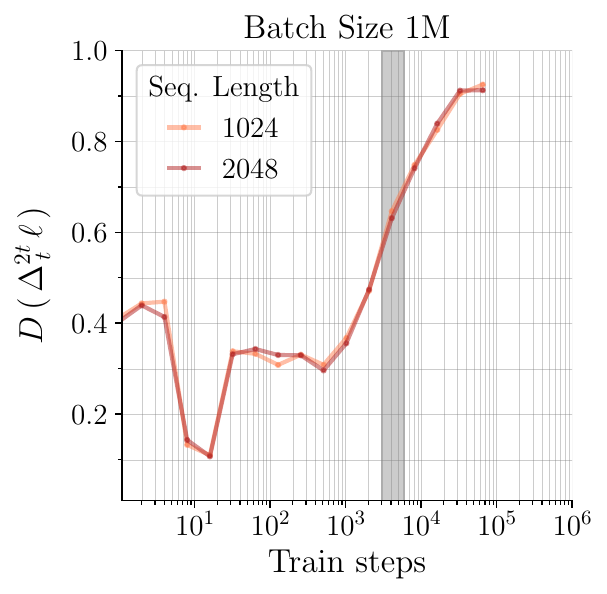}
     \end{subfigure}
     \begin{subfigure}[b]{0.24\linewidth}
         \centering
         \includegraphics[width=\linewidth]{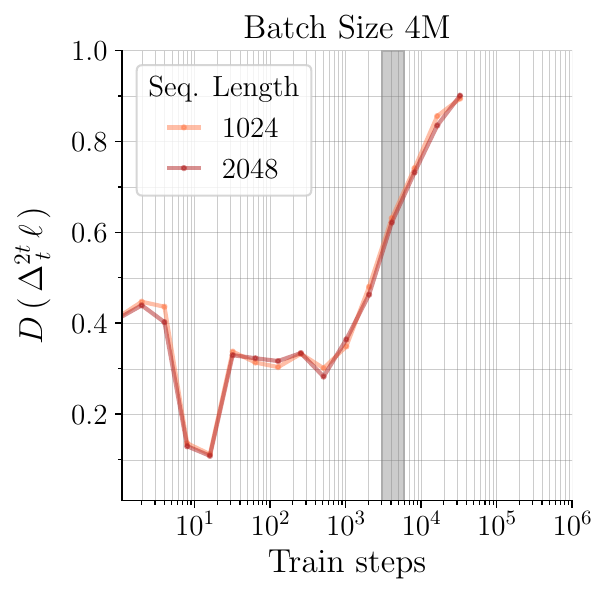}
     \end{subfigure}
     \begin{subfigure}[b]{0.24\linewidth}
         \centering
         \includegraphics[width=\linewidth]{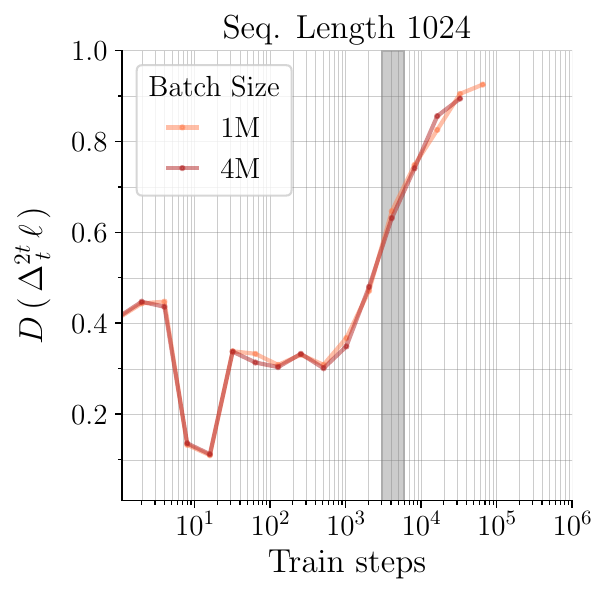}
     \end{subfigure}
     \begin{subfigure}[b]{0.24\linewidth}
         \centering
         \includegraphics[width=\linewidth]{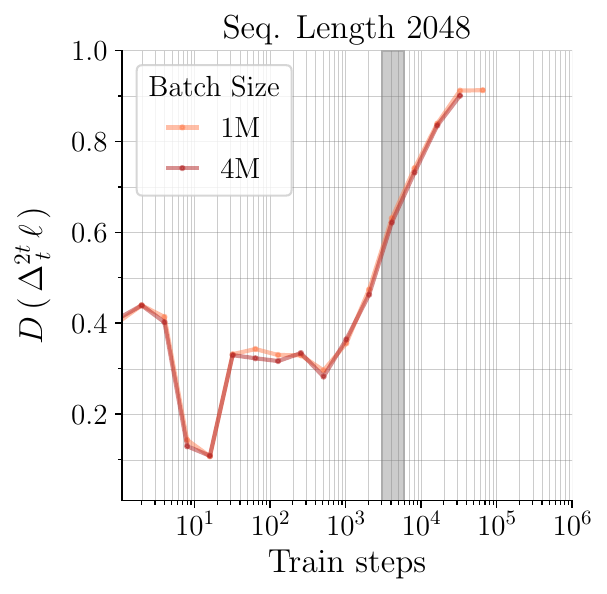}
     \end{subfigure}
\caption{Increasing batch size and sequence length both have no discernible effect on ZSL as measured by destructive interference in loss improvements.}
\label{fig.app.bszseq_ldi}
\end{figure}

\paragraph{Effect on zero-sum learning} We measure destructive interference in loss improvements as in \reffig{fig.zsl.zsl_multistep_log2_DI} and show the results in \reffig{fig.app.bszseq_ldi}. Results are consistent across several batch sizes and sequence lengths, and these hyperparameters, despite their different effects on train and validation loss curves, do not have an effect on zero-sum learning. In the case of sequence length, this is consistent with the observation that sequence length does not improve performance. 
In the case of batch sizes, this is consistent with the observation that improvements are established \textit{before} deceleration and would therefore not involve mitigating zero-sum learning.

\clearpage
\subsection{Effect of Optimizer Variants}
\label{app.ablation_optimizer}
In this set of experiments, we compare different optimizers to the baseline AdamW \cite{loshchilov_decoupled_2019} used in our main experiments, with the 144M model. The tested optimizers include AdEMAMix \cite{pagliardini_ademamix_2024}, and Muon \cite{jordan_muon_2024}. Furthermore, these results are obtained with NanoGPT \cite{karpathy_nanogpt_2022} rather than OLMo as in our main results, primarily because modifying optimizers is less error-prone in this simpler framework and because it is the basis of the original Muon implementation by \citet{jordan_muon_2024}. 
This framework differs in several important aspects which help corroborate the generality of our findings. In particular, these experiments use a different dataset--OpenWebText2 \cite{gao_pile_2020}--and a different model architecture and vocabulary based on GPT-2 \cite{radford_language_2019}. 
\\

\noindent
\begin{minipage}[t]{0.5\textwidth}
\parnl{Effect on deceleration}
Surprisingly, Muon and AdEMAMix have qualitatively different effects on loss deceleration. While Muon appears to improve early performance by accelerating deceleration with smaller $t_d$, AdEMAMix appears to improve performance by improving rate of loss improvement after deceleration with larger $r_d$. 
\\

Furthermore, despite faster convergence, Muon exhibits significant training instabilities after deceleration, coinciding with large spikes in gradient norms. While this has, to the best of our knowledge, not been reported elsewhere, these results were obtained using the original implementation of \citet{jordan_muon_2024}. One key difference which might account for these instabilities is the fact that we don't decay learning rate across our experiments. 
\\
\\
\end{minipage}\hfill
\begin{minipage}[t]{0.45\textwidth}
 \centering\raisebox{\dimexpr \topskip-\height}{
    \includegraphics[width=0.9\linewidth]{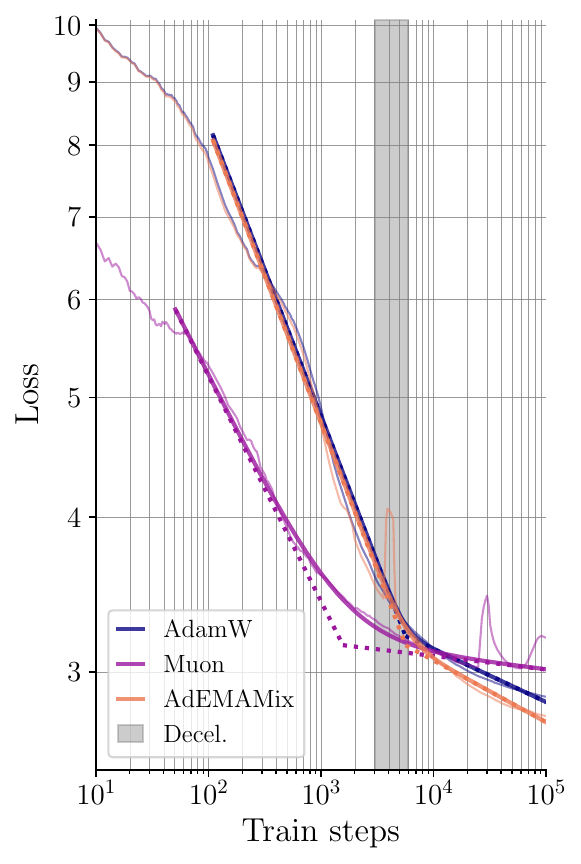}
 }
 \captionof{figure}{
   Loss curves and BNSL fits.
}
 \label{
    fig.app.bnsl_loss_fit_optimizers_144M
}
\end{minipage}

\vspace{1em}
\begin{table}[!h]
  \caption{Deceleration measurements; BNSL fit parameters and errors for different optimizers.}
  \label{tab.app.bnsl_params_optimizer}
  \centering
  \resizebox{\columnwidth}{!}{
    \begin{tabular}{lcccccccccc}
    \toprule
    Model & $L_d$ & $t_d$ & $r_d$ && $b$ & $c_0$ & $c_1$ & $\log(d_1)$ & $f_1$ & RSLE \\ 
    \midrule
    AdamW & $3.21$ & $5650$ & $0.043$ && $24.77 \pm 0.46$ & $0.24 \pm 0.00$ & $-0.19 \pm 0.00$ & $8.64 \pm 0.04$ & $0.15 \pm 0.06$ & $0.026$ \\ 
    Muon &  $3.15$ & $1569$ & $0.011$  && $12.04 \pm 0.15$ & $0.18 \pm 0.00$ & $-0.17 \pm 0.00$ & $7.36 \pm 0.04$ & $0.71 \pm 0.05$ & $0.010$ \\ 
    AdEMAMix & $3.15$ & $5673$ & $0.050$ && $24.73 \pm 0.53$ & $0.24 \pm 0.00$ & $-0.19 \pm 0.01$ & $8.64 \pm 0.05$ & $0.25 \pm 0.07$ & $0.025$ \\ 
    \bottomrule
  \end{tabular}
  }
\end{table}

\clearpage
\paragraph{Effect on zero-sum learning}
We find that the effects of AdEMAMix and Muon on zero-sum learning are consistent with the previously observed effect on loss deceleration. 
Specifically, destructive interference in loss improvements rises notably faster in Muon, consistent with the faster onset of loss deceleration observed. Furthermore, we observe significant oscillations after deceleration which correspond to the training instabilities observed as loss degrades closer to the end of training. 
In contrast, AdEMAMix behaves very similarly to AdamW leading up to deceleration, consistent with the fact that they both reach loss deceleration at similar step and loss values. However, after deceleration, AdEMAMix typically exhibits lower destructive interference than AdamW. In contrast to Muon which has similar oscillations, these are not attributable to training instability and loss degradation, and more likely to account for the improved rate of loss improvements observed in AdEMAMix after deceleration. 
These ablations were intended simply to verify the generality of our findings across different optimizers, however the results suggest that the considered optimizers differ in important ways with respect to zero-sum learning. 
While more comprehensive experiments are required to validate our preliminary findings, this suggests that zero-sum learning can indeed be mitigated independent of scale by targeting optimization and training dynamics directly.

\begin{figure}[!h]
     \centering
     \captionsetup{width=1\linewidth}
     \begin{subfigure}[b]{0.45\linewidth}
         \centering
         \includegraphics[width=\linewidth]{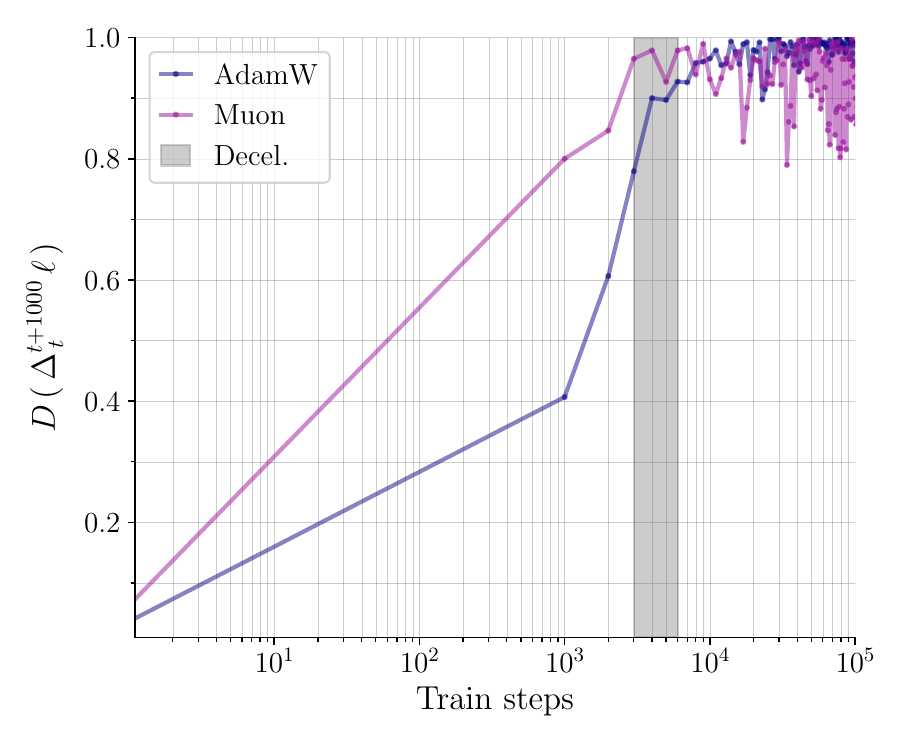}
     \end{subfigure}
     \begin{subfigure}[b]{0.45\linewidth}
         \centering
         \includegraphics[width=\linewidth]{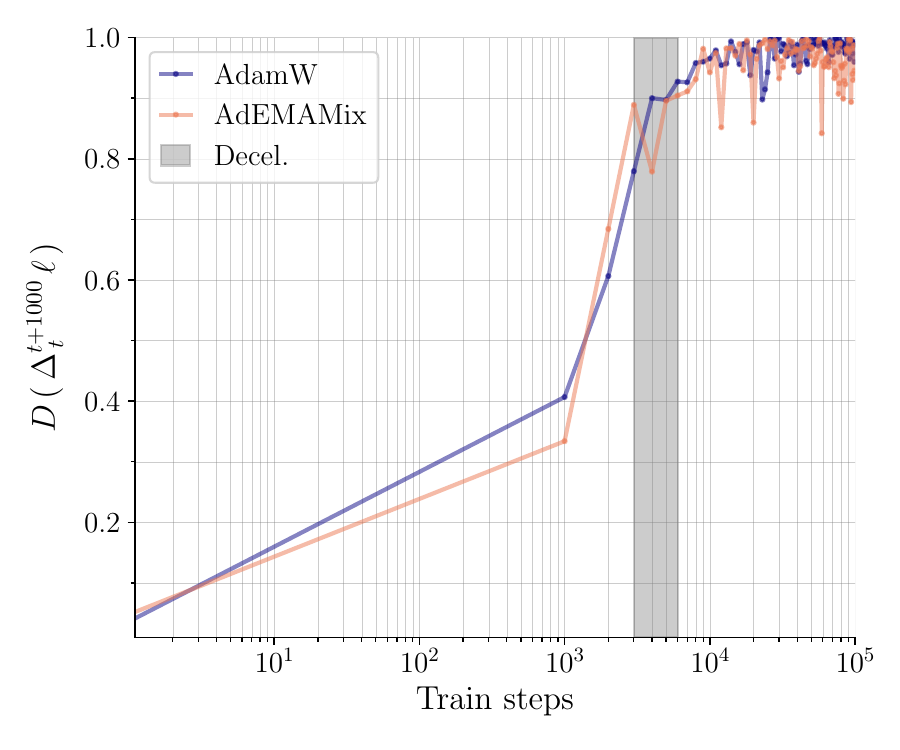}
     \end{subfigure}
\caption{Destructive interference in loss improvements for AdEMAMix and Muon, compared to baseline AdamW}
\label{fig.app.optimizers_ldi}
\end{figure}

\vspace{2em}
\noindent
\begin{minipage}[t]{0.45\textwidth}
\parnl{Note on Cautious AdamW}
We initially included Cautious or C-AdamW \cite{liang_cautious_2025} in our analysis, however the results were effectively indistinguishable from our baseline. This is consistent with the original results reported by \citet{liang_cautious_2025} (\reffig{fig.app.cadamw_results}), where C-AdamW only improves performance on the 1B model (which itself underperforms the 350M model AdamW baseline).

\end{minipage}\hfill
\begin{minipage}[t]{0.5\textwidth}
 \centering\raisebox{\dimexpr \topskip-\height}{
 \includegraphics[width=1\linewidth]{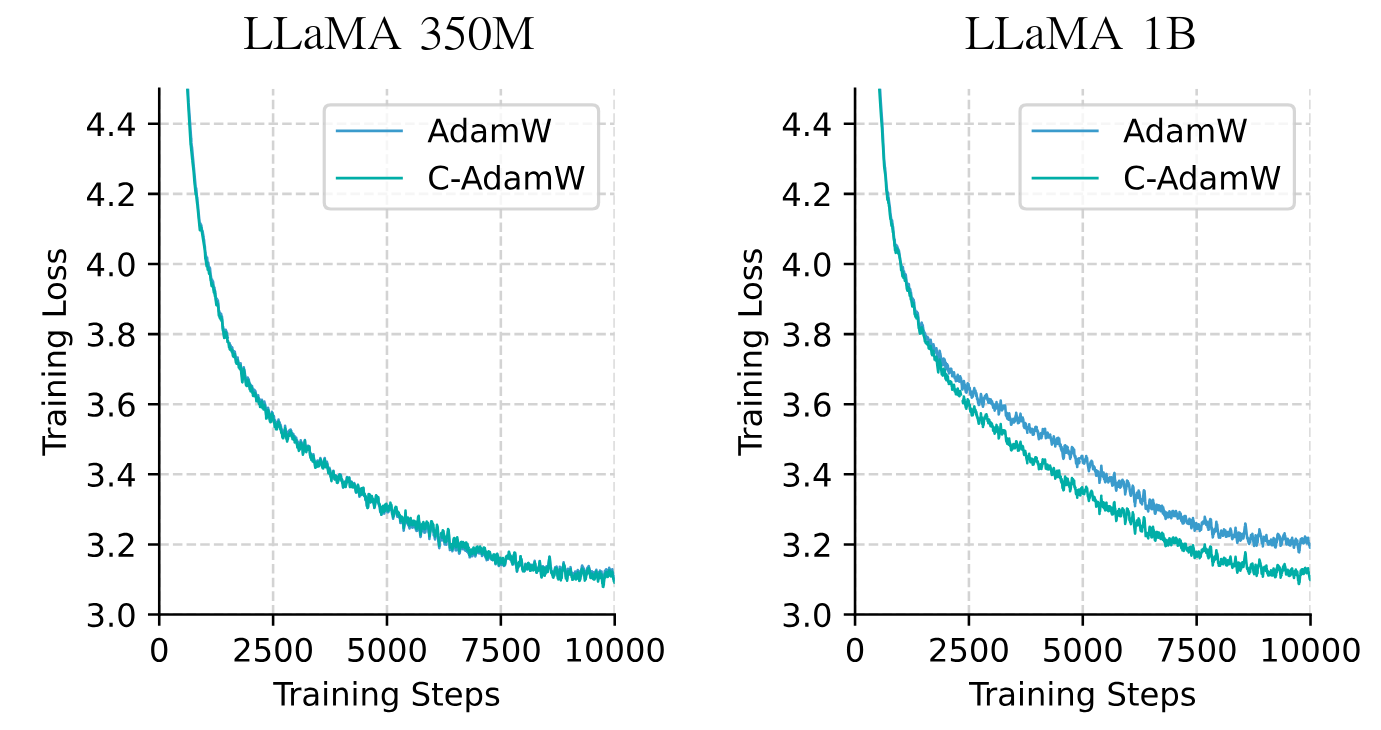}
 }
 \captionof{figure}{Original C-AdamW results \cite{liang_cautious_2025}.}
 \label{fig.app.cadamw_results}
\end{minipage}

\clearpage
\subsection{Effect of Learning Rate Decay}
\label{app.decel_cosine}
Our main results are for training runs where learning rate was warmed up and held constant, in line with \citet{hagele_scaling_2024} and \citet{wen_understanding_2024}. However, typically scaling experiments have been conducted with learning rate decay. In particular, \citet{hoffmann_training_2022} note that consistently decaying to $0.1$ of the peak learning rate as an important difference to \citet{kaplan_scaling_2020}, leading to different compute-optimal scaling. To rule out this potential confound, we replicate our experiments with a cosine learning rate decay in line with \citet{hoffmann_training_2022} (and \citet{groeneveld_olmo_2024}), leaving all else equal.
\\

\noindent
\begin{minipage}[t]{0.55\textwidth}
 \parnl{Effect on loss deceleration}
\reffig{fig.bnsl_loss_fit} is replicated in \reffig{fig.app.bnsl_loss_fit_cosine}, with similar results and quality of fits.  \reftab{tab.app.bnsl_params} is replicated in \reftab{tab.app.bnsl_params_cosine} with again similar results, and generally smaller values for $c_1$, $\log\pp{d_1}$, and $f_1$. Lastly, \reftab{tab.decel_measurements} is replicated in \reftab{tab.app.decel_measurements_cosine}, where we see that $L_d$ resulting from the BNSL fit is increased, but this is offset by improved $r_d$ and $t_d$, leading to better final loss. This improvement in final loss appears to increase with model size, suggesting a complementary mechanism by which scale improves loss under learning rate decay, which is not accounted for by our principal findings.
\\\\
\captionof{table}{Deceleration measurements with lr decay.}
\label{tab.app.decel_measurements_cosine}
\begin{tabular}{lccccccccc}
\toprule
Model             && $L_d$          & $r_d$        & $t_d$      && $\hat{L}_T$ & $L_T$ 
\\
\midrule
14M && $4.08$ & $0.016$ & $5198$ & & $3.83$ & $3.86$ \\ 
37M && $3.65$ & $0.023$ & $5029$ & & $3.34$ & $3.36$ \\ 
78M && $3.45$ & $0.029$ & $4808$ & & $3.07$ & $3.09$ \\ 
144M && $3.35$ & $0.036$ & $4712$ & & $2.90$ & $2.92$ \\ 
285M && $3.28$ & $0.040$ & $3921$ & & $2.77$ & $2.78$ \\ 
472M && $3.24$ & $0.045$ & $3653$ & & $2.68$ & $2.69$ \\ 
\midrule
OLMo-1B && $2.89$ & $0.035$ & $3106$ & & $2.39$ & $2.38$ \\ 
OLMo-7B && $2.66$ & $0.054$ & $3885$ & & $2.03$ & $2.02$ \\ 
\bottomrule
\end{tabular}
\end{minipage}\hfill
\begin{minipage}[t]{0.42\textwidth}
 \centering\raisebox{\dimexpr \topskip-\height}{
    \includegraphics[width=\linewidth]{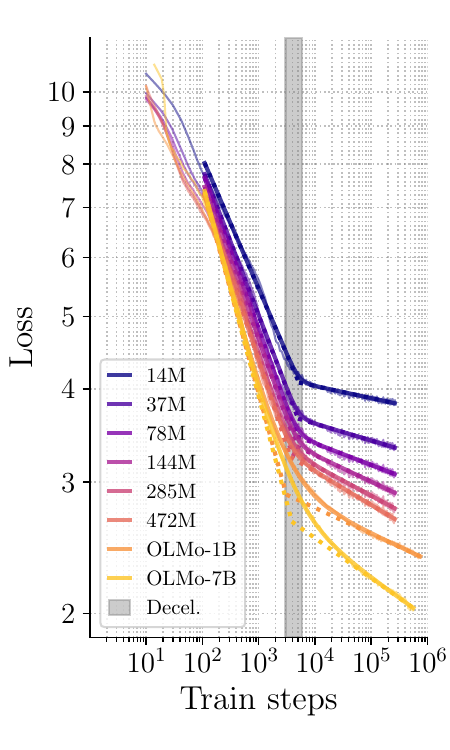}
 }
 \captionof{figure}{
   Loss curves and BNSL fits when training with cosine lr decay.
}
 \label{fig.app.bnsl_loss_fit_cosine}
\end{minipage}\hfill
\vspace{2em}

\begin{table}[!h]
  \caption{BNSL parameters and error when training with cosine lr decay.}
  \label{tab.app.bnsl_params_cosine}
  \centering
  \resizebox{0.8\columnwidth}{!}{
    \begin{tabular}{lcccccc}
    \toprule
    Model & $b$ & $c_0$ & $c_1$ & $\log(d_1)$ & $f_1$ & RSLE \\ 
    \midrule
    14M & $18.32 \pm 0.16$ & $0.18 \pm 0.00$ & $-0.16 \pm 0.00$ & $8.56 \pm 0.02$ & $0.16 \pm 0.03$ & $0.012$ \\ 
    37M & $19.60 \pm 0.22$ & $0.20 \pm 0.00$ & $-0.17 \pm 0.00$ & $8.52 \pm 0.02$ & $0.18 \pm 0.03$ & $0.014$ \\ 
    78M & $20.67 \pm 0.24$ & $0.21 \pm 0.00$ & $-0.18 \pm 0.00$ & $8.48 \pm 0.02$ & $0.22 \pm 0.03$ & $0.014$ \\ 
    144M & $20.31 \pm 0.25$ & $0.21 \pm 0.00$ & $-0.18 \pm 0.00$ & $8.46 \pm 0.03$ & $0.24 \pm 0.03$ & $0.014$ \\ 
    285M & $20.87 \pm 0.28$ & $0.22 \pm 0.00$ & $-0.18 \pm 0.00$ & $8.27 \pm 0.03$ & $0.31 \pm 0.03$ & $0.013$ \\ 
    472M & $21.30 \pm 0.29$ & $0.23 \pm 0.00$ & $-0.18 \pm 0.00$ & $8.20 \pm 0.03$ & $0.31 \pm 0.03$ & $0.013$ \\ 
    \midrule
    OLMo-1B & $26.53 \pm 0.42$ & $0.28 \pm 0.00$ & $-0.24 \pm 0.00$ & $8.04 \pm 0.03$ & $0.76 \pm 0.02$ & $0.008$ \\ 
    OLMo-7B & $28.14 \pm 0.54$ & $0.29 \pm 0.00$ & $-0.23 \pm 0.00$ & $8.26 \pm 0.04$ & $0.78 \pm 0.03$ & $0.008$ \\
    \bottomrule
  \end{tabular}
  }
\end{table}

\clearpage
\section{Related works}
\label{app.relworks}
This work connects several existing areas of research. 
In particular, several recent works attempt to explain scaling laws, typically from the perspective of intrinsic model capacity, long-tailed data distributions, and asymptotic behaviour \cite[e.g.][]{hutter_learning_2021,sharma_scaling_2022,michaud_quantization_2023,bahri_explaining_2024,bordelon_dynamical_2024}. In contrast, our goal is to identify a mechanism grounded in training dynamics that can be targeted independent of scale. 
The mechanism we identify, loss deceleration, is to the best of our knowledge not addressed in relevant prior works on e.g. loss plateaus \cite{yoshida_data-dependence_2020}, learning curve shapes \cite{viering_shape_2022}, or LM saturation \cite{godey_why_2024,mircea_gradient_2024}. 
Lastly, the study of training dynamics based on per-example gradient interactions remains under-explored, with related tangential works on e.g. improving multi-task learning \cite{liu_conflict-averse_2021}, or characterizing outliers in SGD \cite{rosenfeld_outliers_2023}.

\paragraph{Explaining scaling laws}\label{app.rw.esl}
Several works have proposed different explanations for neural scaling laws such as \cite{kaplan_scaling_2020,hoffmann_training_2022,caballero_broken_2023,hagele_scaling_2024,tissue_scaling_2024,everett_scaling_2024}. Notably, \citet{bahri_explaining_2024} explain scaling laws in terms of asymptotic behaviour, identifying variance-limited regimes based on concentration around  infinite limits, and resolution-limited regimes based on distances between train and test data points on their manifold (see also \cite{sharma_scaling_2022}). \citet{atanasov_scaling_2024} analytically explain power-law scaling in high-dimensional ridge regression with tools from random matrix theory.  \citet{michaud_quantization_2023} propose a "quantization model of neural scaling", whereby power law scaling is a result of (1) language models improving loss by learning discrete capabilities from their demonstration in data, (2) larger models being able to learn more capabilities, and (3) rarer capabilities improve loss by smaller and smaller amounts due to their vanishing frequency. Similarly, \citet{hutter_learning_2021} show how power law scaling with data can arise from long-tail feature distributions. 

\paragraph{Improving language models independently of scaling}\label{app.rw.siftd}
Recent work on e.g. data pruning \cite{marion_when_2023,sorscher_beyond_2022},
model distillation \cite{allen-zhu_towards_2023,gemma_team_gemma_2024}, 
and model pruning \cite{raposo_mixture--depths_2024} show that improvements obtained from scaling can be matched (up to a point) with other methods. This suggests that scaling may indirectly improve loss by its effect on training dynamics, and that similar effects/improvements can be obtained without necessarily scaling. 

\paragraph{Gradient opposition}\label{sec.rw.gradopp} 
From the perspective of training dynamics, \citealp{rosenfeld_outliers_2023} discuss the effect of outlier samples with opposing gradients. In the context of multi-task learning, several works have proposed approaches to mitigate gradient opposition between tasks, e.g. \cite{parascandolo_learning_2020,yu_gradient_2020,liu_conflict-averse_2021}. Gradient opposition between tokens in language modeling has, to the best of our knowledge, not been characterized. Related but distinct, is the work of \citet{mircea_gradient_2024} characterizes opposition within token gradients rather than between different tokens. 

\paragraph{Loss deceleration and learning curves} 
To the best of our knowledge, the loss deceleration transition we identify and characterize in this work has not been previously established or explained.  We refer the reader to \citet{viering_shape_2022} for a comprehensive review of learning curve shapes, as well as \citet{hutter_learning_2021} and \citet{yoshida_data-dependence_2020} as examples of attempting to explain features in a learning curve.

\end{document}

%% file: tex/math_commands.tex
\newcommand{\smin}{\text{-}}

\newcommand{\pp}[1]{\left(#1\right)}

\newcommand{\sqnorm}[1]{\lVert#1\rVert_2}

\newcommand{\norm}[1]{\lVert#1\rVert}
\newcommand{\abs}[1]{\left| #1 \right|}
\newcommand{\abstt}[1]{| #1 |}

\makeatletter
\newcommand{\tsum}{\DOTSB\mathop{\tsum@}\nolimits}
\newcommand{\tsum@}{\mathchoice{\textstyle\sum}{\sum}{\sum}{\sum}}
\makeatother

\newcommand{\dotp}[2]{\left<{#1}, {#2}\right>}

\newcommand{\dotpB}[2]{\left<{#1}, {#2}\right>}
\newcommand{\dotpb}[2]{\langle{#1}, {#2}\rangle}
\newcommand{\cossim}[2]{\cos({#1},{#2})}

\newcommand{\mscript}[1]{\mbox{\scriptsize $#1$}} 

\newcommand{\vDI}{\vec{D}}

\newcommand{\vG}{{G}}
\newcommand{\vg}{{g}}
\newcommand{\vidx}[2]{#1\mscript{\left[#2\right]}}
\newcommand{\zslm}[4]{#1_{^{#2}}^{_{#3}}(#4)}

\newcommand{\cg}{\mathbf{C}_g}
\newcommand{\cu}{\mathbf{C}_{(u,g)}}
\newcommand{\cug}{\mathbf{C}_{(u,G)}}

\newcommand{\paramv}{\theta}

\newcommand{\dataset}{\mathcal{D}}

\newcommand{\Loss}{L}

\newcommand{\loss}{\ell}
\newcommand{\lossx}[1]{\loss_{#1}}

\newcommand{\dldwx}[1]{\nabla_{\paramv} \lossx{ #1 }}
\newcommand{\dLdw}{\nabla_{\paramv} \Loss}
\newcommand{\Dlx}[1]{\Delta \lossx{#1}}
\newcommand{\Dlxfote}[1]{\Tilde{\Delta} \lossx{#1}}
\newcommand{\DL}{\Delta \Loss}
\newcommand{\DLfote}{\Tilde{\Delta} \Loss}
\newcommand{\Dw}{\Delta \paramv}
\newcommand{\ptg}[2]{\Tilde{\nabla}_{\theta_{#2}} \lossx{#1}}

\newcommand{\Dltt}{\Delta_{\:t}^{2t} \loss}

\newcommand{\DI}[1]{D(#1)}

\newcommand{\MA}[1]{M(#1)}

\newcommand{\lossh}[1]{\Tilde{\ell}_{#1}}
\newcommand{\R}[1]{\mathbb{R}^{#1}}

%% file: tex/util_commands.tex
\newcommand{\parnl}[1]{\paragraph{#1} ~\\}

\newcommand{\refsec}[1]{\hyperref[#1]{Section~\ref*{#1}}}
\newcommand{\refapp}[1]{\hyperref[#1]{Appendix~\ref*{#1}}}
\newcommand{\refeqn}[1]{\hyperref[#1]{Eqn.~\ref*{#1}}}
\newcommand{\reffig}[1]{\hyperref[#1]{Fig.~\ref*{#1}}}
\newcommand{\reftab}[1]{\hyperref[#1]{Table~\ref*{#1}}}

\usepackage[colorinlistoftodos,prependcaption,textsize=tiny]{todonotes}
\usepackage{xargs}  
\newcommandx{\todomust}[2][1=]{\todo[linecolor=red,backgroundcolor=red!25,bordercolor=red,#1]{#2}}
\newcommandx{\todomaybe}[2][1=]{\todo[linecolor=orange,backgroundcolor=orange!25,bordercolor=orange,#1]{#2}}
\newcommandx{\tdwrite}[2][1=]{\todo[linecolor=blue,backgroundcolor=blue!25,bordercolor=blue,#1]{#2}}
\newcommandx{\tdformat}[2][1=]{\todo[linecolor=yellow,backgroundcolor=yellow!25,bordercolor=yellow,#1]{#2}}
\newcommandx{\tddone}[2][1=]{\todo[linecolor=green,backgroundcolor=green!25,bordercolor=green,#1]{#2}}
\newcommandx{\thiswillnotshow}[2][1=]{\todo[disable,#1]{#2}}